\documentclass[journal,onecolumn,draftclsnofoot]{IEEEtran}
\usepackage{blindtext, graphicx}
\usepackage{amsmath,amsfonts,amssymb,bbm}
\usepackage{algorithm,algorithmic}
\usepackage{epstopdf}
\usepackage{epsfig}
\graphicspath{{images/}}
\usepackage{mathtools,dsfont}
\usepackage{psfrag,bbm,graphicx}
\usepackage{comment,balance}
\usepackage{epstopdf}
\usepackage{epsfig}
\usepackage{multicol}
\usepackage{multirow, cite}
\usepackage{amsfonts}
\usepackage{color}
\usepackage{hyperref}
\usepackage{mathrsfs}
\allowdisplaybreaks

\ifCLASSINFOpdf
  
\else
 
\fi

\newcounter{lecnum}
\renewcommand{\thepage}{\arabic{page}}
\renewcommand{\thesection}{\Roman{section}}
\renewcommand{\theequation}{\arabic{equation}}
\renewcommand{\thefigure}{\arabic{figure}}
\renewcommand{\thetable}{\thelecnum.\arabic{table}}

\newtheorem{theorem}{Theorem}
\newtheorem{lemma}{Lemma}
\newtheorem{proposition}[theorem]{Proposition}

\newtheorem{remark}{Remark}

\begin{document}
%
\title{Best Arm Identification in Restless Markov Multi--Armed Bandits}

\author{P. N. Karthik, Kota Srinivas Reddy, and Vincent Y. F. Tan\\
\normalsize{National University of Singapore}\\
\normalsize{Email: \href{mailto:karthik@nus.edu.sg}{karthik@nus.edu.sg}}, \href{mailto:ksreddy@nus.edu.sg}{ksreddy@nus.edu.sg}, \href{mailto:vtan@nus.edu.sg}{vtan@nus.edu.sg}}

\IEEEoverridecommandlockouts
\IEEEpubid{\makebox[\columnwidth]{978-1-5090-5356-8/17/\$31.00~\copyright~2017 IEEE \hfill}
	\hspace{\columnsep}\makebox[\columnwidth]{ }}

\maketitle

\begin{abstract}
We study the problem of identifying the best arm in a multi-armed bandit environment when each arm is a time-homogeneous and ergodic discrete-time Markov process on a common, finite state space. The state evolution on each arm is governed by the arm's transition probability matrix (TPM). 
A decision entity that knows the set of arm TPMs but not the exact mapping of the TPMs to the arms, wishes to find the index of the best arm as quickly as possible, subject to an upper bound on the error probability. The decision entity selects one arm at a time sequentially, and all the unselected arms continue to undergo state evolution ({\em restless} arms). For this problem, we derive the first-known problem instance-dependent asymptotic lower bound on the growth rate of the expected time required to find the index of the best arm, where the asymptotics is as the error probability vanishes. Further, we propose a sequential policy that, for an input parameter $R$, forcibly selects an arm that has not been selected for $R$ consecutive time instants. We show that this policy achieves an upper bound that depends on $R$ and is monotonically non-increasing as $R\to\infty$. The question of whether, in general, the limiting value of the upper bound as $R\to\infty$ matches with the lower bound, remains open. We identify a special case in which the upper and the lower bounds   match. Prior works on best arm identification have dealt with (a) independent and identically distributed observations from the arms, and (b) rested Markov arms, whereas our work deals with the more difficult setting of restless Markov arms. 
\end{abstract} 
\section{Introduction}
Consider a multi-armed bandit with $K\geq 2$ arms in which each arm is associated with a time-homogeneous and ergodic discrete-time Markov process evolving on a common, finite state space. The state evolutions on each arm are governed by the arm's transition probability matrix (TPM). Given a function $f$ on the common state space of the arms, we define the {\em best arm} as the arm with the largest average value of $f$, averaged over the arm's stationary distribution. A decision entity that has knowledge of the set of TPMs of the arms, but does not know the exact mapping of the TPMs to the arms, wishes to find the index of the best arm as quickly as possible, subject to an upper bound on the error probability. Our interest is in the asymptotics as the error probability vanishes.

The above problem, known popularly in the literature as {\em best arm identification}, is an instance of an optimal stopping problem in decision theory, and can be embedded within the framework of active sequential hypothesis testing studied in the classical works of Chernoff \cite{Chernoff1959} and Albert \cite{albert1961sequential}. Prior works on best arm identification have dealt with (a) independent and identically distributed ({\em i.i.d.}) observations from the arms, as in \cite{garivier2016optimal}, and (b) {\em rested} Markov arms \cite{moulos2019optimal} in which the arms yield Markovian observations and an arm evolves only when selected, and remains frozen otherwise. In this work, we extend the results of \cite{garivier2016optimal,Kaufmann2016,moulos2019optimal} to the more difficult setting when the unselected arms continue to evolve ({\em restless arms}). An examination of the results in \cite{garivier2016optimal,Kaufmann2016,moulos2019optimal} shows that given an error probability threshold $\epsilon>0$, the minimum expected expected time required to find the best arm (or, expected stopping time) with an error probability no more than $\epsilon$ grows as $O(\log (1/\epsilon))$ in the limit as $\epsilon \downarrow 0$. We anticipate a similar growth rate for the expected stopping time in the setting of restless arms. Our goal is to characterise or bound the exact constant multiplying $\log (1/\epsilon)$ for the setting of restless arms and attempt to achieve the lower bound of $O(\log (1/\epsilon))$ in the limit as $\epsilon \downarrow 0$. Additionally, we aim to devise an efficient policy that  achieves the fundamental limit. 

\subsection{A Preliminary Trembling Hand Model}
The continued evolution of the unselected arms in our work makes it necessary for the decision entity to keep a record of (a) the time elapsed since each arm was previously selected (the arm's {\em delay}), and (b) the state of  each arm recorded at its previous selection instant (the arm's {\em last observed state}). The notion of arm delays is superfluous when the arms are rested because the unobserved arms remain frozen. They are also redundant when the arms yield {\em i.i.d.} observations because the observation from an arm at any given time is independent of all its previous observations. When the arms are restless, the arm delays are non-negative, integer-valued, and introduce a countably-infinite dimension to the problem. In a related problem of identifying an anomalous or {\em odd} arm in a restless multi-armed bandit, Karthik and Sundaresan \cite{karthik2021detecting,karthik2021learning} demonstrated that the arm delays and the last observed states constitute a controlled Markov process. A key aspect of the works \cite{karthik2021detecting,karthik2021learning} is the notion of a {\em trembling hand} for selecting the arms that is defined in the system model. Probabilistically, a trembling hand with parameter $\gamma\in [0,1]$ selects an arm uniformly at random with probability $\gamma$, and selects the intended arm with probability $1-\gamma$. As the authors note in \cite{karthik2021detecting}, when $\gamma>0$, the controlled Markov process (of arm delays and last observed states) satisfies a key ergodicity property (see \cite[Lemma 1]{karthik2021detecting}) which is pivotal in establishing matching upper and lower bounds on the expected time to find the odd arm. An understanding of whether the lower bound for the case $\gamma=0$ admits a matching upper bound remains open.

\subsection{Forgoing the Trembling Hand Model and Constraining the Maximum Delay of Each Arm}
The trembling hand model of \cite{karthik2021detecting,karthik2021learning} implies that at any given time, the probability of selecting any arm is at least $ \gamma/K$, irrespective of the arm selection scheme used. 
In this work, we forgo the trembling hand assumption and allow for arm selection schemes to put zero mass on certain arms. We derive a lower bound on the expected time to find the best arm over {\em all} arm selection schemes. However, such a generic lower bound may not be achievable. In order to analyse the achievability of the lower bound, we restrict attention to those arm selection policies which, for an input parameter $R>K$, forcefully select an arm if its delay is equal to $R$. For this class of policies, we derive an upper bound in terms of $R$ and show that the sequence of upper bounds, one for each value of $R$, is monotonically non-increasing as $R$ increases. 
An advantage of our method over the methods of \cite{karthik2021detecting,karthik2021learning} is that while the arm selection schemes in \cite{karthik2021detecting,karthik2021learning} are not practically implementable, the arm selection schemes we propose in this paper are easy to implement. 

The question of whether, in general, the limiting value of the upper bounds matches the generic lower bound appears to be a difficult problem  and remains open. Notwithstanding this, we identify some special cases when the limiting value of the upper bounds matches with the generic lower bound. 

\subsection{Prior Works on Multi-Armed Bandits and Best Arm Identification}
The  problem of minimising regret for multi-armed bandits was introduced in the seminal work of Lai and Robbins~\cite{lai1985asymptotically} when each arm yields {\em i.i.d.} observations. Anantharam {\em et al.} \cite{anantharam1987asymptotically} extended the results of Lai and Robbins to the setting of rested arms. We refer the reader to \cite{bubeck2012regret} for an extensive survey of regret minimization problems. In \cite{mannor2004sample,garivier2016optimal}, the authors derived a lower bound on the sample complexity of best arm identification in multi-armed bandits with {\em i.i.d.} observations from arms. Many follow-up works proposed algorithms to achieve the lower bounds; notable among them are action-elimination algorithms \cite{even2002pac,karnin2013almost}, upper confidence bound (UCB) algorithms \cite{jamieson2014lil,audibert2010best}, and  lower and upper confidence bound (LUCB) algorithms \cite{kalyanakrishnan2012pac,kaufmann2013information}. Several extensions to the classical setup of best-arm identification in multi-armed bandits have appeared in the literature; notable among them are correlated bandits \cite{gupta2021best}, cascading bandits \cite{zhong2020best},  bandits with switching costs \cite{dekel2014bandits}, and bandits with corrupted rewards~\cite{zhong2021prob}.

For the problem of finding the best arm as quickly as possible subject to an upper bound on the error probability, the paper \cite{garivier2016optimal} provided a problem instance-dependent lower bound on the expected time required to find the best arm for the setting of {\em i.i.d.} observations from the arms, and a {\em track-and-stop} scheme that meets the lower bound asymptotically as the error probability vanishes. Moulos \cite{moulos2019optimal} extended the results of  \cite{garivier2016optimal} to the setting of rested arms with hidden Markov observations. However, the lower and the upper bounds in \cite{moulos2019optimal} differ by a constant multiplicative factor; the achievability analysis therein uses novel concentration inequalities for Markov processes derived by the author. For a related problem of finding the anomalous (or odd) arm in multi-armed bandits, the works \cite{karthik2020learning,karthik2021detecting,karthik2021learning} obtain matching upper and lower bounds on the expected time required to find the odd arm subject to an upper bound on the error probability. While \cite{karthik2020learning} studies the setting of rested arms,  the works \cite{karthik2021detecting,karthik2021learning} focus on the setting of restless arms. A key aspect of the works \cite{karthik2021detecting,karthik2021learning} is a certain trembling hand model for selecting the arms, motivated from a certain visual science experiment. The paper \cite{karthik2021detecting} assumes that the TPMs of the odd arm and the non-odd arms are known beforehand to the decision entity, whereas the works \cite{karthik2020learning,karthik2021learning} deal with the case when the arm TPMs are unknown.

The recent works \cite{deshmukh2018controlled,deshmukh2021sequential,prabhu2020sequential} study a more general problem of
sequential hypothesis testing in multi-armed bandits, some special cases of which are the problems of best arm identification and odd arm identification, in the context of \textit{i.i.d.} observations from each arm. An extension of the results of \cite{deshmukh2018controlled,deshmukh2021sequential,prabhu2020sequential} to the settings of rested and restless arms is a possible direction of future work.

\subsection{Contributions}
\label{sec:contribution}
In this paper, we study the problem of finding the best arm in a restless multi-armed bandit as quickly as possible, subject to an upper bound on the error probability. Our technical contributions are as follows:
\begin{itemize}
	\item[--] In Section \ref{sec:lower-bound}, we show that given any $\epsilon>0$, the expected time required to find the best arm with an error probability no more than $\epsilon$ grows as $O(\log (1/\epsilon))$ in the limit as $\epsilon \downarrow 0$. We explicitly characterise the problem instance-dependent constant multiplying $\log(1/\epsilon)$ that is a function of the arm TPMs.
	\item[--] For a problem instance $C$, the constant $T^{\star}(C)$ appearing in our lower bound is the solution of a sup-min optimisation problem, where the supremum is over all possible state-action occupancy measures associated with the countable-state controlled Markov process of arm delays and last observed states; see \eqref{eq:T_a^{star}} for details. The question of whether the supremum in the expression for $T^{\star}(C)$ lower bound is attained is still open. The key difficulty in showing this is the presence of the countably-infinite valued arm delays appearing in the expression for $T^{\star}(C)$, which makes further simplifications of the expression for $T^{\star}(C)$ difficult. In the prior works \cite{garivier2016optimal,Kaufmann2016,moulos2019optimal},  further simplification of the expression for the constants governing the lower bound is possible because the notion of arm delays is superfluous in the settings of those works.
	
	In a related problem of odd arm identification in restless multi-armed bandits, Karthik and Sundaresan  \cite{karthik2021detecting,karthik2021learning} show that under a trembling hand model for arms selection, the supremum in the expression for the lower bounds in these works may be further simplified by restricting the supremum to the class of all stationary arm selection policies; see \cite{karthik2021detecting,karthik2021learning} for more details. However, it is unclear if such a simplification is possible in the absence of the trembling hand (as is the setting of this paper). 
	\item[--]In our achievability analysis presented in Section \ref{sec:achievability}, to ameliorate the difficulty arising from the countably infinite-valued arm delays, we constrain the maximum delay of each arm to be no more than $R$, and focus our attention on those policies which select an arm forcibly if its delay equals $R$. This constraint on the maximum delay of each arm makes the state-action space finite and amenable to further analyses. To the best of our knowledge, this is the first work to analyse delay-constrained policies for restless multi-armed bandits.
	
	It is worth noting that the achievability analyses in the works \cite{karthik2021detecting,karthik2021learning} rely crucially on a key ergodicity property (see \cite[Lemma 1]{karthik2021detecting}) for the controlled Markov process of arm delays and last observed states that is satisfied under a {\em trembling hand} model for arms selection. This ergodicity property is pivotal to the achievability analysis in \cite{karthik2021detecting,karthik2021learning}, and it is unclear if the same property holds in the absence of the trembling hand.
	\item[--] We devise a policy that, for an input parameter $R$, forcibly pulls an arm if its delay equals $R$, and achieves an upper bound of the order $\Theta(\log(1/\epsilon))$. We show that the best (smallest) constant multiplying $\log(1/\epsilon)$ is equal to $1/T_R^\star(C)$ under the problem instance $C$; see \eqref{eq:T-R-star-C} for the exact expression of $T_{R}^{\star}(C)$. Our results imply that $1/T_R^{\star}(C)$ is a valid asymptotic upper bound on the expected stopping time for all integers $R$. 
	\item[--] We show that $T_R^{\star}(C)$ is non-decreasing in $R$, and that $T_R^*(C)\leq T^{\star}(C)$ for all $R$, thus implying that $\lim_{R\to\infty}\ T_{R}^{\star}(C)$ exists. Thus, the lower bound of our work is given by $T^{\star}(C)^{-1}$, whereas the upper bound is governed by $\lim_{R\to\infty} T_{R}^{\star}(C)^{-1}$. While it is certainly true that $\lim_{R\to\infty}\ T_{R}^{\star}(C) \leq T^{\star}(C)$, showing that, in general, this inequality is an equality seems to be a difficult problem and remains open. In the special case when the TPM of each arm has identical rows, which is akin to obtaining {\em i.i.d.} observations from the arms, the above inequality is an equality, thus leading to matching upper and lower bounds in this special setting. We are thus able to recover the results of the prior works \cite{Kaufmann2016, garivier2016optimal} for the case when the {\em i.i.d.} observations from each arm come from a finite alphabet.
\end{itemize}

\subsection{Paper Organisation}
The rest of this paper is organised as follows. In Section \ref{sec:notations}, we setup the notations and state our central goal. In Section \ref{sec:delays-last-observed-states}, we introduce the notions of arm delays, last observed states, and the Markov decision problem arising from the arm delays and the last observed states. We provide expressions for the log-likelihoods and the log-likelihood ratios, the basic quantities of analysis, in Section \ref{sec:LLR}, and present the asymptotic lower bound on the growth rate of the expected time required to find the best arm in Section \ref{sec:lower-bound}. In Section \ref{sec:achievability}, we constrain the maximum delay of each arm to be no more than $R$ for some $R\in \mathbb{N}\cap (K, \infty)$, and describe a policy that forcibly samples an arm whose delay equals $R$. We provide results on the performance of the policy in Section \ref{sec:performance-results}, and state the main result in Section \ref{sec:main_result}. We discuss the convergence of $T_R^\star(C)$ to $T^\star(C)$ as $R\to\infty$ in Section \ref{sec:convergence-of-TRstar-to-Tstar}, where we show that the convergence takes place in the special case when the TPM of each arm has identical rows. We conclude the paper in Section \ref{sec:concluding-remarks-and-discussion}. The proofs of all the results are contained in Appendices \ref{appndx:proof-of-lower-bound}-\ref{appndx:proof-of-convergence-of-T-R-star-to-T-star}.

\section{Notations and Preliminaries}
\label{sec:notations}

We consider a multi-armed bandit with $K\geq 2$ arms, and define $\mathcal{A}\coloneqq \{1,\ldots,K\}$ to be the set of arms. We associate with each arm an ergodic discrete-time Markov process on a common, finite state space $\mathcal{S}$. We assume that the Markov process of each arm is independent of those of the other arms. We write $\{X_t^a: t\geq 0\}$ denote the Markov process of arm $a\in \mathcal{A}$.\footnote{Throughout the paper, time $t\in \{0, 1, 2, \ldots\}.$} The state evolution on each arm is governed by its transition probability matrix (TPM). Given TPMs $P_1, \ldots, P_K$ and a permutation $\sigma:\{1, \ldots, K\} \to \{1, \ldots, K\}$, let $C=(P_{\sigma(1)}, \ldots, P_{\sigma(K)})$ denote an assignment of the TPMs to the arms in which the TPM assigned to arm $a$ is $P_{\sigma(a)}$. In the sequel, we refer to $C$ as an {\em assignment of the TPMs}, and we let $\mathcal{C}$ denote the collection of all possible assignments of the TPMs, i.e.,
\begin{equation}
\mathcal{C} = \{(P_{\sigma(1)}, \ldots, P_{\sigma(K)}):\ \sigma\text{ is a permutation on }\mathcal{A}\}.
\label{eq:all_arms_configurations}
\end{equation}  
For each $k = 1,\ldots, K$, let $\mu_k=\{\mu_k(i):\ i\in \mathcal{S}\}$ denote the unique stationary distribution of the TPM $P_k$. Given a function $f:\mathcal{S} \to \mathbb{R}$, let 
\begin{equation}
\nu_a \coloneqq \sum_{i \in \mathcal{S}}\ f(i)\, \mu_a(i), \quad a \in \mathcal{A},
\label{eq:nu_a}
\end{equation}
denote the average value of $f$ under $\mu_a$. Define the {\em best arm} $a^\star\in \mathcal{A}$ as 
\begin{equation}
a^\star \coloneqq \arg \max _{a\in \mathcal{A}}\ \nu_a = \arg \max _{a\in \mathcal{A}}\ \sum_{i \in \mathcal{S}}\ f(i)\, \mu_a(i).
\label{eq:best_arm}
\end{equation} 
We assume throughout the paper that $a^\star$ is unique. Without loss of generality, let $P_{a^\star} = P_1.$\footnote{The reader may recognise that this does not necessarily imply $a^\star =1$.}

For an integer $d\geq 1$ and a matrix $P$, let $P^{d}$ denote the matrix obtained by multiplying $P$ with itself $d$ times. For $i,j\in\mathcal{S}$ and $d\geq 1$, let $P^{d}(j|i)$ to denote the $(i,j)$th element of $P^{d}$. For $a\in \mathcal{A}$, let each row of $P_a$ be mutually absolutely continuous with the corresponding row of $P_{a'}$ for all $a'\neq a$. It is easy to see that this implies that for all $d\geq 1$, each row of $P_a^d$ is mutually absolutely continuous with the corresponding row of $P_{a'}^d$. The above assumption implies that the decision entity cannot infer the best arm merely by observing certain specific state(s) or state-transition(s) on the arm.

A decision entity that knows $P_1, \ldots, P_K$ only up to a permutation wishes to find the index of the best arm (i.e., the arm whose TPM is $P_1$) as quickly as possible, subject to an upper bound on the error probability. Clearly, this is accomplished if the decision entity finds $C\in \mathcal{C}$ that defines the problem instance. Given $C\in \mathcal{C}$, we write $\textsf{Alt}(C)$ to denote the set of all assignments of the TPMs alternative to $C$, i.e., those assignments of the TPMs in which the location of the best arm is different from the one in $C$.
In order to find the index of the best arm in a problem instance $C$, the decision entity selects the arms sequentially, one at each time $t$.
Let $A_t$ be the arm selected at time $t$. The decision entity observes the state of the arm $A_{t}$, denoted by $\bar{X}_{t}$. In contrast to the previous works \cite{garivier2016optimal,Kaufmann2016,moulos2019optimal} that deal with {\em i.i.d.} observations from the arms and rested arms, we assume that the unobserved arms continue to undergo state evolution whether or not they are selected ({\em restless} arms). Let $(A_{0:t},\bar{X}_{0:t}) \coloneqq (A_0,\bar{X}_0,\ldots,A_{t}\,\bar{X}_{t})$ denote the history of all the arm selections and observations seen up to time $t$. All random variables are defined on a common probability space $(\Omega, \mathcal{F}, \mathbb{P})$. Define the filtration  
\begin{equation}
\mathcal{F}_0\coloneqq \{\Omega, \emptyset\}, \quad \mathcal{F}_{t}\coloneqq \sigma(A_{0:t-1}, \bar{X}_{0:t-1}), \quad t\geq 1.
\label{eq:filtration}
\end{equation}

\subsection{Policy and Problem Definition}
A policy $\pi$ is defined by a collection of functions $\{\pi_{t}:t\geq 0\}$. At each time $t$, $\pi_{t}$ does one of the following based on the history $\mathcal{F}_t$:
\begin{itemize}
	\item stop and declare the index of the best arm;
	\item choose to pull arm $A_t$ according to a deterministic or a randomised rule.
\end{itemize}
Let $\pi$ denote a generic policy, and let $\tau(\pi)$ denote the stopping time of policy $\pi$ (defined with respect to the filtration \eqref{eq:filtration}). Let $\theta(\tau(\pi))$ denote the index of the best arm declared by the policy $\pi$ at the stopping time. 

Let $P_{C}^{\pi}(\cdot)$ denote the probability computed under the assignment of the TPMs $C$ and
under the policy $\pi$. 
For $a\in \mathcal{A}$, let $\mathcal{C}_a \subset \mathcal{C}$ denote the collection of all permutations in which $P_a=P_1$. Clearly, the collection $\{\mathcal{C}_a: \ a\in \mathcal{A}\}$ is a partition of $\mathcal{C}$. 
Given an error probability threshold $\epsilon>0$, let 
\begin{equation}
	\Pi(\epsilon)\coloneqq \{\pi: \ \text{for all }a\in \mathcal{A}, \ \ P_{C}^\pi(\theta(\tau(\pi))\neq a) \leq \epsilon \ \forall C\in \mathcal{C}_{a}\}
\label{eq:Pi(epsilon)}
\end{equation}
denote the collection of all policies whose error probability at the stopping time is no more than $\epsilon$ for all possible assignments of TPMs. 
We anticipate from similar results in the prior works \cite{garivier2016optimal,Kaufmann2016,moulos2019optimal} that $$\inf_{\pi\in\Pi(\epsilon)}\mathbb{E}_{C}^\pi[\tau(\pi)] = \Theta(\log(1/\epsilon)).$$ 
Here, $\mathbb{E}_{C}^\pi[\cdot]$ denotes the expectation under the assignment of the TPMs $C$ and policy $\pi$.
Our interest is in characterising, or at least bounding, the value of 
\begin{equation}
\lim _{\epsilon \downarrow 0} \inf_{\pi \in \Pi(\epsilon)}\ \frac{\mathbb{E}_{C}^\pi[\tau(\pi)]}{\log (1/\epsilon)}.
\label{eq:main_quantity}
\end{equation}
For simplicity, we assume that every policy starts by selecting arm $1$ at time $t=0$, arm $2$ at time $t=1$, etc., and arm $K$ at time $t=K-1$. This ensures that the Markov process of each arm is observed at least once. 

\section{Delays, Last Observed States, and a Markov Decision Problem}
\label{sec:delays-last-observed-states}
The contents of this section are mostly borrowed from \cite[Section II-B, II-C]{karthik2021detecting}, but modified appropriately to reflect the absence of the {\em trembling hand} assumption of \cite{karthik2021detecting}, with an intent to keep the material in the paper self-contained. Recall that the decision entity observes only one of the arms at each time $t$, while the unobserved arms continue to undergo state evolution. This means that at any time $t$, the probability of the observation $\bar{X}_t$ on the selected arm $A_t$ given the history $\mathcal{F}_t$ is a function of (a) the time elapsed since the previous time instant of selection of arm $A_t$ (called the \emph{delay} of arm $A_t$), and (b) the state of arm $A_t$ at its previous selection time instant (called the \emph{last observed state} of arm $A_t$). Notice that when the arms are \emph{rested}, the notion of arm delays is superfluous since each arm remains frozen at its previously observed state until its next selection time instant. Also, the notion of arm delays is redundant in the setting of iid observations because the current state of the arm selected is independent of the state at its previous selection. The notion of arm delays is a key distinguishing feature of the setting of restless arms.

For $t\geq K$, let $d_a(t)$ and $i_a(t)$ respectively denote the delay and the last observed state of arm $a$ at time $t$. Let $\underline{d}(t) \coloneqq (d_1(t),\ldots,d_K(t))$ and $\underline{i}(t) \coloneqq (i_1(t),\ldots,i_K(t))$ denote the vectors of arm delays and the last observed states at time $t$. Note that arm delays and last observed states are defined only for $t\geq K$ as these quantities are well-defined only when at least one observation is available from each arm. Set $\underline{d}(K)=(K,K-1,\ldots,1)$. Thus, $d_a(t)\geq 1$ for all $t\geq K$, and that $d_a(t)=1$ if and only if arm $a$ is selected at time $t-1$.

The rule for updating the arm delays and last observed states is as follows: if $A_{t}=a'$, then
\begingroup \allowdisplaybreaks\begin{align}
	{d}_a(t+1)=\begin{cases}
		d_a(t)+1, &a\neq a',\\
		1,& a=a',
	\end{cases} \qquad \qquad 
	i_a(t+1)=\begin{cases}
		i_a(t),& a\neq a',\\
		X_{t}^{a},& a=a',
	\end{cases}
	\label{eq:specific_transition_pattern}
\end{align}\endgroup
where $X_{t}^{a}=\bar{X}_{t}$ is the state of the arm $A_t=a'$ at time $t$.
Thus, for all $t \geq K$, based on $\{(\underline{d}(s), \underline{i}(s)): K\leq s\leq t\}$, the decision entity chooses to pull $A_{t}$ and observes $\bar{X}_{t}$.\footnote{Note that specifying $\{(\underline{d}(s), \underline{i}(s)): K\leq s\leq t\}$ is equivalent to specifying $(A_{0:t-1}, \bar{X}_{0:t-1})$ for all $t\geq K$.} It then form $(\underline{d}(t+1), \underline{i}(t+1)) $. This repeats until the stopping time, at which time it declares $\theta(\tau(\pi))$ (under policy $\pi$) as the candidate best arm.

From the update rule in \eqref{eq:specific_transition_pattern}, it is clear that the process $\{(\underline{d}(t),\underline{i}(t)):t\geq K\}$ takes values in a subset $\mathbb{S}$ of the {countable} set $\mathbb{N}^K\times\mathcal{S}^K$, where $\mathbb{N}=\{1,2,\ldots\}$ denotes the set of natural numbers. The subset $\mathbb{S}$ is formed based on the constraint that at any time $t\geq K$, exactly one of the components of $\underline{d}(t)$ is equal to $1$, and all the other components are $>1$. Given any assignment of the TPMs $C\in \mathcal{C}$ and policy $\pi$, note that for all $(\underline{d}',\underline{i}') \in \mathbb{S}$ and $t\geq K$,
\begin{align}
	&P_C^\pi(\underline{d}(t+1)=\underline{d}',\underline{i}(t+1)=\underline{i}'\mid \{(\underline{d}(s),\underline{i}(s)), \ K\leq s \leq t\},A_{0:t})\nonumber\\
	&=P_C^\pi(\underline{d}(t+1)=\underline{d}',\underline{i}(t+1)=\underline{i}'\mid (\underline{d}(t),\underline{i}(t)),A_t).\label{eq:controlled_markov_chain}
\end{align}
On account of \eqref{eq:controlled_markov_chain} being satisfied, we say that under any policy $\pi$, the evolution of the process $\{(\underline{d}(t),\underline{i}(t)):t\geq K\}$ is \emph{controlled} by the sequence $\{A_t\}_{t\geq 0}$ of intended arm selections under policy $\pi$. Alternatively, $\{(\underline{d}(t),\underline{i}(t)):t\geq K\}$ is a controlled Markov process, with $\{A_t\}_{t\geq 0}$ being the sequence of controls.\footnote{The terminology used here follows that of Borkar \cite{borkar1988control}.} Thus, we are in a Markov decision problem (MDP) setting. We now make precise the state space, the action space, the transition probabilities and our objective.

The state space of the MDP is $\mathbb{S}$, with the state at time $t$ denoted $(\underline{d}(t), \underline{i}(t))$. The action space of the MDP is $\mathcal{A}$, with action $A_t$ at time $t$ possibly depending on the history $\mathcal{F}_{t}$. The transition probabilities for the MDP under an assignment of the TPMs $C\in \mathcal{C}$ and a policy $\pi$ are given by
\begin{align}
	&P_C^\pi(\underline{d}(t+1)=\underline{d}',\underline{i}(t+1)=\underline{i}'\mid \underline{d}(t)=\underline{d},\underline{i}(t)=\underline{i}, A_t=a)\nonumber\\
	&=\begin{cases}
		(P_C^a)^{d_a}(i_a'|i_a),&\text{if }d_a'=1\text{ and }d'_{\tilde{a}}=d_{\tilde{a}}+1\text{ for all }\tilde{a}\neq a,\\
		&i_{\tilde{a}}'=i_{\tilde{a}}\text{ for all }\tilde{a}\neq a,\\
		0,&\text{otherwise},
	\end{cases}\label{eq:MDP_transition_probabilities}
\end{align}
where $P_C^a$ denotes the TPM of arm $a$ under the assignment of the TPMs $C$.
For instance, if $C=(P_1, \ldots, P_K)$, then $P_C^a=P_a$ for all $a\in \mathcal{A}$.
Note that the right hand side of \eqref{eq:MDP_transition_probabilities} is not a function of $t$ or $\pi$. Let $Q_{C}(\underline{d}', \underline{i}'|\underline{d}, \underline{i},a)$ denote the transition probabilities in \eqref{eq:MDP_transition_probabilities}.

\section{Log-Likelihoods and Log-Likelihood Ratios}
\label{sec:LLR}

Given $C\in\mathcal{C}$, let
\begin{align}
Z_{C}^{\pi}(n)\coloneqq\log P_C^{\pi}(A_{0:n},\Bar{X}_{0:n})
\label{eq:log-likelihood}
\end{align}
denote the log-likelihood of all the controls and observations seen up to time $n$ under the policy $\pi$ when $C$ is the assignment of the TPMs. Under the assumption that $\pi$ selects arm $1$ at time $t=0$, arm $2$ at time $t=1$, etc., and arm $K$ at time $K-1$, \eqref{eq:log-likelihood} may be expressed as
\begin{align}
Z_{C}^{\pi}(n) &= \sum_{a=1}^{K} \log P_{C}^{\pi}(X_{a-1}^{a}) \label{eq:Z_{C}_1} \\
&\hspace{1cm}+ \sum_{t=K}^{n} \ \log P_{C}^{\pi}(A_{t}|A_{0:t-1}, \, \bar{X}_{0:t-1})  \label{eq:Z_{C}_2}\\
&\hspace{2cm}+ \sum_{t=K}^{n} \ \log P_{C}^{\pi}(\bar{X}_{t}|A_{0:t}, \, \bar{X}_{0:t-1})  \label{eq:Z_{C}_3}.
\end{align}
Because $\pi$ is oblivious to the assignment of the TPMs $C$, \eqref{eq:Z_{C}_2} does not depend on $C$. Furthermore, \eqref{eq:Z_{C}_3} may be expressed as
\begin{align}
\sum_{t=K}^{n} \ \log P_{C}^{\pi}(\bar{X}_{t}|A_{0:t}, \, \bar{X}_{0:t-1}) &= \sum_{t=K}^{n} \ \sum_{a=1}^{K}\ \mathbb{I}_{\{A_{t}=a\}}\ \log P_{C}^{\pi}(\bar{X}_{t}|A_{0:t-1}, A_{t}=a, \bar{X}_{0:t-1})\nonumber\\
&=\sum_{t=K}^{n}  \ \sum_{(\underline{d}, \underline{i})\in \mathbb{S}}\ \sum_{a=1}^{K}\ \mathbb{I}_{\{\underline{d}(t)=\underline{d},\, \underline{i}(t)=\underline{i}, \, A_{t}=a\}}\, \log (P_{C}^{a})^{d_{a}}(\bar{X}_{t}|i_{a})\nonumber\\
&= \sum_{t=K}^{n}  \ \sum_{(\underline{d}, \underline{i})\in \mathbb{S}}\ \sum_{a=1}^{K}\ \sum_{j\in \mathcal{S}}\ \mathbb{I}_{\{\underline{d}(t)=\underline{d},\, \underline{i}(t)=\underline{i}, \, A_{t}=a, \, \bar{X}_{t}=j\}}\, \log (P_{C}^{a})^{d_{a}}(j|i_{a})\nonumber\\
&=\sum_{(\underline{d}, \underline{i})\in \mathbb{S}}\ \sum_{a=1}^{K}\ \sum_{j\in \mathcal{S}}\ N(n, \underline{d}, \underline{i}, a, j) \ \log (P_{C}^{a})^{d_{a}}(j|i_{a}),
\label{eq:log-likelihood-final}
\end{align}
where 
\begin{align}
N(n, \underline{d}, \underline{i}, a, j) \coloneqq  \sum_{t=K}^{n} \ \mathbb{I}_{\{\underline{d}(t)=\underline{d},\, \underline{i}(t)=\underline{i}, \, A_{t}=a, \, \bar{X}_{t}=j\}}
\label{eq:N(n,d,i,a,j)}
\end{align}
denotes the number of times up to time $n$ the process $\{(
\underline{d}(t), \underline{i}(t)):\, t\geq K\}$ is in the state $(\underline{d}, \underline{i})$, arm $a$ is selected subsequently, and the state of arm $a$ is observed to be $j$. Let 
\begin{align}
N(n, \underline{d}, \underline{i}, a) \coloneqq \sum_{j\in \mathcal{S}} \ N(n, \underline{d}, \underline{i}, a, j), \quad N(n, \underline{d}, \underline{i}) \coloneqq \sum_{a=1}^{K}\ \sum_{j\in \mathcal{S}}\ N(n, \underline{d}, \underline{i}, a, j).
\label{eq:N(n,d,i,a)-and-N(n,d,i)}
\end{align}

Given $C, C'\in \mathcal{C}$, let
\begin{align}
Z_{CC'}^{\pi}(n) \coloneqq Z_{C}^{\pi}(n) - Z_{C'}^{\pi}(n)=\log \frac{P_C^{\pi}(A_{0:n},\Bar{X}_{0:n})}{P_{C'}^{\pi}(A_{0:n},\Bar{X}_{0:n})}
\label{eq:LLR}
\end{align}
denote the log-likelihood ratio (LLR) of the controls and observations seen under the policy $\pi$ when the assignment of the TPMs is $C$ with respect to that when the assignment of the TPMs is $C'$. Using \eqref{eq:log-likelihood-final} in conjunction with the fact that~\eqref{eq:Z_{C}_2} does not depend on $C$, we get
\begin{align}
Z_{CC'}^{\pi}(n)=\sum_{a=1}^{K} \log \frac{P_{C}^{\pi}(X_{a-1}^{a})}{P_{C'}^{\pi}(X_{a-1}^{a})} + \sum_{(\underline{d}, \underline{i})\in \mathbb{S}}\ \sum_{a=1}^{K}\ \sum_{j\in \mathcal{S}}\ N(n, \underline{d}, \underline{i}, a, j) \ \log \frac{(P_{C}^{a})^{d_{a}}(j|i_{a})}{(P_{C'}^{a})^{d_{a}}(j|i_{a})}.
\label{eq:LLR-final}
\end{align}

\section{Lower Bound}
\label{sec:lower-bound}
We now present a lower bound for \eqref{eq:main_quantity}. Given two probability distributions $\mu$ and $\nu$ on the finite state space $\mathcal{S}$, the {\em Kullback--Leibler (KL) divergence} (also called the {\em relative entropy}) between $\mu$ and $\nu$ is defined as
\begin{equation}
	D_{\textsf{KL}}(\mu\|\nu)\coloneqq \sum_{i\in\mathcal{S}}\mu(i)\log \frac{\mu(i)}{\nu(i)},\label{eq:KL_divergence}
\end{equation}
where, by convention, $0\log \frac{0}{0}=0$.

\begin{proposition}
    \label{prop:lower_bound}
    Suppose $C\in \mathcal{C}$ is the underlying assignment of the TPMs. Then, 
\begin{equation}
	\liminf_{\epsilon\downarrow 0}\inf_{\pi\in\Pi(\epsilon)}\frac{\mathbb{E}_{C}^\pi[\tau(\pi)]}{\log(1/\epsilon)}\geq \frac{1}{T^{\star}(C)},\label{eq:lower_bound}
\end{equation}
where $T^{\star}(C)$ is given by
\begin{align}
T^{\star}(C)\coloneqq \sup_{\nu}\ \min_{C'\in \textsf{Alt}(C)} \ \sum_{(\underline{d},\underline{i})\in\mathbb{S}}\ \sum_{a=1}^{K}\  \nu(\underline{d},\underline{i},a) \ D_{\textsf{KL}}((P_C^a)^{d_a}(\cdot|i_a)\|(P_{C'}^a)^{d_a}(\cdot|i_a)).
\label{eq:T_a^{star}}
\end{align}
In \eqref{eq:T_a^{star}}, the supremum is over all $\nu=\{\nu(\underline{d}, \underline{i}, a): (\underline{d}, \underline{i})\in \mathbb{S}, a\in \mathcal{A}\}$ satisfying 
\begin{align}
	&\sum_{(\underline{d},\underline{i})\in\mathbb{S}}~\sum_{a=1}^{K}\,\nu(\underline{d},\underline{i},a)=1,\label{eq:lower_bound_6_2}\\
	&\nu(\underline{d},\underline{i},a)\geq 0\quad \text{for all }(\underline{d},\underline{i},a)\in\mathbb{S}\times\mathcal{A}.\label{eq:lower_bound_6_3}
\end{align}
\end{proposition}
\begin{IEEEproof}
The proof proceeds in several steps. First, we derive an analogue of the ubiquitous change-of-measure result \cite[Lemma 18]{Kaufmann2016} for the setting of restless arms. We then lower bound the expected LLR of any policy whose stoppage error probability is at most $\epsilon$ by $d(\epsilon, 1-\epsilon)$, where $d(x, y)$ denotes the relative entropy between two Bernoulli distributions with parameters $x$ and $y$. Next, we derive an upper bound on the expected LLR in terms of the expected stopping time. This involves deriving an analogue of Wald's identity for the setting of restless arms. Combining the upper and the lower bounds for the expected LLR, we get \eqref{eq:lower_bound}. The details are in Appendix \ref{appndx:proof-of-lower-bound}.
\end{IEEEproof}

\subsection{A Flow Constraint}
Notice that $T^{\star}(C)$ is the optimal value of an infinite-dimensional linear program (LP). The question of whether there exists $\nu$ attaining the supremum in \eqref{eq:T_a^{star}} remains open. The key difficulty in showing this is that because the set $\mathbb{S}$ is countably infinite, it is not clear whether the set of all $\nu$ satisfying \eqref{eq:lower_bound_6_2}-\eqref{eq:lower_bound_6_3} (which is akin to the space of all probability distributions on $\mathbb{S}\times \mathcal{A}$) is compact. Also, because this supremum is over {\em all} probability distributions on $\mathbb{S}\times \mathcal{A}$ which is a large class of distributions,  the lower bound in~\eqref{eq:lower_bound}   may not be achievable. It seems necessary to introduce additional constraints on $\nu$ to render the lower bound achievable. Indeed, given $\delta>0$, suppose $\nu_{\delta}$ is a probability distribution on $\mathbb{S}\times \mathcal{A}$ such that $$ \min_{C'\in \textsf{Alt}(C)} \ \sum_{(\underline{d},\underline{i})\in\mathbb{S}}\ \sum_{a=1}^{K}\  \nu_{\delta}(\underline{d},\underline{i},a) \ D_{\textsf{KL}}((P_C^a)^{d_a}(\cdot|i_a)\|(P_{C'}^a)^{d_a}(\cdot|i_a)) \geq T^{\star}(C)-\delta. $$ One way to achieve the quantity on the left hand side of the above equation is to ensure that for all $(\underline{d}, \underline{i}, a)\in \mathbb{S}\times \mathcal{A}$, the value of the fraction $N(n, \underline{d}, \underline{i}, a)/n$ is close to $\nu_{\delta}(\underline{d}, \underline{i}, a)$ for all $n$ large (the regime of large $n$ is akin to the regime of vanishing error probabilities). It seems difficult to accomplish this in the absence of more structure on $\nu_{\delta}$. It is worth noting here that in a related problem of odd arm identification, the authors of \cite{karthik2021detecting} are confronted with a similar difficulty in showing the achievability of the lower bound therein. To ameliorate the difficulty, they introduce a version of the following {\em flow} constraint on $\nu$:
\begin{equation}
\textsf{flow constraint}:\quad \sum_{a=1}^{K}\nu(\underline{d}', \underline{i}', a) = \sum_{(\underline{d}, \underline{i})\in \mathbb{S}}\ \sum_{a=1}^{K}\ \nu(\underline{d}, \underline{i}, a)\ Q_{C}(\underline{d}', \underline{i}'|\underline{d}, \underline{i}, a)\quad \text{for all }(\underline{d}', \underline{i}')\in \mathbb{S}.
\label{eq:flow-constraint}
\end{equation}
In \eqref{eq:flow-constraint}, $Q_{C}$ is the MDP transition matrix defined in \eqref{eq:MDP_transition_probabilities}. The flow constraint is, in fact, a {\em global balance} equation, and dictates that for any $(\underline{d}', \underline{i}')\in \mathbb{S}$, the long-term probability of a transition from $(\underline{d}', \underline{i}')$ (the {\em flow} out of $(\underline{d}', \underline{i}')$, captured by the left hand side of \eqref{eq:flow-constraint}) is equal to the probability of a transition to $(\underline{d}', \underline{i}')$ (the {\em flow} into $(\underline{d}', \underline{i}')$, captured by the right hand side of \eqref{eq:flow-constraint}). The authors of \cite{karthik2021detecting} show that their lower bound, after including the flow constraint, can be achieved by a certain {\em trembling hand}-based policy; see \cite[Section V]{karthik2021detecting} for a description of the policy. However, the policy in \cite{karthik2021detecting} is not practically implementable. 

With an end goal of showing achievability of our lower bound, we take \eqref{eq:flow-constraint} into consideration along with \eqref{eq:lower_bound_6_2}-\eqref{eq:lower_bound_6_3} when evaluating the supremum in \eqref{eq:T_a^{star}}, and wish to design a policy that (a) is  computationally  feasible/tractable and easy-to-implement, and (b) achieves the lower bound in \eqref{eq:lower_bound}. This forms the content of the next section.


\section{Achievability}
\label{sec:achievability}

As alluded to in the previous section, the {\em trembling hand}-based policy of \cite{karthik2021detecting} is not practically implementable. One reason for this is that the arm delays (which appear in the policy of \cite{karthik2021detecting}) take countably infinitely many values and therefore cannot be handled on a machine with  finite-size memory. To alleviate the difficulty arising from the countably infinite-valued arm delays, we study a simplified setting where the maximum delay of each arm is restricted to be at most $R$ for some $R\in \mathbb{N}\cap (K, \infty),$\footnote{We consider $R>K$ to be consistent with our assumption that each of the arms is sampled once in the first $K$ time instants.} and an arm whose delay is equal to $R$ at any given time is forcibly selected in the following time instant. Let $\mathbb{S}_{R}$ denote the subset of $\mathbb{S}$ in which the delay of each arm is no more than $R$. Further, for $a\in \mathcal{A}$, let $\mathbb{S}_{R,a}$ denote the subset of $\mathbb{S}_{R}$ in which the delay of arm $a$ is equal to $R$. Notice that $\mathbb{S}_{R,a}$ is a finite set for each $a$ and that $\mathbb{S}_{R,a}$ and $\mathbb{S}_{R,a'}$ are disjoint for all $a'\neq a$.

%

\subsection{Modifications to the MDP Transition Probabilities Under Maximum Delay Constraint}

Recall that $\{(\underline{d}(t), \underline{i}(t)): t\geq K\}$ and $\{A_{t}: t\geq 0\}$ together define a Markov decision problem (MDP) whose state space is $\mathbb{S}$, the action space is $\mathcal{A}$, the state at time $t$ is $(\underline{d}(t), \underline{i}(t))$, and the control (or action) at time $t$ is $A_{t}$. From Section \ref{sec:delays-last-observed-states}, we know that the transition probabilities of the MDP are given by \eqref{eq:MDP_transition_probabilities}. When the delay of each arm is constrained to be no more than $R$, the modified state space of the MDP is $\mathbb{S}_{R}$, and the modified transition probabilities for the MDP are as follows:
\begin{itemize}
    \item \textit{Case 1:} $(\underline{d},\underline{i})\notin \bigcup_{a=1}^{K}\ \mathbb{S}_{R,a}$. In this case, the transition probabilities are as in \eqref{eq:MDP_transition_probabilities}.
\item \textit{Case 2:} $(\underline{d},\underline{i})\in \mathbb{S}_{R,a}$ for some $a\in \mathcal{A}$. In this case, when $A_{t}=a$,
    \begin{align}
	&P_C^\pi(\underline{d}(t+1)=\underline{d}',\underline{i}(t+1)=\underline{i}'\mid \underline{d}(t)=\underline{d},\underline{i}(t)=\underline{i}, A_t=a)\nonumber\\
	&=\begin{cases}
		(P_C^a)^{R}(i_a'|i_a),&\text{if }d_a'=1\text{ and }d'_{\tilde{a}}=d_{\tilde{a}}+1\text{ for all }\tilde{a}\neq a,\\
		&i_{\tilde{a}}'=i_{\tilde{a}}\text{ for all }\tilde{a}\neq a,\\
		0,&\text{otherwise},
	\end{cases}
\label{eq:modified_MDP_transition_probabilities2}
\end{align}
and when $A_{t} \neq a$, the transition probabilities are undefined. 
\end{itemize}
We write $Q_{C,R}(\underline{d}', \underline{i}'|\underline{d}, \underline{i},a)$ to denote the transition probabilities in \eqref{eq:modified_MDP_transition_probabilities2}.

\subsection{Capturing the Maximum Delay Constraint and a Finite Dimensional Linear Program}

Recall that in the absence of any constraints on the maximum delay of each arm, the lower bound is as in \eqref{eq:lower_bound}, with the constant $T^{\star}(C)$ in \eqref{eq:lower_bound} as given in \eqref{eq:T_a^{star}}. Further, the supremum in \eqref{eq:T_a^{star}} is over all $\nu$ satisfying \eqref{eq:lower_bound_6_2}-\eqref{eq:flow-constraint}. When the delay of each arm is constrained to be no more than $R$, the following additional constraint on $\nu$ comes into play:
\begin{equation}
R\textsf{-max-delay-constraint}: \quad \nu(\underline{d}, \underline{i}, a)=\sum_{a'=1}^{K}\ \nu(\underline{d}, \underline{i}, a')\quad \text{for all }(\underline{d}, \underline{i})\in \mathbb{S}_{R,a}.
\label{eq:additional-condition-with-delay-restriction}
\end{equation}
The condition in \eqref{eq:additional-condition-with-delay-restriction} captures the observation that any occurrence of the state $(\underline{d}, \underline{i})\in \mathbb{S}_{R,a}$ is followed by selecting arm $a$ forcibly (i.e., with probability $1$), thus implying that
$\nu(\underline{d}, \underline{i}, a')=0$ for all $a'\neq a$ which is equivalent to~\eqref{eq:additional-condition-with-delay-restriction}. 
Let $T_{R}^{\star}(C)$ be the optimal value of the following optimisation problem: 
\begin{align}
&\sup_{\nu}\ \min_{C'\in \textsf{Alt}(C)}\ \sum_{(\underline{d}, \underline{i})\in \mathbb{S}_{R}}\ \sum_{a=1}^{K}\ \nu(\underline{d}, \underline{i}, a)\ D_{\textsf{KL}}((P_{C}^{a})^{d_{a}}(\cdot | i_{a}) \| (P_{C'}^{a})^{d_{a}}(\cdot | i_{a})),
\label{eq:T-R-star-C}\\
&\text{subject to}\nonumber\\
&\sum_{a=1}^{K}\nu(\underline{d}',\underline{i}',a)=\sum_{(\underline{d},\underline{i})\in\mathbb{S}_{R}}
\ \sum_{a=1}^{K}\,\nu(\underline{d},\underline{i},a)\ Q_{C, R}(\underline{d}',\underline{i}'|\underline{d},\underline{i},a)\quad \text{for all }(\underline{d}',\underline{i}')\in\mathbb{S}_{R},\label{eq:lower_bound_6_1_2}\\
	&\sum_{(\underline{d},\underline{i})\in\mathbb{S}_{R}}\ \sum_{a=1}^{K}\ \nu(\underline{d},\underline{i},a)=1,\label{eq:lower_bound_6_2_2}\\
	&\nu(\underline{d},\underline{i},a) \geq 0\quad \text{for all }(\underline{d},\underline{i},a)\in\mathbb{S}_{R}\times\mathcal{A},\label{eq:lower_bound_6_3_3}\\
	&\nu(\underline{d},\underline{i},a)=\sum_{a'=1}^{K}\ \nu(\underline{d}, \underline{i}, a')\quad \text{for all }(\underline{d}, \underline{i})\in \mathbb{S}_{R,a}, \quad a\in \mathcal{A}. \label{eq:lower_bound_6_3_4}
\end{align}
Notice that the above optimisation problem is a finite-dimensional LP, and is the analogue of the infinite-dimensional LP in \eqref{eq:T_a^{star}} with the additional condition in \eqref{eq:lower_bound_6_3_4} to account for the case when an arm is forcibly selected if its delay equals $R$. Because (a) $\mathbb{S}_{R}\times \mathcal{A}$ is finite, (b) the space of all probability distributions on $\mathbb{S}_{R}\times \mathcal{A}$ (say, $\mathscr{P}(\mathbb{S}_{R}\times \mathcal{A})$) is compact with respect to the topology arising from the Euclidean metric in $\mathbb{R}^{|\mathbb{S}_{R}\times \mathcal{A}|}$, (c) the set of all $\nu$ satisfying \eqref{eq:lower_bound_6_1_2}-\eqref{eq:lower_bound_6_3_4} is a closed subset of $\mathscr{P}(\mathbb{S}_{R}\times \mathcal{A})$ (and therefore compact), and (d) the expression $$\min_{C'\in \textsf{Alt}(C)}\ \sum_{(\underline{d}, \underline{i})\in \mathbb{S}_{R}}\ \sum_{a=1}^{K}\ \nu(\underline{d}, \underline{i}, a)\ D_{\textsf{KL}}((P_{C}^{a})^{d_{a}}(\cdot | i_{a}) \| (P_{C'}^{a})^{d_{a}}(\cdot | i_{a}))$$ is continuous in $\nu$, it follows by Weierstrass extreme value theorem that there exists $\nu_{C,R}^{\star}=\{\nu_{C,R}^{\star}(\underline{d}, \underline{i}, a): (\underline{d}, \underline{i}, a)\in \mathbb{S}_{R}\times \mathcal{A}\}$ that attains the supremum in \eqref{eq:T-R-star-C}. Although a closed-form expression for $\nu_{C,R}^\star$ is not currently available, it can easily be evaluated numerically. In the next section, we fix $R\in \mathbb{N}\cap (K, \infty)$ and design a policy for finding the best arm that samples an arm forcibly if its delay is equal to $R$. Additionally, we demonstrate that our policy (a) stops in finite time almost surely, (b) satisfies the desired error probability, and (c) achieves an upper bound of $1/T_{R}^{\star}(C)$ asymptotically as the error probability vanishes. We shall see that our policy is easy to implement as it operates on the finite set $\mathbb{S}_{R}$ instead of the countable set $\mathbb{S}$.

\subsection{A Policy for Finding the Best Arm Under a Maximum Delay Constraint}
\label{subsec:policy-description}
Fix an $R\in \mathbb{N}\cap (K, \infty)$. In this section, we design a policy for finding the best arm when the delay of each arm is constrained to be no more than $R$. Towards this, we first analyse a uniform arm selection policy that, for all $t\geq K$, selects the arms uniformly whenever the delay of each arm is $<R$, and forcibly selects an arm whose delay is equal to $R$. Let this policy be denoted $\pi_{R}^{\textsf{unif}}$. It is clear that $\{(\underline{d}(t), \underline{i}(t)): t\geq K\}$ is a Markov process under $\pi_{R}^{\textsf{unif}}$ with $\mathbb{S}_{R}$ as its state space. The following result shows that this Markov process is, in fact, ergodic.
\begin{lemma}
\label{lemma:ergodicity-under-unif-SRS-policy}
Fix $R\in \mathbb{N}\cap (K, \infty)$. Under every assignment of the TPMs $C\in \mathcal{C}$, the process $\{(\underline{d}(t), \underline{i}(t)):\ t\geq K\}$ is ergodic under the policy $\pi_{R}^{\textsf{unif}}$.
\end{lemma}
\begin{IEEEproof}
See  Appendix \ref{appndx:proof-of-lemma-ergodicity-under-unif-SRS-policy}.
\end{IEEEproof}
As a consequence of Lemma \ref{lemma:ergodicity-under-unif-SRS-policy}, the process $\{(\underline{d}(t), \underline{i}(t)):\ t\geq K\}$ has a unique stationary distribution, say $\mu_{C,R}^{\textsf{unif}}=\{\mu_{C,R}^{\textsf{unif}}(\underline{d}, \underline{i}): (\underline{d}, \underline{i})\in \mathbb{S}_{R}\}$, under the policy $\pi_{R}^{\textsf{unif}}$ and under the assignment of the TPMs $C$. We note that $\mu_{C,R}^{\textsf{unif}}(\underline{d}, \underline{i})>0$ for all $(\underline{d}, \underline{i})\in \mathbb{S}_{R}$. Let
\begin{equation}
\nu_{C,R}^{\textsf{unif}}(\underline{d}, \underline{i}, a)\coloneqq 
\begin{cases}
\frac{\mu_{C,R}^{\textsf{unif}}(\underline{d}, \underline{i})}{K}, & (\underline{d}, \underline{i})\notin \bigcup_{a'=1}^{K}\ \mathbb{S}_{R,a'},\\
\mu_{C,R}^{\textsf{unif}}(\underline{d}, \underline{i}), & (\underline{d}, \underline{i})\in \mathbb{S}_{R,a},\\
0, &(\underline{d}, \underline{i}) \in \bigcup_{a'\neq a}\  \mathbb{S}_{R,a'}.
\end{cases}
\label{eq:ergodic-state-action-under-unif}
\end{equation}
denote the corresponding {\em ergodic} state-action occupancy measure. Observe that $\nu_{C,R}^{\textsf{unif}}$ satisfies \eqref{eq:lower_bound_6_1_2}-\eqref{eq:lower_bound_6_3_4}.

For $\eta\in (0,1]$ and $C\in \mathcal{C}$, let 
\begin{align}
\nu_{\eta,R,C}(\underline{d}, \underline{i}, a) &\coloneqq \eta\ \nu_{C,R}^{\textsf{unif}}(\underline{d},  \underline{i}, a) + (1-\eta) \ \nu_{C,R}^{\star}(\underline{d}, \underline{i}, a), \label{eq:nu-eta-R-C}\\
\mu_{\eta, R, C}(\underline{d}, \underline{i}) &\coloneqq \sum_{a=1}^{K}\ \nu_{\eta,R,C}(\underline{d}, \underline{i}, a).
\label{eq:mu-eta-R-C}
\end{align}
Observe that $\mu_{\eta, R, C}(\underline{d}, \underline{i})\geq \frac{\eta}{K}\ \mu_{C,R}^{\textsf{unif}}(\underline{d}, \underline{i})>0$ for all $(\underline{d}, \underline{i})\in \mathbb{S}_{R}$. Also, $\nu_{\eta,R,C}$ satisfies \eqref{eq:lower_bound_6_1_2}-\eqref{eq:lower_bound_6_3_4} by virtue of the fact that both $\nu_{C,R}^{\textsf{unif}}$ and $\nu_{C,R}^{\star}$ satisfy \eqref{eq:lower_bound_6_1_2}-\eqref{eq:lower_bound_6_3_4}. Let $\lambda_{\eta,R,C}=\{\lambda_{\eta,R,C}(a|\underline{d}, \underline{i}): a\in \mathcal{A}, \ (\underline{d},\underline{i})\in \mathbb{S}_{R}\}$ be defined as
\begin{equation}
\lambda_{\eta,R,C}(a|\underline{d}, \underline{i}) = \frac{\nu_{\eta,R,C}(\underline{d}, \underline{i}, a)}{\mu_{\eta, R, C}(\underline{d}, \underline{i})} \quad (\underline{d}, \underline{i})\in \mathbb{S}_{R}, \ a\in \mathcal{A}.
\label{eq:lambda-eta-R-C}
\end{equation}
Notice that for all $(\underline{d}, \underline{i})\notin \bigcup_{a'=1}^{K}\ \mathbb{S}_{R,a'}$,
\begin{align}
\lambda_{\eta,R,C}(a|\underline{d}, \underline{i}) \geq \frac{\eta}{K}\ \mu_{C,R}^{\textsf{unif}}(\underline{d}, \underline{i}) 
&\geq \frac{\eta}{K}\ \mu_{R}^{\textsf{min}},
\end{align}
where 
\begin{equation}
 \mu_{R}^{\textsf{min}} \coloneqq \min_{C\in \mathcal{C}}\ \min_{(\underline{d}, \underline{i})\in \mathbb{S}_{R}} \mu_{C, R}^{\textsf{unif}}(\underline{d}, \underline{i})>0. 
\label{eq:mu-min}
\end{equation}
That is, whenever the delay of each arm is $<R$, the distribution $\lambda_{\eta, R, C}(\cdot|\underline{d}, \underline{i})$ puts a strictly positive mass on each of the arms.
Using the arm selection rule in \eqref{eq:lambda-eta-R-C} instead of the uniform sampling rule and following the proof template in Appendix \ref{appndx:proof-of-lemma-ergodicity-under-unif-SRS-policy}, it can be shown that the policy $\pi^{\lambda_{\eta,R,C}}$, which selects the arms at each time instant according to the rule in \eqref{eq:lambda-eta-R-C}, renders the process $\{(\underline{d}(t), \underline{i}(t)): t\geq K\}$ ergodic. We now claim that the stationary distribution of the process $\{(\underline{d}(t), \underline{i}(t)): t\geq K\}$ under $\pi^{\lambda_{\eta,R,C}}$ is $\mu_{\eta, R, C}$. Indeed, suppose $Q^{C,R}=\{Q^{C,R}(\underline{d}', \underline{i}'|\underline{d}, \underline{i}): (\underline{d}', \underline{i}'), (\underline{d}, \underline{i})\in \mathbb{S}_{R}\}$ denotes the transition probability matrix of the Markov process  $\{(\underline{d}(t), \underline{i}(t)): t\geq K\}$ under the policy $\pi^{\lambda_{\eta,R,C}}$. Then,
\begin{align}
Q^{C,R}(\underline{d}', \underline{i}'|\underline{d}, \underline{i}) &= \sum_{a=1}^{K}\ \lambda_{\eta, R, C}(a|\underline{d}, \underline{i})\ Q_{C,R}(\underline{d}', \underline{i}'|\underline{d}, \underline{i}, a),
\label{eq:TPM-of-ergodic-d(t)-i(t)-process}
\end{align}
from which it follows that for all $(\underline{d}', \underline{i}')\in \mathbb{S}_{R}$,
\begin{align}
\sum_{(\underline{d}, \underline{i})\in \mathbb{S}_{R}}\ \mu_{\eta, R, C}(\underline{d}, \underline{i})\ Q^{C,R}(\underline{d}', \underline{i}'|\underline{d}, \underline{i})&= \sum_{(\underline{d}, \underline{i})\in \mathbb{S}_{R}}\ \sum_{a=1}^{K}\ \mu_{\eta, R, C}(\underline{d}, \underline{i})\ \lambda_{\eta, R, C}(a|\underline{d}, \underline{i})\ Q_{C,R}(\underline{d}', \underline{i}'|\underline{d}, \underline{i}, a)\nonumber\\
&=\sum_{(\underline{d}, \underline{i})\in \mathbb{S}_{R}}\ \sum_{a=1}^{K}\  \nu_{\eta, R, C}(\underline{d}, \underline{i}, a)\ Q_{C,R}(\underline{d}', \underline{i}'|\underline{d}, \underline{i}, a) \nonumber\\
&\stackrel{(a)}{=}\sum_{a=1}^{K}\ \nu_{\eta, R, C}(\underline{d}', \underline{i}', a)\nonumber\\
&=\mu_{\eta, R, C}(\underline{d}', \underline{i}'), 
\end{align}
thus proving the claim. In the above set of equations, $(a)$ follows from the fact that $\nu_{\eta, R, C}$ satisfies \eqref{eq:lower_bound_6_1_2}.

\begin{remark}
Observe that $\nu_{\eta,R,C}$ in \eqref{eq:nu-eta-R-C} is a {\em mixture} of two terms, one arising from the uniform arm selection rule under maximum delay constraints (the term corresponding to $\nu_{C,R}^{\textsf{unif}}$), and the other arising from the optimal solution to the finite-dimensional LP under a constraint on the delay of each arm (the term corresponding to $\nu_{C,R}^{\star}$). A similar, {\em trembling hand}-based mixture term appears in the works \cite{karthik2021detecting,karthik2021learning}. While the mixtures in  \cite{karthik2021detecting,karthik2021learning} arise from a restrictive system model that forces each arm to be selected with a strictly positive probability, the mixture in \eqref{eq:nu-eta-R-C} ensures that each arm is selected with a strictly positive probability (except when the delay of an arm is equal to the maximum allowed delay $R$ in which case it is forcibly selected) without any restrictions on the system model. It is also worth noting that the mixtures in \cite{karthik2021detecting,karthik2021learning} give rise to a conditional probability distribution on the arms, conditioned on the arm delays and the last observed states (a quantity akin to $\lambda(a|\underline{d}, \underline{i}$)), whereas the mixture in \eqref{eq:nu-eta-R-C} gives rise to a joint probability distribution on the set $\mathbb{S}_{R}\times \mathcal{A}$ (a quantity akin to $\nu(\underline{d}, \underline{i}, a)$). 
\end{remark}

Our policy, which we call {\em $R$-Delay-Constrained-Restless-BAI} (or {\sc $R$-DCR-BAI} in short) or alternatively as $\pi^{\star}(L, \eta, R)$, is then as follows. Here, $L>1$, $\eta\in (0,1]$ and $R\in \mathbb{N}\cap (K, \infty)$ are parameters of the policy. As we shall see, the parameter $L$ controls the policy's error probability. In fact, we will see from Lemma \ref{lem:pi_star(L)_in_Pi(epsilon)} that if we set $L=1/\epsilon$, the error probability is bounded above by $\epsilon$.
\vspace{.2in}
\hrule
\vspace{.1in}
\noindent \textbf{\underline{\emph{Policy } {\sc $R$-DCR-BAI} / $\pi^{\star}(L, \eta, R)$}}:\\
\noindent Fix $L>1$, $\eta\in (0,1]$, and $R\in \mathbb{N}\cap (K, \infty)$. Assume that $A_0=1$, $A_1=2$, \ldots, $A_{K-1}=K$.  Let $$M_{C}^{\pi^{\star}(L, \eta, R)}(n)=\min_{C'\in \textsf{Alt}(C)}Z_{CC'}^{\pi^{\star}(L, \eta, R)}(n), \quad C\in \mathcal{C}.$$ Implement the following steps for all $n\geq K$.\\
\noindent (1) Compute $\bar{C}(n)\in \arg\max_{C\in \mathcal{C}}\ M_C^{\pi^{\star}(L, \eta, R)}(n)$. Resolve ties uniformly at random.\\
\noindent (2) If $M^{\pi^{\star}(L, \eta, R)}_{\bar{C}(n)}(n)\geq \log(L(K-1)(K-1)!)$, stop and declare the index of the best arm in $\bar{C}(n)$.\\
\noindent (3) If $M^{\pi^{\star}(L, \eta, R)}_{\bar{C}(n)}(n)<\log(L(K-1)(K-1)!)$, select arm $A_{n}$ according to the distribution $$ P^{\pi^{\star}(L, \eta, R)} (A_{n}=a|A_{0:n-1}, \bar{X}_{0:n-1}) = \lambda_{\eta,R, \bar{C}(n)}(a|\underline{d}(n), \underline{i}(n)).$$ 
\hrule
\vspace{.2in}
In item (1) in {\sc $R$-DCR-BAI}, $\bar{C}(n)$ denotes the estimate of the underlying assignment of the TPMs based on all the controls (arm selections) and observations seen up to time $n$. If the LLR between $\bar{C}(n)$ and its nearest alternative assignment of the TPMs exceeds a certain threshold (i.e.,  $\geq \log (L(K-1)(K-1)!)$), then the policy is sufficiently confident that $\bar{C}(n)$ is indeed the underlying assignment of the TPMs, and therefore stops and declares the index of the best arm in $\bar{C}(n)$. Else, it samples the next arm based on the value of $(\underline{d}(n), \underline{i}(n))$ according to the distribution $\lambda_{\eta,R, \bar{C}(n)}(\cdot|\underline{d}(n), \underline{i}(n))$.

In the following section, we demonstrate that for a suitable choice of $L$, the policy {\sc $R$-DCR-BAI} achieves the desired error probability. Further, letting $L\to\infty$, we show that the growth rate of its expected stopping time satisfies an asymptotic upper bound that is arbitrarily close to $1/T_{R}^{\star}(C)$ under the assignment of the TPMs $C$ for a suitable choice of $\eta$.

\section{Results on the Performance of Policy \texorpdfstring{\sc $R$-DCR-BAI}{R DCR BAI}}

\label{sec:performance-results}
This section is organised as follows. In Section 
\ref{subsection:strictly-positive-drift-llr}, we establish that for a given $C\in \mathcal{C}$ and any 
$C'\in \textsf{Alt}(C)$, the LLR $Z_{CC'}^{\pi}(n)$ has a strictly positive drift almost surely under the assignment of the TPMs $C$ (Lemma \ref{lem:liminf_Z_{CC'}(n)_strictly_positive}), and therefore the policy {\sc $R$-DCR-BAI} stops in finite time almost surely. In Section \ref{subsec:desired-error-probability}, we show that {\sc $R$-DCR-BAI} satisfies any desired error probability for a suitable choice of $L$ (Lemma \ref{lem:pi_star(L)_in_Pi(epsilon)}). In Section \ref{subsec:correct-asymptotic-drift-for-LLRs}, we strengthen the result of Section 
\ref{subsection:strictly-positive-drift-llr} by showing that the LLR of $C$ with respect to its nearest alternative $C'\in \textsf{Alt}(C)$ is a certain constant that, in the limit as $\eta\downarrow 0$, converges to $T_{R}^{\star}(C)$ (Lemma \ref{prop:Z_{CC'}(n)_has_the_right_drift}). In Section \ref{subsec:asymptotic-growth-stopping-time}, we show that the stopping time of {\sc $R$-DCR-BAI} grows almost surely as $L\to\infty$ (Lemma \ref{Lemma:stopping_time_of_policy_goes_to_infinity}). In Section \ref{subsec:almost-sure-upper-bound-on-stopping-time}, we derive an almost sure upper bound on the stopping time of {\sc $R$-DCR-BAI} that, in the limit as $\eta\downarrow 0$, converges to $1/T_{R}^{\star}(C)$. In Section \ref{subsec:asymptotic-upper-bound-on-expected-stopping-time}, we show that the expected stopping time of {\sc $R$-DCR-BAI} satisfies the same upper bound as that derived in Section \ref{subsec:almost-sure-upper-bound-on-stopping-time}. This is based on (a) showing that the family $\{\tau(\pi^{\star}(L, \eta, R))/\log L:L>1\}$ is uniformly integrable, and (b) combining the almost sure upper bound with the uniform integrability result to obtain an upper bound in expectation. The proofs of all the results are relegated to the appendices.

\subsection{Strictly Positive Drift for the LLRs}
\label{subsection:strictly-positive-drift-llr}
 
Let $\pi_{\textsf{NS}}^{\star}(L,\eta,R)$ denote a version of {\sc $R$-DCR-BAI} that never stops, i.e., it does not check the second step in {\sc $R$-DCR-BAI} and continues to the last step indefinitely. We now show that under $\pi_{\textsf{NS}}^{\star}(L,\eta,R)$, the LLRs have a strictly positive drift as the number of rounds of arm selection $n\to\infty$.
\begin{lemma}
\label{lem:liminf_Z_{CC'}(n)_strictly_positive}
Fix $L>1$, $\eta\in (0,1]$, $R\in \mathbb{N}\cap(K, \infty)$, and $C\in \mathcal{C}$. Under the assignment of the TPMs $C$ and the policy $\pi=\pi_{\textsf{NS}}^{\star}(L,\eta,R)$,
	\begin{equation}
		\liminf_{n\to\infty}\frac{Z_{CC'}^{\pi}(n)}{n}>0\quad \text{almost surely} \quad \text{for all }C'\in \textsf{Alt}(C).\label{eq:liminf_Z_{hh'}(n)_strictly_positive}
	\end{equation}
\end{lemma}

\begin{IEEEproof}
See Appendix \ref{appndx:proof-of-lemma-liminf_Z_{CC'}(n)_strictly_positive}.
\end{IEEEproof}
Lemma \ref{lem:liminf_Z_{CC'}(n)_strictly_positive} asserts that under the assignment of the TPMs $C$, we have
\begin{equation}
\liminf_{n\to\infty} \frac{M_{C}^{\pi}(n)}{n}>0\quad \text{almost surely} 
\label{eq:liminf-M_C(pi)(n)/n-strictly-positive}
\end{equation}
when $\pi=\pi_{\textsf{NS}}^{\star}(L,\eta,R)$. This means that $M_{C}^{\pi}(n)\geq \log(L(K-1)(K-1)!)$ for all $n$ large, almost surely. This proves that {\sc $R$-DCR-BAI} stops in finite time with probability $1$.  

\subsection{Desired Error Probability}
\label{subsec:desired-error-probability}
In this section, we show that for an appropriate choice of the parameter $L$, the policy {\sc $R$-DCR-BAI} achieves any desired error probability.
\begin{lemma}
\label{lem:pi_star(L)_in_Pi(epsilon)}
Fix an error probability threshold $\epsilon>0$. If $L=1/\epsilon$, then $\pi^\star(L,\eta,R)\in\Pi(\epsilon)$ for all $\eta\in (0,1]$ and $R\in \mathbb{N}\cap (K, \infty)$. Here, $\Pi(\epsilon)$ is as defined in \eqref{eq:Pi(epsilon)}.
\end{lemma}
\begin{IEEEproof}
The proof uses the fact that the policy stops in finite time almost surely, and is given in Appendix \ref{appndx:proof-of-admissibility-of-policy}.
\end{IEEEproof}

\subsection{The Correct Asymptotic Drift of the LLRs}
\label{subsec:correct-asymptotic-drift-for-LLRs}
In this section, we strengthen the result of Section \ref{subsection:strictly-positive-drift-llr} by showing that under the constraint that the delay of each arm is at most $R$, the asymptotic drift of the LLRs is arbitrarily close to $T_{R}^{\star}(C)$.
\begin{proposition}
\label{prop:Z_{CC'}(n)_has_the_right_drift}
	Fix $L>1$, $\eta\in (0,1]$, $R\in \mathbb{N}\cap (K, \infty)$, and $C\in \mathcal{C}$. Consider the policy $\pi=\pi_{\textsf{NS}}^{\star}(L, \eta, R)$. Under the assignment of the TPMs $C$, for all $C'\in \textsf{Alt}(C)$,
\begin{equation}
\lim_{n\to\infty}\frac{Z_{CC'}^{\pi}(n)}{n}=\sum_{(\underline{d},\underline{i})\in\mathbb{S}_{R}}\ \sum_{a=1}^{K}\ \nu_{\eta,R,C}(\underline{d},\underline{i},a) \ D((P_{C}^{a})^{d_{a}}(\cdot|i_{a}) \| (P_{C'}^{a})^{d_{a}}(\cdot | i_{a})) \quad \text{almost surely}.
\label{eq:limit_Z_{hh'}(n)/n}
\end{equation}
Consequently, it follows that 
\begin{align}
\lim_{n\to\infty}\frac{M_C^{\pi}(n)}{n}=\min_{C'\in \textsf{Alt}(C)}\ \sum_{(\underline{d},\underline{i})\in\mathbb{S}_{R}}\ \sum_{a=1}^{K}\ \nu_{\eta,R,C}(\underline{d},\underline{i},a)\ D((P_{C}^{a})^{d_{a}}(\cdot|i_{a}) \| (P_{C'}^{a})^{d_{a}}(\cdot | i_{a})) \quad \text{almost surely}.
\label{eq:limit_M_h(n)/n}
\end{align}
\end{proposition}
\begin{IEEEproof}
See Appendix \ref{appndx:proof-of-prop-Z_{CC'}(n)_has_the_right_drift}.
\end{IEEEproof}
From \eqref{eq:nu-eta-R-C}, we note that the right hand side of \eqref{eq:limit_M_h(n)/n} may be lower bounded by
\begin{align}
&\eta\ \min_{C'\in \textsf{Alt}(C)}\ \sum_{(\underline{d},\underline{i})\in\mathbb{S}_{R}}\ \sum_{a=1}^{K}\ \nu_{C, R}^{\textsf{unif}}(\underline{d},\underline{i},a)\ D((P_{C}^{a})^{d_{a}}(\cdot|i_{a}) \| (P_{C'}^{a})^{d_{a}}(\cdot | i_{a})) \nonumber\\
&\hspace{1cm}+(1-\eta)\  \min_{C'\in \textsf{Alt}(C)}\ \sum_{(\underline{d},\underline{i})\in\mathbb{S}_{R}}\ \sum_{a=1}^{K}\ \nu_{C, R}^{\star}(\underline{d},\underline{i},a)\ D((P_{C}^{a})^{d_{a}}(\cdot|i_{a}) \| (P_{C'}^{a})^{d_{a}}(\cdot | i_{a}))\nonumber\\
&=\eta\ \min_{C'\in \textsf{Alt}(C)}\ \sum_{(\underline{d},\underline{i})\in\mathbb{S}_{R}}\ \sum_{a=1}^{K}\ \nu_{C, R}^{\textsf{unif}}(\underline{d},\underline{i},a)\ D((P_{C}^{a})^{d_{a}}(\cdot|i_{a}) \| (P_{C'}^{a})^{d_{a}}(\cdot | i_{a})) + (1-\eta)\  T_{R}^{\star}(C),
\end{align}
which, as $\eta\downarrow 0$, converges to $T_{R}^{\star}(C)$. Using this observation, we shall show later that our policy achieves an upper bound of $1/T_R^\star(C)$ in the limit as $\eta\downarrow 0$.

\subsection{Asymptotic Growth of Stopping Time}
\label{subsec:asymptotic-growth-stopping-time}
In this section, we demonstrate that $\tau(\pi^{\star}(L, \eta, R))$ grows as $L\to \infty$ (equivalently $\epsilon\downarrow 0$).
\begin{lemma}\label{Lemma:stopping_time_of_policy_goes_to_infinity}
Fix $\eta\in (0,1]$, $R\in \mathbb{N}\cap (K, \infty)$, and $C\in \mathcal{C}$. Under the assignment of the TPMs $C$,
\begin{equation}
\liminf_{L\to\infty}\tau(\pi^{\star}(L, \eta, R))=\infty\text{ almost surely.}
\label{eq:stopping_time_of_policy_goes_to_infinity}
\end{equation}
\end{lemma}
\begin{IEEEproof}
See Appendix \ref{appndx:proof-of-lemma-stopping-time-goes-to-infty}.
\end{IEEEproof}
As a consequence of Lemma \ref{Lemma:stopping_time_of_policy_goes_to_infinity}, we get that under the assignment of the TPMs $C$ and under $\pi=\pi^{\star}(L, \eta, R)$,
\begin{equation}
\lim_{L\to\infty}\ \frac{M_{C}^{\pi}(\tau(\pi))}{\tau(\pi)}
=\sum_{(\underline{d},\underline{i})\in\mathbb{S}_{R}}\ \sum_{a=1}^{K}\ \nu_{\eta,R,C}(\underline{d},\underline{i},a) \ D((P_{C}^{a})^{d_{a}}(\cdot|i_{a}) \| (P_{C'}^{a})^{d_{a}}(\cdot | i_{a})) \quad \text{almost surely}.
\label{eq:M_h(N(pi))/N(pi)_has_almost_correct_drift}
\end{equation}

\subsection{Almost Sure Asymptotic Upper Bound on the Stopping Time}
\label{subsec:almost-sure-upper-bound-on-stopping-time}
In this section, we derive an almost sure asymptotic upper bound on the stopping time of {\sc $R$-DCR-BAI} as $L\to\infty$, and show that this upper bound is arbitrarily close to $1/T_{R}^{\star}(C)$ under the assignment of the TPMs $C$. In the next section, we combine the almost sure upper bound of this section with a certain uniform integrability result to claim that the expected stopping time of {\sc $R$-DCR-BAI} satisfies the same upper bound as that derived in this section.
\begin{lemma}\label{Lemma:almost_sure_upper_bound_for_policy_pi_star}
Fix $\eta\in (0,1]$, $R\in \mathbb{N}\cap(K, \infty)$, and $C\in \mathcal{C}$. Under the assignment of the TPMs $C$ and under the policy $\pi=\pi^*(L,\eta, R)$, 
\begin{align}
\limsup_{L\to\infty}\, \frac{\tau(\pi)}{\log L}
&\leq  \frac{1}{\min\limits_{C'\in \textsf{Alt}(C)}\ \sum\limits_{(\underline{d},\underline{i})\in\mathbb{S}_{R}}\ \sum\limits_{a=1}^{K}\ \nu_{\eta,R,C}(\underline{d},\underline{i},a) \ D((P_{C}^{a})^{d_{a}}(\cdot|i_{a}) \| (P_{C'}^{a})^{d_{a}}(\cdot | i_{a}))}\nonumber\\
&\leq \frac{1}{\eta\,T_{R}^{\textsf{unif}}(C) + (1-\eta)\ T_{R}^{\star}(C)}\quad \text{almost surely},
\label{eq:almost_sure_upper_bound_for_policy_pi_star}
\end{align}	
where $$T_{R}^{\textsf{unif}}(C)\coloneqq \min_{C'\in \textsf{Alt}(C)}\ \sum_{(\underline{d},\underline{i})\in\mathbb{S}_{R}}\ \sum_{a=1}^{K}\ \nu_{C,R}^{\textsf{unif}}(\underline{d},\underline{i},a) \ D((P_{C}^{a})^{d_{a}}(\cdot|i_{a}) \| (P_{C'}^{a})^{d_{a}}(\cdot | i_{a})).$$
\end{lemma}
\begin{IEEEproof}
See Appendix \ref{appndx:proof-of-almost-sure-upper-bound}
\end{IEEEproof}

\subsection{Asymptotic Upper Bound on the Expected Stopping Time}
\label{subsec:asymptotic-upper-bound-on-expected-stopping-time}

In this section, we show that the expected value of the stopping time of policy {\sc $R$-DCR-BAI} satisfies an asymptotic upper bound that matches with the right hand side of \eqref{eq:almost_sure_upper_bound_for_policy_pi_star} as $L\to\infty$.
\begin{proposition}
\label{prop:upper_bound}
Fix $\eta\in (0,1]$, $R\in \mathbb{N}\cap (K, \infty)$, and $C\in \mathcal{C}$. Under the assignment of the TPMs $C$, the policy {\sc $R$-DCR-BAI} satisfies
\begin{equation}
\limsup_{L\to\infty}\frac{\mathbb{E}_{C}^{\pi}[\tau(\pi)]}{\log L}\leq \frac{1}{\eta\,T_{R}^{\textsf{unif}}(C) + (1-\eta)\ T_{R}^{\star}(C)}.
\label{eq:upper_bound}
\end{equation}
\end{proposition}
\begin{IEEEproof}
In the proof, which we provide in Appendix \ref{appndx:proof_of_upper_bound}, we first show that the family $\{\tau(\pi)/\log L:L>1\}$ is uniformly integrable. Combining \eqref{eq:almost_sure_upper_bound_for_policy_pi_star} with the uniform integrability result yields \eqref{eq:upper_bound}.
\end{IEEEproof}

\section{A Key Monotonicity Property and the Main Result}\label{sec:main_result}
In this section, we establish a key monotonicity property for $T_{R}^{\star}(C)$, which is that $T_{R}^{\star}(C)\leq T_{R'}^{\star}(C)$ for all $R<R'$. This, combined together with the fact that $T_{R}^{\star}(C)\leq T^{\star}(C)$, implies that $\lim_{R\to\infty}\ T_{R}^{\star}(C)$ exists. We conclude the section by stating the main result of the paper.

\subsection{A Key Monotonicity Property}
\label{subsec:key-monotonicity-property}
The below result asserts that $T_{R}^{\star}(C)$ is monotonically non-decreasing as $R$ increases.
\begin{lemma}
\label{lemma:T-R-star-nondecreasing-in-R}
	 $T_{R}^{\star}(C)\leq T_{R+1}^{\star}(C)$ for all $R\in \mathbb{N}\cap (K, \infty)$.
\end{lemma}
\begin{IEEEproof}
	Fix $R\in \mathbb{N}\cap (K, \infty)$. The key idea behind the proof is to note that (a) $\mathbb{S}_{R}\subset \mathbb{S}_{R+1}$, and (b) any $\nu$ that satisfies \eqref{eq:lower_bound_6_1_2}-\eqref{eq:lower_bound_6_3_4} with parameter $R$ also satisfies them with parameter $R+1$. The details follow. Let $\nu_{C,R}^{\star}$ and $\nu_{C,R+1}^{\star}$ be the optimal state-action measures when the arm delays are constrained to no more than $R$ and $R+1$ respectively. Note that both $\nu_{C,R}^{\star}$ and $\nu_{C,R+1}^{\star}$ satisfy \eqref{eq:lower_bound_6_1_2}-\eqref{eq:lower_bound_6_3_4} (with the corresponding parameters $R$ and  $R+1$). Further, $\nu_{C,R}^{\star}$ satisfies \eqref{eq:lower_bound_6_1_2}-\eqref{eq:lower_bound_6_3_4} with parameter $R+1$. Define $\widetilde{\nu}_{C,R+1}$ as
	\begin{align*}
		\widetilde{\nu}_{C,R+1}(\underline{d},\underline{i},a)\coloneqq \begin{cases}
			\nu_{C,R}^{\star}(\underline{d},\underline{i},a) & \text{ if } (\underline{d},\underline{i})\in \mathbb{S}_R, \\
			0, & \text{ otherwise}.\\
		\end{cases}
	\end{align*}
Clearly, $\widetilde{\nu}_{C,R+1}$ satisfies \eqref{eq:lower_bound_6_1_2}-\eqref{eq:lower_bound_6_3_4} with parameter $R+1$. We therefore have
\begin{align*}
T_{R+1}^{\star}(C)&=\min_{C'\in \textsf{Alt}(C)}\ \sum_{(\underline{d}, \underline{i})\in \mathbb{S}_{R+1}}\ \sum_{a=1}^{K}\ \nu_{C,R+1}^{\star}(\underline{d}, \underline{i}, a)\ D_{\textsf{KL}}((P_{C}^{a})^{d_{a}}(\cdot | i_{a}) \| (P_{C'}^{a})^{d_{a}}(\cdot | i_{a}))\\
	&\stackrel{(a)}{\geq} \min_{C'\in \textsf{Alt}(C)}\ \sum_{(\underline{d}, \underline{i})\in \mathbb{S}_{R+1}}\ \sum_{a=1}^{K}\ \widetilde{\nu}_{C,R+1}(\underline{d}, \underline{i}, a)\ D_{\textsf{KL}}((P_{C}^{a})^{d_{a}}(\cdot | i_{a}) \| (P_{C'}^{a})^{d_{a}}(\cdot | i_{a}))\\
	&= \min_{C'\in \textsf{Alt}(C)}\ \sum_{(\underline{d}, \underline{i})\in \mathbb{S}_{R}}\ \sum_{a=1}^{K}\ \widetilde{\nu}_{C,R+1}(\underline{d}, \underline{i}, a)\ D_{\textsf{KL}}((P_{C}^{a})^{d_{a}}(\cdot | i_{a}) \| (P_{C'}^{a})^{d_{a}}(\cdot | i_{a}))\\
	&=\min_{C'\in \textsf{Alt}(C)}\ \sum_{(\underline{d}, \underline{i})\in \mathbb{S}_{R}}\ \sum_{a=1}^{K}\ {\nu}_{C,R}^{\star}(\underline{d}, \underline{i}, a)\ D_{\textsf{KL}}((P_{C}^{a})^{d_{a}}(\cdot | i_{a}) \| (P_{C'}^{a})^{d_{a}}(\cdot | i_{a}))\\
	&= T_{R}^{\star}(C),
\end{align*}
thus establishing the desired result.
\end{IEEEproof}
Lemma \ref{lemma:T-R-star-nondecreasing-in-R}, in conjunction with the observation that $T_R^\star(C)\leq T^{\star}(C)$ for all $R$, implies that $\lim_{R\to\infty} T_{R}^{\star}(C)$ exists and $\lim_{R\to\infty} T_{R}^{\star}(C) \leq T^{\star}(C)$. The question of whether this inequality is an equality seems difficult to prove, and is discussed in the next section. 

\subsection{Main Result}
We are now ready to state the main result of the paper.
\begin{theorem}
\label{prop:main_result}
	Consider a multi-armed bandit with $K\geq 2$ arms in which each arm is a time homogeneous and ergodic discrete-time Markov process on the finite state space $\mathcal{S}$. Given TPMs $P_{1}, \ldots, P_{K}$ and a permutation $\sigma:\{1,\ldots, K\}\to \{1, \ldots, K\}$, let $C=(P_{\sigma(1)}, \ldots, P_{\sigma(K)})$ be the underlying assignment of the TPMs where $P_{\sigma(a)}$ denotes the TPM of arm $a$. The growth rate of the expected time required to find the best arm in $C$ satisfies the lower bound
	\begin{align}
		\liminf_{\epsilon\downarrow 0}\inf_{\pi\in\Pi(\epsilon)}\frac{\mathbb{E}_{C}^\pi[\tau(\pi)]}{\log(1/\epsilon)}\geq \frac{1}{T^{\star}(C)}.\label{eq:lower_bound_main_result}
	\end{align}
	Further, given any $\epsilon>0$, the policy $\pi^{\star}(1/\epsilon, \eta, R) \in \Pi(\epsilon)$ for all $\eta\in (0,1]$ and $R\in \mathbb{N}\cap (K, \infty)$. Additionally,
	\begin{align}		
	\limsup_{R\to\infty}\ \limsup_{\eta \downarrow 0} \ \limsup_{L\to\infty}\ \frac{\mathbb{E}_{C}^{\pi^\star(L, \eta, R)}[\tau(\pi^\star(L, \eta, R))]}{\log L}\leq \frac{1}{\lim_{R\to\infty}\ T_{R}^{\star}(C)},\label{eq:upper_bound_main_result}
	\end{align}
	thereby yielding
\begin{align}
\frac{1}{T^{\star}(C)} &\leq \liminf_{\epsilon\downarrow 0}\inf_{\pi\in\Pi(\epsilon)}\frac{\mathbb{E}_{C}^\pi[\tau(\pi)]}{\log(1/\epsilon)}\nonumber\\
&\leq \limsup_{\epsilon\downarrow 0}\inf_{\pi\in\Pi(\epsilon)}\frac{\mathbb{E}_{C}^\pi[\tau(\pi)]}{\log(1/\epsilon)}\nonumber\\
&\leq \limsup_{R\to\infty}\ \limsup_{\eta \downarrow 0} \ \limsup_{L\to\infty}\ \frac{\mathbb{E}_{C}^{\pi^\star(L, \eta, R)}[\tau(\pi^\star(L, \eta, R))]}{\log L} \leq \frac{1}{\lim_{R\to\infty}\ T_{R}^{\star}(C)}.
\label{eq:upper-and-lower-bound-combined-main-result}
\end{align}
Thus, the lower bound on the growth rate of the expected stopping time is $1/T^{\star}(C)$, and the upper bound is $1/(\lim_{R\to\infty}\ T_{R}^{\star}(C))$.
\end{theorem}

\begin{IEEEproof}
The asymptotic lower bound in \eqref{eq:lower_bound_main_result} follows from Proposition \ref{prop:lower_bound}. From Lemma \ref{lem:pi_star(L)_in_Pi(epsilon)}, we know that for any $\epsilon>0$, the policy $\pi^\star(1/\epsilon, \eta, R)\in \Pi(\epsilon)$ for all $\eta\in (0,1]$ and $R\in \mathbb{N}\cap (K, \infty)$. Therefore, it follows that
\begin{equation}
\inf_{\pi \in \Pi(\epsilon)} \frac{\mathbb{E}_{C}^\pi[\tau(\pi)]}{\log \left(1/\epsilon\right)} \leq \frac{\mathbb{E}^{\pi^\star(1/\epsilon, \eta, R)}[\tau(\pi^\star(1/\epsilon, \eta, R))]}{\log (1/\epsilon)}.
\label{eq:proof_of_main_result_temp_1}
\end{equation}
Fixing $\eta$, $R$, and letting $\epsilon\downarrow 0$ (or equivalently, substituting $L=1/\epsilon$ and letting $L\to\infty$) in \eqref{eq:proof_of_main_result_temp_1}, and using the upper bound in \eqref{eq:upper_bound}, we get
\begin{align}
\limsup_{\epsilon\downarrow 0}\inf_{\pi\in\Pi(\epsilon)}\frac{\mathbb{E}_{C}^\pi[\tau(\pi)]}{\log(1/\epsilon)} \leq \limsup_{L\to\infty}\frac{\mathbb{E}_{C}^{\pi^\star(L, \eta, R)}[\tau(\pi^\star(L, \eta, R))]}{\log L}
\leq \frac{1}{\eta\ T_{R}^{\textsf{unif}}(C)+(1-\eta)\ T_{R}^{\star}(C)}.
\label{eq:proof_of_main_result_temp_2}
\end{align}
Letting $\eta\downarrow 0$ in \eqref{eq:proof_of_main_result_temp_2} and noting that the leftmost term in \eqref{eq:proof_of_main_result_temp_2} does not depend on $\eta$, we get
\begin{align}
\limsup_{\epsilon\downarrow 0}\inf_{\pi\in\Pi(\epsilon)}\frac{\mathbb{E}_{C}^\pi[\tau(\pi)]}{\log(1/\epsilon)} \leq \limsup_{\eta\downarrow 0}\ \limsup_{L\to\infty}\frac{\mathbb{E}_{C}^{\pi^\star(L, \eta, R)}[\tau(\pi^\star(L, \eta, R))]}{\log L}
\leq \frac{1}{T_{R}^{\star}(C)}.
\label{eq:proof_of_main_result_temp_3}
\end{align}
Finally, letting $R\to\infty$ in \eqref{eq:proof_of_main_result_temp_3}, we arrive at \eqref{eq:upper-and-lower-bound-combined-main-result}.
\end{IEEEproof}
\section{On the Convergence of \texorpdfstring{$T_R^\star(C)$ to $T^\star(C)$ as $R\to\infty$}{T-R-Star-to-T-star}}
\label{sec:convergence-of-TRstar-to-Tstar}

Recall that $T^\star(C)$ is the optimal value of the infinite-dimensional LP in \eqref{eq:T_a^{star}}, where the supremum in \eqref{eq:T_a^{star}} is over all $\nu$ satisfying \eqref{eq:lower_bound_6_2}-\eqref{eq:flow-constraint}, and $T_R^\star(C)$ is the optimal value of the finite-dimensional LP in \eqref{eq:T-R-star-C} that arises when the the delay of each arm is constrained to be no more than $R$. From our exposition in Section \ref{subsec:key-monotonicity-property}, we know that $\lim_{R\to\infty} T_R^\star(C)\leq T^\star(C)$. Showing that, in general, this inequality is an equality appears to be difficult. In this section, we show that in the special case when the arm TPMs $P_1, \ldots, P_K$ have identical rows, which is akin to obtaining {\em i.i.d.} observations from the arms, we have $\lim_{R\to\infty} T_R^\star(C)= T^\star(C)$, thus leading to matching upper and lower bounds in this special setting.  

Assume that the TPMs $P_1, \ldots, P_K$ have identical rows, and suppose that $\mu_1, \ldots, \mu_K$ are the unique stationary distributions associated with $P_1, \ldots, P_K$ respectively. Then, by the convergence result \cite[Theorem 4.9]{levin2017markov} for finite-state Markov processes, each row of $P_k$ must be equal to $\mu_k$, $k=1, \ldots, K$. In this special setting, the below result states that $ T_R^\star(C)=T^\star(C)$ for all $R\in \mathbb{N}\cap (K, \infty)$, and therefore $\lim_{R\to\infty} T_R^\star(C)= T^\star(C)$.
\begin{lemma}
\label{lemma:lim-T-R-star-matches-T-star-when-TPMs-have-identical-rows}
Suppose each row of $P_k$ is equal to $\mu_k$, $k=1, \ldots, K$. In this special setting, $T^\star(C)=T_{R}^\star(C)$ for all $R\in \mathbb{N}\cap (K, \infty)$. Consequently, $\lim_{R\to\infty} T_R^\star(C)= T^\star(C)$.
\end{lemma}
\begin{IEEEproof}
The proof uses the key idea that for a given $C\in \mathcal{C}$ and for all $d\in \mathbb{N}$, $i\in \mathcal{S}$, and $C'\in \textsf{Alt}(C)$, $$ D_{\textsf{KL}}((P_{C}^{a})^{d}(\cdot|i) \| (P_{C'}^{a})^{d}(\cdot|i)) = D_{\textsf{KL}}(\mu_C^a \| \mu_{C'}^a), $$
where $\mu_C^a$ denotes the stationary distribution associated with the TPM $P_C^a$. The complete proof of Lemma~\ref{lemma:lim-T-R-star-matches-T-star-when-TPMs-have-identical-rows} is given in Appendix~\ref{appndx:proof-of-convergence-of-T-R-star-to-T-star}.
\end{IEEEproof}

As a consequence of Lemma \ref{lemma:lim-T-R-star-matches-T-star-when-TPMs-have-identical-rows}, the common expressions for $T^\star(C)$ and $T_R^\star(C)$ for all $R\in \mathbb{N}\cap (K, \infty)$ specialise to the lower bounds of \cite{Kaufmann2016,garivier2016optimal} for the case when the {\em i.i.d.} observations from each arm come from a finite alphabet, thus enabling us to recover the results of \cite{Kaufmann2016,garivier2016optimal}. For general arm TPMs with non-identical rows, we leave open the question of whether $\lim_{R\to\infty} T_R^\star(C)$ equals $T^\star(C)$ for future study.
\section{Concluding Remarks and Discussion}
\label{sec:concluding-remarks-and-discussion}

\begin{enumerate}
    \item We studied the problem of finding the best arm in a restless Markov multi-armed bandit as quickly as possible, subject to an upper bound on the error probability. For this optimal stopping problem, we showed that under the problem instance $C$, the expected time required to find the best arm with an error probability no more than $\epsilon$ is lower bounded by $\log(1/\epsilon)/T^\star(C)$ in the limit as $\epsilon\downarrow 0$ (converse). Here, $T^\star(C)$ is a problem-instance dependent constant that captures the hardness of the problem. We also devised a policy that, for an input parameter $R\in \mathbb{N}\cap (K, \infty)$, forcibly selects an arm which has not been selected for $R$ consecutive time instants, and finds the best arm in at most $\log(1/\epsilon)/T_R^\star(C)$ time instants on the average as $\epsilon\downarrow 0$ (achievability).
    
    \item We showed that $T_R^\star(C)$ is monotonically non-decreasing in $R$, and that $\lim_{R\to \infty} T_R^\star(C)\leq T^\star(C)$. Showing that, in general, this inequality is an equality appears to be a difficult problem and remains open. Notwithstanding this, we showed that in the special case when the TPM of each arm has identical rows (which is akin to obtaining {\em i.i.d.} observations from each arm), the above inequality is indeed an equality. We were thus able to recover the results of \cite{Kaufmann2016,garivier2016optimal} for the case of arms with a common, finite alphabet.

    \item The trembling hand-based policy of \cite{karthik2021detecting} is not practically implementable because it operates on the countable set $\mathbb{S}$ of {\em all} delays and last observed states which cannot be handled on a machine with finite-size memory. However, for any given $R\in \mathbb{N}\cap (K, \infty)$, our policy operates on the finite set $\mathbb{S}_R$ which can easily be stored in finite-size memory on a machine, thereby making it practically implementable.
    
    \item In our achievability analysis, we assumed that the initial state of each arm follows a certain distribution $\phi$ that is independent of the underlying assignment of the TPMs. However, this may not actually be the case. For instance, if the Markov process of each arm has evolved for a sufficiently long time and reached stationarity before the decision entity begins sampling the arms at $t=0$, then the initial state of each arm follows the arm's stationary distribution. This will lead to a mismatch between the LLR expressions in our work (resulting from $\phi$) and the actual LLR expressions (resulting from the stationary distributions). Suppose $\bar{Z}_{CC'}^\pi(n)$ denotes the analogue of \eqref{eq:LLR-final} resulting from using the stationary distributions in place of $\phi$. Then, fixing $C\in \mathcal{C}$, it can be shown that $$ \lim_{n\to \infty} \frac{Z_{CC'}^\pi(n)}{n} - \frac{\bar{Z}_{CC'}^\pi(n)}{n}=0. $$ That is, the asymptotic drift of $Z_{CC'}^\pi(n)$ is identical to that of $\bar{Z}_{CC'}^\pi(n)$, and therefore the assumption $X_0^a\sim \phi$ does not affect the asymptotic analysis in any way.
    
    \item It will be interesting to extend the results of our paper to the case when the arm TPMs $P_1, \ldots, P_K$ are not known to the decision entity beforehand. The difficulty here is that for any given $C\in \mathcal{C}$, the set of alternatives $\textsf{Alt}(C)$ is uncountably infinite. Also, the arm TPMs they must be estimated on-the-fly using the observations from the arms. In this case, showing that the TPM estimates converge to their true values is the key challenge. It will be interesting to explore these issues further. 
    
    \item The function $f$ appearing in the definition of the best arm in \eqref{eq:best_arm}, also appears implicitly in the analyses of the lower and the upper bounds wherever one evaluates $\textsf{Alt}(C)$ for any given $C$. A more realistic setting where we anticipate that $f$ will appear explicitly in the analyses of the lower and the upper bounds is one in which the decision entity only observes $Y_t^a = f(X_t^a)$ and not the underlying state $X_t^a$ of arm $a$ at time~$t$, i.e., the arms yield hidden Markov observations. More generally, suppose that $Y_t^a|X_t^a \sim P^a(\cdot|X_t^a)$ for some conditional probability distribution $P^a$, $a\in \mathcal{A}$. Because $\{Y_t^a:t\geq 0\}$ is not a Markov process in general, the analyses of the lower and the upper bounds in this setting appear to be quite challenging. It will be interesting to explore this setting in more detail.  
\end{enumerate}
\section*{Acknowledgements}

The authors are supported by a Singapore National Research Foundation Fellowship (under grant number R-263-000-D02-281) and  by the  National Research Foundation, Singapore under its AI Singapore Programme (AISG Award No: AISG2-RP-2020-018).
\appendices


\section{Proof of Proposition \ref{prop:lower_bound}}\label{appndx:proof-of-lower-bound}

It suffices to prove \eqref{eq:lower_bound} for all $\pi$ such that $\mathbb{E}_{C}^{\pi}[\tau(\pi)] < \infty$, as \eqref{eq:lower_bound} trivially holds when $\mathbb{E}_{C}^{\pi}[\tau(\pi)] = \infty$. This proof is organised as follows. In Section \ref{subsec:change-of-measure-restless-arms}, we derive a change-of-measure result that is the analogue of \cite[Lemma~18]{Kaufmann2016} for the setting of restless arms (see \eqref{eq:change_of_measure}). Using the change-of-measure result together with \cite[Lemma~19]{Kaufmann2016}, we derive in Section \ref{subsec:lower-bound-for-expected-LLR} a lower bound for the expected LLR in terms of the error probability. In Section~\ref{subsec:upper-bound-for-expected-LLR}, we derive an upper bound for the expected LLR in terms of the expected stopping time (see \eqref{eq:lower_bound_3}). Combining the lower bound of Section \ref{subsec:lower-bound-for-expected-LLR} and the upper bound of Section \ref{subsec:upper-bound-for-expected-LLR}, and letting the error probability vanish, we arrive at the lower bound \eqref{eq:lower_bound}.

\subsection{A Change-of-Measure Result for Restless Arms}
\label{subsec:change-of-measure-restless-arms}
The following change-of-measure result is the analogue of \cite[Lemma 18]{Kaufmann2016} for the setting of restless arms. The proof technique is along the lines of the proof of \cite[Lemma 18]{Kaufmann2016}.

\begin{lemma}\label{lem:change_of_measure}
	Fix $C, C'\in \mathcal{C}$. Given a policy $\pi$ with stopping time $\tau(\pi)$ such that $P_{C}^{\pi}(\tau(\pi)<\infty)=1$, $P_{C'}^{\pi}(\tau(\pi)<\infty)=1$, let
	\begin{equation}
		\mathcal{F}_{\tau(\pi)} \coloneqq \{E\in\mathcal{F}:E\cap \{\tau(\pi)= t\}\in\mathcal{F}_t\text{ for all }t\geq 0\},\label{eq:F_tau}
	\end{equation}
	where $\{\mathcal{F}_{t}:t\geq 0\}$ is as defined in \eqref{eq:filtration}. Then,  
	\begin{equation}
		P_{C'}^{\pi}(E)=\mathbb{E}_{C}^{\pi}\big[\mathbb{I}_{E}\, \exp\big(-Z_{CC'}^{\pi}(\tau(\pi))\big)\big], \quad E\in \mathcal{F}_{\tau(\pi)}.
\label{eq:change_of_measure}
	\end{equation}
\end{lemma}

\begin{IEEEproof}[Proof of Lemma \ref{lem:change_of_measure}]
We prove \eqref{eq:change_of_measure} by first demonstrating, through mathematical induction, that the relation
 \begin{equation}
 	\mathbb{E}_{C'}[g(A_{0:t},\bar{X}_{0:t})]=\mathbb{E}_{C}^{\pi}\big[g(A_{0:t},\bar{X}_{0:t})\,\exp\big(-Z_{CC'}^{\pi}(t)\big)\big]
	\label{eq:change_of_measure_equiv}
 \end{equation}
 holds for all $t\geq 0$ and for all measurable functions $g:\mathcal{A}^{t+1}\times\mathcal{S}^{t+1}\to\mathbb{R}$. Then, \eqref{eq:change_of_measure} follows from  \eqref{eq:change_of_measure_equiv} by noting that for any $E\in\mathcal{F}_{\tau(\pi)}$,
\begin{align}
 	P_{C'}^{\pi}(E) 
	&=\mathbb{E}_{C'}^{\pi}[\mathbb{I}_{E}]\nonumber\\
 	&=\mathbb{E}_{C'}^{\pi}\bigg[\sum_{t\geq 0} \ \mathbb{I}_{E\cap \{\tau(\pi)= t\}}\bigg]\nonumber\\
 	&\stackrel{(a)}{=}\sum_{t\geq 0}\ \mathbb{E}_{C'}^{\pi}\left[\mathbb{I}_{E\cap \{\tau(\pi)= t\}}\right]\nonumber\\
 	&\stackrel{(b)}{=}\sum_{t\geq 0}\ \mathbb{E}_{C}^{\pi}\left[\mathbb{I}_{E\cap \{\tau(\pi)= t\}}\,\exp\big(-Z_{CC'}^{\pi}(t)\big)\right]\nonumber\\
 	&=\sum_{t\geq 0}\ \mathbb{E}_{C}^{\pi}\left[\mathbb{I}_{E\cap \{\tau(\pi)= t\}}\,\exp\big(-Z_{CC'}^{\pi}(\tau(\pi))\big)\right]\nonumber\\
 	&=\mathbb{E}_{C}^{\pi}\left[\mathbb{I}_{E}\,\exp\big(-Z_{CC'}^{\pi}(\tau(\pi))\big)\right],\label{eq:change_of_measure_9}
 \end{align}
 where $(a)$ is due to the monotone convergence theorem, and $(b)$ above follows from \eqref{eq:change_of_measure_equiv} and the fact that $E\in\mathcal{F}_{\tau(\pi)}$ implies that $E\cap \{\tau(\pi)=t\}\in\mathcal{F}_t$ for all $t\geq 0$.
 
The proof of \eqref{eq:change_of_measure_equiv} for the case $t=0$ may be obtained as follows: for any measurable $g:\mathcal{A}\times\mathcal{S}\to\mathbb{R}$,
 \begingroup \allowdisplaybreaks\begin{align}
 	\mathbb{E}_{C'}^{\pi}[g(A_0,\bar{X}_0)]&=\sum_{a=1}^{K}\ \sum_{i\in\mathcal{S}}\ g(a,i)\ P_{C'}^{\pi}(A_0=a,\bar{X}_0=i)\nonumber\\
 	&=\sum_{a=1}^{K}\ \sum_{i\in\mathcal{S}}\ g(a,i)\ P_{C'}^{\pi}(A_0=a)\ P_{C'}^{\pi}(\bar{X}_0=i|A_0=a)\nonumber\\
 	&\stackrel{(a)}{=}\sum_{a=1}^{K}\ \sum_{i\in\mathcal{S}}\ g(a,i)\ P_{C}^{\pi}(A_0=a)\ \phi(i)\nonumber\\
 	&=\sum_{a=1}^{K}\ \sum_{i\in\mathcal{S}}\ g(a,i) \ P_{C}^{\pi}(A_0=a)\ P_{C}^{\pi}(X_0^a=i|A_{0}=a)\nonumber\\
	&=\mathbb{E}_{C}^{\pi}[g(A_0,\bar{X}_0)]\nonumber\\
	&\stackrel{(b)}{=}\mathbb{E}_{C}^{\pi}\big[g(A_0,\bar{X}_0)\,\exp\big(-Z_{CC'}^{\pi}(0)\big)\big],\label{eq:change_of_measure_1}
 \end{align}\endgroup
where in writing $(a)$, we   make use of (i) the fact that $P_{C'}^{\pi}(A_{0}=a)=P_{C}^{\pi}(A_0=a)$ because the policy $\pi$ selects arms without the knowledge of the underlying assignment of the TPMs, and (ii) the assumption that $X_{0}^{a} \sim \phi$ for all $a\in \mathcal{A}$, where $\phi$ is a probability distribution on $\mathcal{S}$ that does not depend on the underlying assignment of the TPMs. In writing $(b)$ above, we make use of the observation that
\begin{align}
 	Z_{CC'}^{\pi}(0)=\log \frac{P_{C}^{\pi}(A_0,\bar{X}_0)}{P_{C'}^{\pi}(A_0,\bar{X}_0)}=0.
	\label{eq:change_of_measure_3}
 \end{align}
 
We now assume that \eqref{eq:change_of_measure_equiv} is true for some $t>0$, and show that it also true for $t+1$. By the law of iterated expectations, $\mathbb{E}_{C'}^{\pi}[g(A_{0:t+1}, \bar{X}_{0:t+1})]=\mathbb{E}_{C'}^{\pi}[\mathbb{E}_{C'}^{\pi}[g(A_{0:t+1}, \bar{X}_{0:t+1})|\mathcal{F}_{t+1}]]$. 
Because $\mathbb{E}_{C'}^{\pi}[g(A_{0:t+1}, \bar{X}_{0:t+1})|\mathcal{F}_{t+1}]$ is a measurable function of $(A_{0:t},\bar{X}_{0:t})$, by the induction hypothesis, we have
\begin{align}
 	\mathbb{E}_{C'}^{\pi}[g(A_{0:t+1},\bar{X}_{0:t+1})|\mathcal{F}_{t+1}]&=\mathbb{E}_{C}^{\pi}\big[\mathbb{E}_{C'}^{\pi}[g(A_{0:t+1},\bar{X}_{0:t+1})\,\big|\,\mathcal{F}_{t+1}]\,\exp\big(-Z_{CC'}^{\pi}(t)\big)\big]\nonumber\\
	&=\mathbb{E}_{C}^{\pi}\big[\mathbb{E}_{C'}^{\pi}\big[g(A_{0:t+1},\bar{X}_{0:t+1})\,\exp\big(-Z_{CC'}^{\pi}(t)\big)~\big|~\mathcal{F}_{t+1}\big]\big],\label{eq:change_of_measure_5}
 \end{align}
 where the last line above follows by noting that $Z_{CC'}^{\pi}(t)$ is measurable with respect to $\mathcal{F}_{t+1}$.
 We now note that
\begin{align}
 	&\mathbb{E}_{C'}^{\pi}\big[g(A_{0:t+1},\bar{X}_{0:t+1})\ \exp\big(-Z_{CC'}^{\pi}(t)\big)~\big|~\mathcal{F}_{t+1}\big]\nonumber\\
 	&=\sum_{a=1}^{K}\ \sum_{i\in\mathcal{S}}\ g(A_{0:t},\bar{X}_{0:t}, a, i)\ P_{C'}^{\pi}(A_{t+1}=a|\mathcal{F}_{t+1})\  P_{C'}^{\pi}(\bar{X}_{t+1}=i|A_{t+1}=a,\mathcal{F}_{t+1})\  \exp\big(-Z_{CC'}^{\pi}(t)\big)\nonumber\\
 	&\stackrel{(a)}{=}\sum_{a=1}^{K}\ \sum_{i\in\mathcal{S}}\ g(A_{0:t},\bar{X}_{0:t}, a, i)\ P_{C}^{\pi}(A_{t+1}=a|\mathcal{F}_{t+1})\  P_{C'}^{\pi}(\bar{X}_{t+1}=i|A_{t+1}=a,\mathcal{F}_{t+1})\  \exp\big(-Z_{CC'}^{\pi}(t)\big),
	\label{eq:change_of_measure_6}
 \end{align}
 where in writing $(a)$ above, we make use of the fact that $P_{C'}^{\pi}(A_{t+1}=a|\mathcal{F}_{t+1})=P_{C}^{\pi}(A_{t+1}=a|\mathcal{F}_{t+1})$ because $\pi$ selects arms without the knowledge of the underlying assignment of the TPMs. Also, we note that
\begin{align}
& P_{C'}^{\pi}(\bar{X}_{t+1}=i|A_{t+1}=a, \mathcal{F}_{t+1})\,\exp\big(-Z_{CC'}^{\pi}(t)\big)\nonumber\\
 	&=\frac{P_{C'}^{\pi}(\bar{X}_{t+1}=i|A_{t+1}=a,\mathcal{F}_{t+1})}{P_{C}^{\pi}(\bar{X}_{t+1}=i|A_{t+1}=a,\mathcal{F}_{t+1})}\ \exp\big(-Z_{CC'}^{\pi}(t)\big)\ P_{C}^{\pi}(\bar{X}_{t+1}=i|A_{t+1}=a,\mathcal{F}_{t+1})\nonumber\\
 	&=\exp\big(-Z_{CC'}^{\pi}(t+1)\big)\,P_{C}^{\pi}(\bar{X}_{t+1}=i|A_{t+1}=a,\mathcal{F}_{t+1}).\label{eq:change_of_measure_7}
 \end{align}
Substituting \eqref{eq:change_of_measure_7} in \eqref{eq:change_of_measure_6} and simplifying, we get
\begin{align}
& \mathbb{E}_{C'}^{\pi}\big[g(A_{0:t+1},\bar{X}_{0:t+1})\ \exp\big(-Z_{CC'}^{\pi}(t)\big)~\big|~\mathcal{F}_{t+1}\big]\nonumber\\
&=\sum_{a=1}^{K}\ \sum_{i\in\mathcal{S}}\ g(A_{0:t},\bar{X}_{0:t}, a, i)\ P_{C}^{\pi}(A_{t+1}=a|\mathcal{F}_{t+1})\  P_{C}^{\pi}(\bar{X}_{t+1}=i|A_{t+1}=a,\mathcal{F}_{t+1})\  \exp\big(-Z_{CC'}^{\pi}(t+1)\big)\nonumber\\
&=\mathbb{E}_{C}^{\pi}\big[g(A_{0:t+1},\bar{X}_{0:t+1})\ \exp\big(-Z_{CC'}^{\pi}(t+1)\big)~\big|~\mathcal{F}_{t+1}\big].
\label{eq:change_of_measure_8}
 \end{align}
 Substituting \eqref{eq:change_of_measure_8} in \eqref{eq:change_of_measure_5}, we get
 \begin{align}
 \mathbb{E}_{C'}^{\pi}[g(A_{0:t+1},\bar{X}_{0:t+1})|\mathcal{F}_{t+1}]
	&=\mathbb{E}_{C}^{\pi}\big[\mathbb{E}_{C}^{\pi}\big[g(A_{0:t+1},\bar{X}_{0:t+1})\,\exp\big(-Z_{CC'}^{\pi}(t+1)\big)~\big|~\mathcal{F}_{t+1}\big]\big]\nonumber\\
	&=\mathbb{E}_{C}^{\pi}\big[g(A_{0:t+1},\bar{X}_{0:t+1})\,\exp\big(-Z_{CC'}^{\pi}(t+1)\big)\big].
	\label{eq:change_of_measure_10}
 \end{align}
Taking $\mathbb{E}_{C'}^{\pi}[\cdot]$ on both sides of \eqref{eq:change_of_measure_10}, we get the desired result.
\end{IEEEproof}

\subsection{A Lower Bound for \texorpdfstring{$\mathbb{E}_{C}^{\pi}[Z_{CC'}^{\pi}(\tau(\pi))]$}{Expected LLR Between C and CPrime} when \texorpdfstring{$\pi\in\Pi(\epsilon)$}{PiEpsilon}}
\label{subsec:lower-bound-for-expected-LLR}

We note the following lower bound on the expected LLR. We omit the proof as it follows directly from the proof of \cite[Lemma 19]{Kaufmann2016}. 
\begin{lemma}
\label{lemma:lower-bound-for-expected-LLR}
Fix $C\in \mathcal{C}$, $C'\in \textsf{Alt}(C)$, and $\pi$ such that $P_{C}^{\pi}(\tau(\pi) < \infty)=1$, $P_{C'}^{\pi}(\tau(\pi) < \infty)=1$. Then,
\begin{enumerate}
\item $P_{C}^{\pi}$ and $P_{C'}^{\pi}$ are mutually absolutely continuous, and
\item  for all $E\in \mathcal{F}_{\tau(\pi)}$ such that $P_{C}^{\pi}(E)>0$, $P_{C'}^{\pi}(E)>0$,
\begin{equation}
	 \mathbb{E}_{C}^{\pi}[Z_{CC'}^{\pi}(\tau(\pi))] \geq  d(P_{C}^{\pi}(E), P_{C'}^{\pi}(E)),
	\label{eq:Kaufmann_DPI_bound}
\end{equation} 
where $d(x,y)$ denotes the relative entropy between two Bernoulli distributions with parameters $x$ and $y$.
\end{enumerate}
\end{lemma}
Fix $\epsilon>0$. Recall the set $\Pi(\epsilon)$ in \eqref{eq:Pi(epsilon)}. A direct consequence of Lemma \ref{lemma:lower-bound-for-expected-LLR} is that for any $\pi\in \Pi(\epsilon)$, setting $E=\{\omega\in \Omega: \theta(\tau(\pi))= a^{\star}(C) \}$, where $a^{\star}(C)$ is the index of the best arm in $C$, noting that $P_{C}^{\pi}(E)\geq 1-\epsilon$, $P_{C'}^{\pi}(E)\leq \epsilon$ for all $C'\in \textsf{Alt}(C)$, and using the fact that $x\mapsto d(x,y)$ is monotone increasing for $x<y$ and the $y\mapsto d(x,y)$ is monotone decreasing for any fixed $x$, we get
\begin{equation}
\mathbb{E}_{C}^{\pi}[Z_{CC'}^{\pi}(\tau(\pi))]  \geq d(\epsilon, 1-\epsilon)
\label{eq:lower-bound-on-expected-LLR-in-terms-of-error-probability}
\end{equation}
for all $C'\in \textsf{Alt}(C)$. Thus, it follows that $\min_{C' \in \textsf{Alt}(C)}\ \mathbb{E}_{C}^{\pi}[Z_{CC'}^{\pi}(\tau(\pi))] \geq d(\epsilon, 1-\epsilon)$ whenever $\pi\in \Pi(\epsilon)$.

\subsection{An Upper Bound for  \texorpdfstring{$\mathbb{E}_{C}^{\pi}[Z_{CC'}^{\pi}(\tau(\pi))]$}{Expected LLR Between C and CPrime} in terms of \texorpdfstring{$ \mathbb{E}_{C}^{\pi}[\tau(\pi)]$}{Expected Stopping Time under C}}
\label{subsec:upper-bound-for-expected-LLR}

We first note the following result.
\begin{lemma}
\label{lemma:a-useful-relation}
Fix $\pi$ and $C\in \mathcal{C}$. For all $(\underline{d}, \underline{i})\in \mathbb{S}$, $a\in \mathcal{A}$, and $j\in \mathcal{S}$, 
\begin{equation}
\mathbb{E}_{C}^{\pi}\big[N(\tau(\pi), \underline{d}, \underline{i}, a, j)\big] = (P_{C}^{a})^{d_{a}}(j|i_{a})\ \mathbb{E}_{C}^{\pi}\big[N(\tau(\pi), \underline{d}, \underline{i}, a)\big].
\label{eq:a-useful-relation}
\end{equation}
\end{lemma}
\begin{IEEEproof}[Proof of Lemma \ref{lemma:a-useful-relation}]
We note that
\begin{align}
 	\mathbb{E}_{C}^{\pi}[\mathbb{E}_{C}^{\pi}[N(\tau(\pi),\underline{d},\underline{i},a,j)|X_{a-1}^a]|\tau(\pi)]&=\mathbb{E}_{C}^{\pi}\bigg[\mathbb{E}_{C}^{\pi}\bigg[\sum_{t=K}^{\tau(\pi)} 1_{\{\underline{d}(t)=\underline{d},\underline{i}(t)=\underline{i},A_{t}=a,X_{t}^a=j\}}\bigg|X_{a-1}^a\bigg]\bigg|\tau(\pi)\bigg]\nonumber\\
 	&=\mathbb{E}_{C}^{\pi}\bigg[\sum_{t=K}^{\tau(\pi)}P_{C}^{\pi}(\underline{d}(t)=\underline{d},\underline{i}(t)=\underline{i},A_{t}=a,X_{t}^a=j|X_{a-1}^a)\,\bigg|\tau(\pi)\bigg].\label{eq:relation_btw_two_qty_1}
 \end{align}
 For each $t$ in the range of the summation in \eqref{eq:relation_btw_two_qty_1}, the conditional probability term for $t$ may be expressed as
 \begingroup \allowdisplaybreaks\begin{align}
 	& P_{C}^{\pi}(\underline{d}(t)=\underline{d},\underline{i}(t)=\underline{i},A_{t}=a,X_{t}^a=j|X_{a-1}^a)\nonumber\\
 	&=P_{C}^{\pi}(\underline{d}(t)=\underline{d},\underline{i}(t)=\underline{i},A_{t}=a|X_{a-1}^a)\cdot P_{C}^{\pi}(X_{t}^a=j|A_{t}=a,\underline{d}(t)=\underline{d},\underline{i}(t)=\underline{i},X_{a-1}^a)\nonumber\\
 	&=P_{C}^{\pi}(\underline{d}(t)=\underline{d},\underline{i}(t)=\underline{i},A_{t}=a|X_{a-1}^a)\cdot (P_{C}^a)^{d_a}(j|i_a).
 	\label{eq:relation_btw_two_qty_2}
 \end{align}\endgroup
 Plugging \eqref{eq:relation_btw_two_qty_2} back in \eqref{eq:relation_btw_two_qty_1} and taking $\mathbb{E}_{C}^{\pi}[\cdot]$ on both sides of \eqref{eq:relation_btw_two_qty_1}, we arrive at \eqref{eq:a-useful-relation}.
\end{IEEEproof}

From \eqref{eq:LLR-final}, we note that for all $C'\in \textsf{Alt}(C)$,
\begin{align}
	& \mathbb{E}_{C}^{\pi}[Z_{CC'}^{\pi}(\tau(\pi))]\nonumber\\
	&\stackrel{(a)}{=}\mathbb{E}_{C}^{\pi}\bigg[\sum_{a=1}^{K}\ \log \frac{P_{C}^{\pi}(X_{a-1}^a)}{P_{C'}^{\pi}(X_{a-1}^a)}\bigg]+\mathbb{E}_{C}^{\pi}\bigg[\sum_{(\underline{d},\underline{i})\in\mathbb{S}}\ \sum_{a=1}^{K}\ \sum_{j\in\mathcal{S}}\ N(\tau(\pi),\underline{d},\underline{i},a,j)\ \log\frac{(P_C^{a})^{d_a}(j|i_a)}{(P_{C'}^{a})^{d_a}(j|i_a)}\bigg]\nonumber\\
	&=\mathbb{E}_{C}^{\pi}\bigg[\sum_{a=1}^{K}\ \log \frac{P_{C}^{\pi}(X_{a-1}^a)}{P_{C'}^{\pi}(X_{a-1}^a)}\bigg]+\sum_{(\underline{d},\underline{i})\in\mathbb{S}}\ \sum_{a=1}^{K}\ \sum_{j\in\mathcal{S}}\ \mathbb{E}_{C}^{\pi}\big[N(\tau(\pi),\underline{d},\underline{i},a,j)\big]\ \log\frac{(P_C^{a})^{d_a}(j|i_a)}{(P_{C'}^{a})^{d_a}(j|i_a)}\nonumber\\
	&\stackrel{(b)}{=}\mathbb{E}_{C}^{\pi}\bigg[\sum_{a=1}^{K}\ \log \frac{P_{C}^{\pi}(X_{a-1}^a)}{P_{C'}^{\pi}(X_{a-1}^a)}\bigg]+\sum_{(\underline{d},\underline{i})\in\mathbb{S}}\ \sum_{a=1}^{K}\ \sum_{j\in\mathcal{S}}\ \mathbb{E}_{C}^{\pi}\big[N(\tau(\pi),\underline{d},\underline{i},a)\big]\ (P_{C}^{a})^{d_{a}}(j|i_{a}) \ \log\frac{(P_C^{a})^{d_a}(j|i_a)}{(P_{C'}^{a})^{d_a}(j|i_a)}\nonumber\\
	&=\mathbb{E}_{C}^{\pi}\bigg[\sum_{a=1}^{K}\ \log \frac{P_{C}^{\pi}(X_{a-1}^a)}{P_{C'}^{\pi}(X_{a-1}^a)}\bigg]+\sum_{(\underline{d},\underline{i})\in\mathbb{S}}\ \sum_{a=1}^{K}\ \mathbb{E}_{C}^{\pi}\big[N(\tau(\pi),\underline{d},\underline{i},a)\big]\ D_{\textsf{KL}}((P_C^{a})^{d_a}(\cdot|i_a) \| (P_{C'}^{a})^{d_a}(j|i_a)).
	\label{eq:lower_bound_1}
\end{align}
In the above chain of equations, $(a)$ follows from the dominated convergence theorem (noting that each row of $(P_{C}^{a})^{d}$ is mutually absolutely continuous with respect to the corresponding row of $(P_{C'}^{a})^{d}$ for all $d\geq 1$), and $(b)$ follows from Lemma \ref{lemma:a-useful-relation}. Continuing with \eqref{eq:lower_bound_1}, we have
\begin{align}
& \mathbb{E}_{C}^{\pi}[Z_{CC'}^{\pi}(\tau(\pi))] \leq \mathbb{E}_{C}^{\pi}\bigg[\sum_{a=1}^{K}\ \log \frac{1}{P_{C'}^{\pi}(X_{a-1}^a)}\bigg] \nonumber\\*
&\hspace{3cm}+ \sum_{(\underline{d},\underline{i})\in\mathbb{S}}\ \sum_{a=1}^{K}\ \mathbb{E}_{C}^{\pi}\big[N(\tau(\pi),\underline{d},\underline{i},a)\big]\ D_{\textsf{KL}}((P_C^{a})^{d_a}(\cdot|i_a) \| (P_{C'}^{a})^{d_a}(j|i_a))\nonumber\\
& \leq \mathbb{E}_{C}^{\pi}\bigg[\sum_{a=1}^{K}\ \log \frac{1}{P_{C'}^{\pi}(X_{a-1}^a)}\bigg] \nonumber\\
&\hspace{1.5cm}+ (\mathbb{E}_{C}^{\pi}[\tau(\pi)-K+1])\  \sum_{(\underline{d},\underline{i})\in\mathbb{S}}\ \sum_{a=1}^{K}\ \frac{\mathbb{E}_{C}^{\pi}\big[N(\tau(\pi),\underline{d},\underline{i},a)\big]}{\mathbb{E}_{C}^{\pi}[\tau(\pi)-K+1]}\ D_{\textsf{KL}}((P_C^{a})^{d_a}(\cdot|i_a) \| (P_{C'}^{a})^{d_a}(j|i_a)),
\label{eq:eq:lower_bound_2}
\end{align}
for all $C'\in \textsf{Alt}(C)$, from which it follows that 
\begin{align}
&\min_{C'\in \textsf{Alt}(C)}\ \mathbb{E}_{C}^{\pi}[Z_{CC'}^{\pi}(\tau(\pi))] \leq  \min_{C'\in \textsf{Alt}(C)}\ \mathbb{E}_{C}^{\pi}\bigg[\sum_{a=1}^{K}\ \log \frac{1}{P_{C'}^{\pi}(X_{a-1}^a)}\bigg] \nonumber\\
&\hspace{1cm}+ (\mathbb{E}_{C}^{\pi}[\tau(\pi)-K+1])\  \min_{C'\in \textsf{Alt}(C)}\  \sum_{(\underline{d},\underline{i})\in\mathbb{S}}\ \sum_{a=1}^{K}\ \frac{\mathbb{E}_{C}^{\pi}\big[N(\tau(\pi),\underline{d},\underline{i},a)\big]}{\mathbb{E}_{C}^{\pi}[\tau(\pi)-K+1]}\ D_{\textsf{KL}}((P_C^{a})^{d_a}(\cdot|i_a) \| (P_{C'}^{a})^{d_a}(j|i_a))\nonumber\\
&\leq \min_{C'\in \textsf{Alt}(C)}\ \mathbb{E}_{C}^{\pi}\bigg[\sum_{a=1}^{K}\ \log \frac{1}{P_{C'}^{\pi}(X_{a-1}^a)}\bigg] \nonumber\\
&\hspace{1cm}+ (\mathbb{E}_{C}^{\pi}[\tau(\pi)-K+1])\  \bigg\lbrace\sup_{\nu}\ \min_{C'\in \textsf{Alt}(C)}\  \sum_{(\underline{d},\underline{i})\in\mathbb{S}}\ \sum_{a=1}^{K}\ \nu(\underline{d}, \underline{i}, a)\ D_{\textsf{KL}}((P_C^{a})^{d_a}(\cdot|i_a) \| (P_{C'}^{a})^{d_a}(j|i_a))\bigg\rbrace,
\label{eq:lower_bound_3}
\end{align}
where the supremum in \eqref{eq:lower_bound_3} is over all $\nu$ that satisfy \eqref{eq:lower_bound_6_2}-\eqref{eq:lower_bound_6_3}. The expression within braces in \eqref{eq:lower_bound_3} is $T^{\star}(C)$.

\subsection{The Final Steps}
Combining the results of Sections \ref{subsec:lower-bound-for-expected-LLR} and \ref{subsec:upper-bound-for-expected-LLR}, we get
\begin{align}
d(\epsilon, 1-\epsilon)\leq \min_{C'\in \textsf{Alt}(C)}\ \mathbb{E}_{C}^{\pi}\bigg[\sum_{a=1}^{K}\ \log \frac{1}{P_{C'}^{\pi}(X_{a-1}^a)}\bigg] + (\mathbb{E}_{C}^{\pi}[\tau(\pi)-K+1])\  T^{\star}(C).
\label{eq:lower_bound_4}
\end{align}
Noting that (a) $d(\epsilon, 1-\epsilon)/\log(1/\epsilon) \to 1$ as $\epsilon \downarrow 0$, and (b) the first term on the right hand side of \eqref{eq:lower_bound_4} is bounded from above, by dividing \eqref{eq:lower_bound_4} throughout by $d(\epsilon, 1-\epsilon)$ and letting $\epsilon\downarrow 0$, we arrive at the lower bound \eqref{eq:lower_bound}.


\section{Proof of Lemma \ref{lemma:ergodicity-under-unif-SRS-policy}}
\label{appndx:proof-of-lemma-ergodicity-under-unif-SRS-policy}
Fix an assignment of the TPMs $C\in \mathcal{C}$. In this proof, we establish that under the policy $\pi_{R}^{\textsf{unif}}$, the process $\{(\underline{d}(t), \underline{i}(t)):\ t\geq K\}$ is irreducible, aperiodic, positive recurrent, and therefore ergodic.
\begin{IEEEproof}[Proof of Irreducibility]
	Consider any two states $(\underline{d},\underline{i}), (\underline{d}',\underline{i}')\in\mathbb{S}_{R}$, and suppose that the process $\{(\underline{d}(t), \underline{i}(t)):t\geq K\}$ is in the state $(\underline{d},\underline{i})$ at time $t=T_0$. We now demonstrate that there exists an integer $N$ (possibly depending on $(\underline{d},\underline{i})$ and $(\underline{d}',\underline{i}')$) such that $$P_{C}^{\pi_{R}^{\textsf{unif}}}(\underline{d}(T_{0}+N)=\underline{d}', \ \underline{i}(T_{0}+N)=\underline{i}'|\underline{d}(T_{0})=\underline{d}, \ \underline{i}(T_{0})=\underline{i})>0.$$
Assume without loss of generality that $\underline{d}'$, the vector of arm delays in the destination state $(\underline{d}', \underline{i}')$, is such that $d_1'>d_2'>\cdots>d_K^\prime=1$. Noting that $P_1, \ldots, P_{K}$ are TPMs on the finite set $\mathcal{S}$, we use \cite[Proposition 1.7]{levin2017markov} for finite state Markov processes to deduce that there exist integers $M_{1}, \ldots, M_{K}$ such that for all $m\geq M\coloneqq \max\{M_{1}, \ldots, M_{K}\}$,
\begin{equation}
	P_1^m(j|i)>0, \ldots, P_K^m(j|i)>0 \quad \text{ for all }i,j\in\mathcal{S}.\label{eq:P_a^M_strictly_pos_entries}
\end{equation}
Order the components of $\underline{d}$, the vector of arm delays in the starting state $(\underline{d}, \underline{i})$, in decreasing order. Under $\pi_{R}^{\textsf{unif}}$, consider the sequence of arm selections and observations as follows: for a total of $M$ time instants, from $t=T_{0}$ to $t=T_{0}+M-1$, select the arms in a round robin fashion in the decreasing order of their component values in $\underline{d}$. At time $t=T_{0}+M$, select arm $1$ and observe the state $i_{1}'$ on it. Thereafter, select arms $2, \ldots, K$ in a round robin fashion in the decreasing order of their component values in $\underline{d}$ until time $t=T_{0}+M+d_{1}'-d_{2}'-1$. At time $t=T_{0}+M+d_{1}'-d_{2}'$, select arm $2$ and observe the state $i_{2}'$ on it. Continue the round robin sampling on  arms $3, \ldots, K$ till time $t=T_{0}+M+d_{1}'-d_{3}'-1$. At time $t=T_{0}+M+d_{1}'-d_{3}'$, select arm $3$ and observe the state $i_{3}'$ on it. Continue this process till arm $K$ is selected at time $t=T_{0}+M+d_{1}'-1$ and the state $i_{K}'$ is observed on it.

Clearly, because of the round robin selection procedure, the delay of each arm at any time is no more than $K$. Also, the above sequence of arm selections and observations leads to the state $(\underline{d}',\underline{i}')$ at time  $t=T_{0}+M+d_1'$. Thus, the probability of starting from the state $(\underline{d},\underline{i})$ and reaching the state $(\underline{d}',\underline{i}')$ may be lower bounded by the probability that the above sequence of actions and observations occur under $\pi_{R}^{\textsf{unif}}$, which in turn may be lower bounded by
\begin{align}
	&\bigg(\frac{1}{K}\bigg)^{M+d_{1}'} \cdot  \left[\prod_{a=1}^{K}(P_C^a)^{M+d_{a}+d_1'-d_a'}(i_a'|i_a)\right].
	\label{eq:irreducibility-1}
\end{align}
Noting that $M\leq M+d_{1}+d_{1}'-d_{a}' \leq M+2R-1$, let
\begin{equation}
\bar{\varepsilon} \coloneqq \min\Big\lbrace (P_{C}^{a})^{m}(j|i):\ i, j\in \mathcal{S}, \ a\in \mathcal{A}, \ M \leq m \leq M+2R-1 \Big\rbrace.
\label{eq:bar-varepsilon}
\end{equation}
It is clear that $\bar{\varepsilon}>0$, and \eqref{eq:irreducibility-1} may further be lower bounded by
\begin{equation}
\bigg(\frac{1}{K}\bigg)^{M+d_{1}'}  \bar{\varepsilon}^{K}>0.
\end{equation}
Thus, we see that the Markov process $\{(\underline{d}(t),\underline{i}(t):t\geq K)\}$ is in the state $(\underline{d}',\underline{i}')$ after $N=M+d_1'$ time instants with a strictly positive probability. This establishes irreducibility.
\end{IEEEproof}

\begin{IEEEproof}[Proof of Aperiodicity]
	It suffices to show that for each $(\underline{d},\underline{i})\in\mathbb{S}_{R}$, there exists $N$ (possibly depending on $(\underline{d}, \underline{i})$) such that the probability of the process $\{(\underline{d}(t),\underline{i}(t)): t\geq K\}$ starting from the state $(\underline{d},\underline{i})$ at some time $t=T_{0}$ and returning to the state $(\underline{d},\underline{i})$ after $N$ time instants and also after $N+1$ time instants is strictly positive. This follows directly from the proof of irreducibility presented above by setting $(\underline{d}', \underline{i}')=(\underline{d}, \underline{i})$ and $N=M+d_{1}$,  where $M$ is such that \eqref{eq:P_a^M_strictly_pos_entries} holds for all $m\geq M$. 
\end{IEEEproof}

\begin{IEEEproof}[Proof of positive recurrence]
	 This follows from the facts that (a) $\mathbb{S}_{R}$ is finite, (b) $\{(\underline{d}(t),\underline{i}(t)): t\geq K\}$ is irreducible under $\pi_{R}^{\textsf{unif}}$, and (c) an irreducible Markov process evolving on a finite state space is positive recurrent.
\end{IEEEproof}


\section{Proof of Lemma \ref{lem:liminf_Z_{CC'}(n)_strictly_positive}}
\label{appndx:proof-of-lemma-liminf_Z_{CC'}(n)_strictly_positive}

This proof is organised as follows. First, we show in Section \ref{subsec:liminf-N(n,d,i)} that for all $(\underline{d}, \underline{i})\in \mathbb{S}_{R}$, 
\begin{equation}
	\liminf_{n\to\infty}\frac{N(n,\underline{d},\underline{i})}{n}>0\quad \text{almost surely}.\label{eq:lim_N(n,d,i)}
\end{equation}
Next, we show in Section \ref{subsec:liminf-N(n,d,i,a)} that almost surely,
\begin{equation}
\liminf_{n\to\infty}\ \frac{N(n, \underline{d}, \underline{i}, a)}{n} 
\begin{cases}
>0, &\text{if } (\underline{d}, \underline{i}) \in \mathbb{S}_{R,a} \text{ or }(\underline{d}, \underline{i}) \notin \bigcup_{a'=1}^{K}\ \mathbb{S}_{R, a'},\\
=0, &\text{if } (\underline{d}, \underline{i}) \in \mathbb{S}_{R,a'}\text{ for some }a'\neq a.
\end{cases}
\label{eq:liminf-N(n,d,i,a)/n}
\end{equation}
Using the above results, we establish \eqref{eq:liminf_Z_{hh'}(n)_strictly_positive} in Section \ref{subsec:completing-proof-of-strict-positivity-of-drift}.

\subsection{Limiting Drift of \texorpdfstring{$N(n, \underline{d}, \underline{i})$}{Number of Times d i is Observed}}
\label{subsec:liminf-N(n,d,i)}

Let $M$ be sufficiently large so that \eqref{eq:P_a^M_strictly_pos_entries} holds for all $m\geq M$.
Fix an arbitrary $(\underline{d}, \underline{i})\in\mathbb{S}_{R}$, and assume without loss of generality that $\underline{d}$ is such that $d_1>d_2>\cdots>d_K=1$. Let $p_{C}(\underline{d}, \underline{i})$ denote the probability of the process  $\{(\underline{d}(t),\underline{i}(t)):t\geq K\}$ starting in the state $(\underline{d},\underline{i})$ and returning back to the state $(\underline{d},\underline{i})$ under the assignment of the TPMs $C$. Following the exposition in Appendix \ref{appndx:proof-of-lemma-ergodicity-under-unif-SRS-policy}, it can be shown that $p_{C}(\underline{d}, \underline{i})>0$. Now, the term $N(n,\underline{d},\underline{i})$ may be lower bounded almost surely by the number of visits to the state $(\underline{d},\underline{i})$ measured only at times $t=K+M+d_1, K+2(M+d_1),K+3(M+d_1)$ and so on until time $t=n$. At each of these time instants, the probability that the process $\{(\underline{d}(t),\underline{i}(t)):t\geq K\}$ is in the state $(\underline{d},\underline{i})$ under the assignment of the TPMs $C$ is equal to $p_{C}(\underline{d},\underline{i})$. Thus, we have
\begin{align}
N(n,\underline{d},\underline{i})\geq \text{Bin}\left(\frac{n-K+1}{M+d_1},\ p_{C}(\underline{d}, \underline{i})\right)\quad \text{almost surely},\label{eq:N(n,d,i)_lower_bounded_by_binomial}
\end{align}
where the notation $\text{Bin}(m,q)$ denotes a Binomial random variable with parameters $m$ and $q$. It then follows that, almost surely,
\begin{align}
\liminf_{n\to\infty}\frac{N(n,\underline{d},\underline{i})}{n}&\geq \liminf_{n\to\infty}\frac{\text{Bin}\left(\frac{n-K+1}{M+d_1},~~p_{C}(\underline{d}, \underline{i})\right)}{n}\nonumber\\
&=\liminf_{n\to\infty} \frac{\text{Bin}\left(\frac{n-K+1}{M+d_1},\ p_{C}(\underline{d}, \underline{i})\right)}{\frac{n-K+1}{M+d_1}}\cdot \frac{n-K+1}{n}\cdot \frac{1}{M+d_1}\nonumber\\
&\stackrel{(a)}{=}\frac{p_{C}(\underline{d}, \underline{i})}{M+d_1}\nonumber\\
&>0,\label{eq:liminf_N(n,d,i)_strictly positive}
\end{align}
where $(a)$ above is due to the strong law of large numbers. This establishes \eqref{eq:lim_N(n,d,i)}.

\subsection{Limiting Drift of \texorpdfstring{$N(n, \underline{d}, \underline{i}, a)$}{Number of Times d i a is Observed}}
\label{subsec:liminf-N(n,d,i,a)}

If $(\underline{d}, \underline{i})\in \mathbb{S}_{R,a}$, then $N(n, \underline{d}, \underline{i}, a)=N(n, \underline{d}, \underline{i})$, and consequently $$\liminf_{n\to\infty}\ \frac{N(n, \underline{d}, \underline{i}, a)}{n}=\liminf_{n\to\infty} \ \frac{N(n, \underline{d}, \underline{i})}{n}>0 \quad \text{almost surely}.$$ If $(\underline{d}, \underline{i})\in \mathbb{S}_{R,a'}$ for some $a'\neq a$, then $N(n, \underline{d}, \underline{i}, a)=0$ for all $n\geq K$. Thus, it remains to show \eqref{eq:liminf-N(n,d,i,a)/n} for $(\underline{d}, \underline{i})\notin \bigcup_{a'=1}^{K}\ \mathbb{S}_{R,a'}$. Fix one such arbitrary $(\underline{d}, \underline{i})$ and define
\begin{equation}
S(n,\underline{d},\underline{i},a)\coloneqq \sum_{t=K}^{n}\Big[\mathbb{I}_{\{A_t=a,\,\underline{d}(t)=\underline{d},\,\underline{i}(t)=\underline{i}\}}-P_{C}^{\pi}(A_t=a,\,\underline{d}(t)=\underline{d},\,\underline{i}(t)=\underline{i}|A_{0:t-1},\bar{X}_{0:t-1})\Big].
\label{eq:S(n,d,i.a)}
\end{equation}
For each $t\geq K$, because $|\mathbb{I}_{\{A_t=a,\underline{d}(t)=\underline{d},\,\underline{i}(t)=\underline{i}\}}-P_{C}^{\pi}(A_t=a,\,\underline{d}(t)=\underline{d},\,\underline{i}(t)=\underline{i}|A_{0:t-1},\bar{X}_{0:t-1})|\leq 2$ almost surely, and $\mathbb{E}_{C}^{\pi}[\mathbb{I}_{\{A_t=a,\,\underline{d}(t)=\underline{d},\,\underline{i}(t)=\underline{i}\}}-P_{C}^{\pi}(A_t=a,\,\underline{d}(t)=\underline{d},\,\underline{i}(t)=\underline{i}|A_{0:t-1},\bar{X}_{0:t-1})|A_{0:t-1},\bar{X}_{0:t-1}]=0$ almost surely, the collection $\{\mathbb{I}_{\{A_t=a,\,\underline{d}(t)=\underline{d},\,\underline{i}(t)=\underline{i}\}}-P_{C}^{\pi}(A_t=a,\,\underline{d}(t)=\underline{d},\,\underline{i}(t)=\underline{i}|A_{0:t-1},\,\bar{X}_{0:t-1})\}_{t\geq K}$ is a bounded martingale difference sequence. Using the concentration result \cite[Theorem 1.2A]{victor1999general} for bounded martingale difference sequences, and subsequently applying the Borel--Cantelli lemma, we get that
\begin{equation}
	\frac{S(n,\underline{d},\underline{i},a)}{n}\longrightarrow 0\quad \text{as }n\to\infty,\quad \text{almost surely}.
\end{equation}
 This implies that for every choice of $\varepsilon>0$, there exists $N_\varepsilon=N_{\varepsilon}(\underline{d}, \underline{i}, a)$ sufficiently large such that
\begin{equation}
\frac{N(n,\underline{d},\underline{i},a)}{n}\geq \frac{1}{n}\ \sum_{t=K}^{n}P_{C}^{\pi}(A_t=a,\,\underline{d}(t)=\underline{d},\,\underline{i}(t)=\underline{i}|A_{0:t-1},\bar{X}_{0:t-1}) -\varepsilon \quad \forall \ n\geq N_\varepsilon, \text{ almost surely}.
\label{eq:frac{N(n,d,i,a)}{n}_is_lower_bounded}
\end{equation}
Now, for each $t\geq K$, under $\pi=\pi^\star(L, \eta, R)$,
\begin{align}
	&P_{C}^{\pi}(A_t=a,\,\underline{d}(t)=\underline{d},\,\underline{i}(t)=\underline{i}|A_{0:t-1},\,\bar{X}_{0:t-1})\nonumber\\
	&=P_{C}^{\pi}(A_t=a|\underline{d}(t)=\underline{d},\underline{i}(t)=\underline{i}, A_{0:t-1}, \bar{X}_{0:t-1})\cdot P_{C}^{\pi}(\underline{d}(t)=\underline{d},\,\underline{i}(t)=\underline{i}|A_{0:t-1},\bar{X}_{0:t-1})\nonumber\\
	&=\lambda_{\eta, R, \bar{C}(t)}(a|\underline{d}, \underline{i})\cdot P_{C}^{\pi}(\underline{d}(t)=\underline{d},\,\underline{i}(t)=\underline{i}|A_{0:t-1},\bar{X}_{0:t-1})\nonumber\\
	&\stackrel{(a)}{=}\frac{\eta\ \nu_{\bar{C}(t), R}^{\textsf{unif}}(\underline{d}, \underline{i}, a) + (1-\eta)\ \nu_{\bar{C}(t), R}^{\star}(\underline{d}, \underline{i}, a)}{\eta\ \mu_{\bar{C}(t), R}^{\textsf{unif}}(\underline{d}, \underline{i})+(1-\eta)\ \sum_{a'=1}^{K}\ \nu_{\bar{C}(t), R}^{\star}(\underline{d}, \underline{i}, a')}\cdot \mathbb{I}_{\{(\underline{d}(t)=\underline{d},\,\underline{i}(t)=\underline{i}\}}\nonumber\\
	&\geq \frac{\eta}{K}\cdot \mu_{\bar{C}(t), R}^{\textsf{unif}}(\underline{d}, \underline{i}) \cdot \mathbb{I}_{\{(\underline{d}(t)=\underline{d},\underline{i}(t)=\underline{i}\}}\nonumber\\
	&\geq \frac{\eta}{K}\cdot \mu_{R}^{\textsf{min}}\cdot \mathbb{I}_{\{(\underline{d}(t)=\underline{d},\underline{i}(t)=\underline{i}\}},\label{eq:lower_bound_P(A_t=a,d,i)}
\end{align}
where in writing $(a)$ above, we use the fact that $(\underline{d}(t),\underline{i}(t))$ is measurable with respect to the history $(A_{0:t-1},\,\bar{X}_{0:t-1})$, and $\mu_{R}^{\textsf{min}}$ in \eqref{eq:lower_bound_P(A_t=a,d,i)} is as defined in \eqref{eq:mu-min}.

Plugging \eqref{eq:lower_bound_P(A_t=a,d,i)} in \eqref{eq:frac{N(n,d,i,a)}{n}_is_lower_bounded}, we get
\begin{align}
\frac{N(n,\underline{d},\underline{i},a)}{n} &\geq \frac{\eta}{K}\cdot \mu_{R}^{\textsf{min}} \cdot \frac{N(n,\underline{d},\underline{i})}{n}-\varepsilon\quad \forall~ n\geq N_\varepsilon, \quad \text{almost surely}.
\label{eq:frac{N(n,d,i,a)}{n}_is_lower_bounded_1}
\end{align}
Using \eqref{eq:liminf_N(n,d,i)_strictly positive} in \eqref{eq:frac{N(n,d,i,a)}{n}_is_lower_bounded_1}, we get that
\begin{equation}
	\frac{N(n,\underline{d},\underline{i},a)}{n-K+1} \geq \frac{\eta}{K}\cdot \mu_{R}^{\textsf{min}}\cdot  \frac{p_{C}(\underline{d}, \underline{i})}{2(M+d_1)}-\varepsilon\label{eq:frac{N(n,d,i,a)}{n}_is_lower_bounded_2}
\end{equation}
for all $n$ large, almost surely.
Setting $\varepsilon=\frac{\eta}{2K}\cdot \mu_{R}^{\textsf{min}} \cdot \frac{p_{C}(\underline{d}, \underline{i})}{2(M+d_1)}$ establishes \eqref{eq:liminf-N(n,d,i,a)/n}. 

\subsection{Completing the Proof of Lemma \ref{lem:liminf_Z_{CC'}(n)_strictly_positive}}
\label{subsec:completing-proof-of-strict-positivity-of-drift}

From \eqref{eq:liminf-N(n,d,i,a)/n}, it follows that whenever $\liminf_{n\to\infty} \ N(n, \underline{d}, \underline{i}, a)/n>0$ almost surely, we may apply the ergodic theorem to deduce that under the assignment of the TPMs $C$,
\begin{equation}
	\frac{N(n,\underline{d},\underline{i},a,j)}{N(n,\underline{d},\underline{i},a)}\longrightarrow (P_C^a)^{d_a}(j|i_a)\quad \text{as }n\to\infty, \quad \text{almost surely}.\label{eq:limit_for_N(n,d,i,a,j)/N(n,d,i,a)}
\end{equation}
Under the constraint that the delay of each arm is at most $R$,
\begin{align}
\frac{Z^{\pi}_{CC'}(n)}{n}
=\frac{1}{n}\ \sum_{a=1}^{K} \log \frac{P_{C}^{\pi}(X_{a-1}^{a})}{P_{C'}^{\pi}(X_{a-1}^{a})} + \sum_{(\underline{d}, \underline{i})\in \mathbb{S}_{R}}\ \sum_{a=1}^{K}\ \sum_{j\in \mathcal{S}}\ \frac{N(n, \underline{d}, \underline{i}, a, j)}{n} \ \log \frac{(P_{C}^{a})^{d_{a}}(j|i_{a})}{(P_{C'}^{a})^{d_{a}}(j|i_{a})}.
\label{eq:Z_{CC'}(n)/n_lower_bound_1}
\end{align}
The first term in \eqref{eq:Z_{CC'}(n)/n_lower_bound_1} may be lower bounded as follows: assuming that $X_{0}^{a}\sim \phi$ for all $a\in \mathcal{A}$, where $\phi$ is a probability distribution on $\mathcal{S}$ that is independent of the underlying TPMs $C$ and puts a strictly positive mass on each state in $\mathcal{S}$, it follows that
\begin{align}
\frac{1}{n}\ \sum_{a=1}^{K} \log \frac{P_{C}^{\pi}(X_{a-1}^{a})}{P_{C'}^{\pi}(X_{a-1}^{a})} &\geq \frac{1}{n}\ \sum_{a=1}^{K} \log P_{C}^{\pi}(X_{a-1}^{a})\nonumber\\
&= \frac{1}{n}\ \sum_{j\in \mathcal{S}}\ \mathbb{I}_{\{X_{a-1}^{a}=j\}}\ \sum_{a=1}^{K} \log \left(\sum_{i\in \mathcal{S}}\ \phi(i)\ (P_{C}^{a})^{a-1}(j|i)\right).
\label{eq:strict-positive-drift-first-term-1}
\end{align}
Because the right hand side of \eqref{eq:strict-positive-drift-first-term-1} converges to $0$ as $n\to\infty$, given any $\varepsilon>0$, there exists $N_{1}=N_{1}(\varepsilon)$ such that 
\begin{align}
\frac{1}{n}\ \sum_{a=1}^{K} \log \frac{P_{C}^{\pi}(X_{a-1}^{a})}{P_{C'}^{\pi}(X_{a-1}^{a})} &\geq -\varepsilon \quad \text{for all }n\geq N_{1},\quad \text{almost surely}
\label{eq:strict-positive-drift-first-term-2}
\end{align}
The second term in \eqref{eq:Z_{CC'}(n)/n_lower_bound_1} may be expressed as 
\begin{align}
\sum_{(\underline{d}, \underline{i})\notin \bigcup_{a'=1}^{K} \mathbb{S}_{R, a'}}\ \sum_{a=1}^{K}\ \frac{N(n,\underline{d}, \underline{i}, a, j)}{n} \ \log \frac{(P_{C}^{a})^{d_{a}}(j|i_{a})}{(P_{C'}^{a})^{d_{a}}(j|i_{a})} + \sum_{a=1}^{K}\ \sum_{(\underline{d}, \underline{i})\in \mathbb{S}_{R,a}}\  \frac{N(n,\underline{d}, \underline{i}, a, j)}{n} \ \log \frac{(P_{C}^{a})^{d_{a}}(j|i_{a})}{(P_{C'}^{a})^{d_{a}}(j|i_{a})}.
\label{eq:strict-positive-drift-first-term-3}
\end{align}
Using the convergence in \eqref{eq:limit_for_N(n,d,i,a,j)/N(n,d,i,a)} and noting that $\mathbb{S}_{R}\times \mathcal{A}$ is finite, we get that there exists $N_{2}=N_{2}(\varepsilon)$ such that for all $n\geq N_{2}$, \eqref{eq:strict-positive-drift-first-term-3} is almost surely lower bounded by
\begin{align}
&\sum_{(\underline{d}, \underline{i})\notin \bigcup_{a'=1}^{K} \mathbb{S}_{R, a'}}\ \sum_{a=1}^{K}\ \frac{N(n,\underline{d}, \underline{i}, a)}{n} \ D_{\textsf{KL}}((P_C^{a})^{d_a}(\cdot|i_a) \| (P_{C'}^{a})^{d_a}(j|i_a)) \nonumber\\
&\hspace{3cm}+ \sum_{a=1}^{K}\ \sum_{(\underline{d}, \underline{i})\in \mathbb{S}_{R,a}}\  \frac{N(n,\underline{d}, \underline{i}, a)}{n} \ D_{\textsf{KL}}((P_C^{a})^{d_a}(\cdot|i_a) \| (P_{C'}^{a})^{d_a}(j|i_a)) - \varepsilon.
\label{eq:strict-positive-drift-first-term-4}
\end{align}
Combining \eqref{eq:strict-positive-drift-first-term-2} and \eqref{eq:strict-positive-drift-first-term-3}, we get that 
\begin{align}
\frac{Z^{\pi}_{CC'}(n)}{n} &\geq -2\varepsilon + \sum_{(\underline{d}, \underline{i})\notin \bigcup_{a'=1}^{K} \mathbb{S}_{R, a'}}\ \sum_{a=1}^{K}\ \frac{N(n,\underline{d}, \underline{i}, a)}{n} \ D_{\textsf{KL}}((P_C^{a})^{d_a}(\cdot|i_a) \| (P_{C'}^{a})^{d_a}(j|i_a)) \nonumber\\
&\hspace{3cm}+ \sum_{a=1}^{K}\ \sum_{(\underline{d}, \underline{i})\in \mathbb{S}_{R,a}}\  \frac{N(n,\underline{d}, \underline{i}, a)}{n} \ D_{\textsf{KL}}((P_C^{a})^{d_a}(\cdot|i_a) \| (P_{C'}^{a})^{d_a}(j|i_a))
\label{eq:strict-positive-drift-first-term-5}
\end{align}
for all $n\geq \max\{N_{1}, N_{2}\}$, almost surely.
Using the results in \eqref{eq:lim_N(n,d,i)} and \eqref{eq:liminf-N(n,d,i,a)/n}, we see that the limit infimum of the last two terms in \eqref{eq:strict-positive-drift-first-term-5} is strictly positive, almost surely. Because $\varepsilon>0$ is arbitrary, the desired result follows.


\section{Proof of Lemma \ref{lem:pi_star(L)_in_Pi(epsilon)}}
\label{appndx:proof-of-admissibility-of-policy}

The policy $\pi=\pi^\star(L,\eta,R)$ commits an error if one of the following events is true:
\begin{enumerate}
	\item The policy does not stop in finite time.
	\item The policy stops in finite time and declares an incorrect best arm index.
\end{enumerate}
The event in item $1$ above has zero probability, thanks to Lemma \ref{lem:liminf_Z_{CC'}(n)_strictly_positive}.
Thus, the probability of error of policy $\pi^\star(L,\eta,R)$ may be evaluated as follows: for $C\in \mathcal{C}$, recall that $a^{\star}(C)$ is the index of the best arm in $C$. Then, under the assignment of the TPMs $C$, the error probability of $\pi^\star(L,\eta,R)$ is given by
\begin{align}
	&P^{\pi}_C(\theta(\tau(\pi))\neq a)\nonumber\\
	&=P_C^{\pi}\bigg(\exists \ n\text{ and }a'\neq a\text{ such that }
	\tau(\pi)=n\text{ and } \theta(n)=a'\bigg)\nonumber\\
	&=P_C^{\pi}\bigg(\exists \ n\text{ and }C'\in\textsf{Alt}(C)\text{ such that }
	\tau(\pi)=n\text{ and } \theta(n)=a^{\star}(C')\bigg).\label{eq:P_e_partial_1}
\end{align}
Let
$\mathcal{R}_{a}(n)\coloneqq\{\omega\in \Omega:\tau(\pi)(\omega)=n,\,\theta(n)(\omega)=a\}$, $a\in \mathcal{A}$,
denote the set of all sample paths for which the policy stops at time $n$ and declares $a$ as the index of the best arm. Clearly, $\{\mathcal{R}_{a}(n):a\in \mathcal{A},\ n\geq 0\}$ is a collection of mutually disjoint sets. Therefore, we have
\begin{align}
&P_{C}^{\pi}(\theta(\tau(\pi))\neq a) \nonumber\\
&=P_C^{\pi}\left(\bigcup_{a'\neq a}\,\bigcup_{n=0}^{\infty}\mathcal{R}_{a'}(n)\right)\nonumber\\
&= \sum_{a'\neq a}\ \sum_{n=0}^{\infty}\ P_C^{\pi}(\tau(\pi)=n,\ \theta(n)=a')\nonumber\\
&=  \sum_{C'\in \textsf{Alt}(C)}\ \sum_{n=0}^{\infty}\ P_C^{\pi}(\tau(\pi)=n,\ \theta(n)=a^{\star}(C'))\nonumber\\
&= \sum_{C'\in \textsf{Alt}(C)}\ \sum_{n=0}^{\infty}\ \int_{\mathcal{R}_{a^{\star}(C')}(n)}\,dP_C^{\pi}(\omega)\nonumber\\
&=\sum_{C'\in \textsf{Alt}(C)}\ \sum_{n=0}^{\infty}~\int_{\mathcal{R}_{a^{\star}(C')}(n)}\ \exp(Z_C^{\pi}(n,\omega))\  d(A_{0:n}(\omega),\bar{X}_{0:n}(\omega))\nonumber\\
&=\sum_{C'\in \textsf{Alt}(C)}\ \sum_{n=0}^{\infty}\ \int_{\mathcal{R}_{a^{\star}(C')}(n)}\ \exp({-Z_{C'C}^{\pi}(n,\omega)})\ \exp(Z_{C'}^{\pi}(n,\omega))\ d(A_{0:n}(\omega),\bar{X}_{0:n}(\omega))\nonumber\\
&\stackrel{(a)}{\leq} \sum_{C'\in \textsf{Alt}(C)}\sum_{n=0}^{\infty}\ \int_{\mathcal{R}_{a^{\star}(C')}(n)}\ \frac{1}{L(K-1)(K-1)!}~dP^{\pi}_{C'}(\omega)\nonumber\\
&=\sum_{C'\in \textsf{Alt}(C)}\frac{1}{L(K-1)(K-1)!}~P_{C'}^{\pi}\left(\bigcup_{n=0}^{\infty}\mathcal{R}_{a^{\star}(C')}(n)\right)\nonumber\\
&\leq \frac{1}{L},
\label{eq:error-probability-proof-1}
\end{align}
where $(a)$ above follows by noting that for any $C'\in \textsf{Alt}(C)$, the condition $M_{C'}^{\pi}(n)\geq \log(L(K-1)(K-1)!)$ holds at the stopping time $\tau(\pi)=n$ on the set $R_{a^\star(C')}$. In particular, this implies that $Z^{\pi}_{C'C}(n)\geq \log(L(K-1)(K-1)!)$. Setting $L=1/\epsilon$ in \eqref{eq:error-probability-proof-1} yields the desired result.


\section{Proof of Proposition \ref{prop:Z_{CC'}(n)_has_the_right_drift}}
\label{appndx:proof-of-prop-Z_{CC'}(n)_has_the_right_drift}

{\color{black} We note that for all $C'\in \textsf{Alt}(C)$, under the policy $\pi=\pi_{\textsf{NS}}^\star(L, \eta, R)$, almost surely, 
\begin{align}
\limsup_{n\to\infty} M_{C'}^{\pi}(n)
&= \limsup_{n\to\infty} \min_{C''\in \textsf{Alt}(C')}\ Z_{C'C''}^{\pi}(n)\nonumber\\
&\leq \limsup_{n\to\infty}\ Z_{C'C}^{\pi}(n)\nonumber\\
&= \limsup_{n\to\infty}\ -Z_{CC'}^{\pi}(n)\nonumber\\
&=-\liminf_{n\to\infty}\ Z_{CC'}^{\pi}(n)\nonumber\\
&\leq -\liminf_{n\to\infty}\ M_{C}^{\pi}(n)\nonumber\\*
&<0,
\label{eq:limsup-M(C')}
\end{align}
where the last line above is due to Lemma \ref{lem:liminf_Z_{CC'}(n)_strictly_positive}. 
Furthermore, for any $\bar{C}\neq C$ such that $a^\star(\bar{C})=a^\star(C)$,\footnote{Recall that $a^\star(C)$ denotes the index of the best arm in $C$.} following the exposition in Appendix  \ref{subsec:completing-proof-of-strict-positivity-of-drift} with $C'$ replaced by $\bar{C}$, we get that $\liminf_{n\to\infty} Z_{C\bar{C}}^\pi(n)/n>0$ almost surely, and therefore $\liminf_{n\to\infty} Z_{C\bar{C}}^\pi(n)>0$ almost surely. This implies that $\liminf_{n\to\infty} Z_{CC'}^\pi(n) - Z_{\bar{C}C'}^\pi(n)>0$ almost surely for all $C'\in \textsf{Alt}(C)$, which in turn implies that $\liminf_{n\to\infty} Z_{CC'}^\pi(n) > \limsup_{n\to\infty} Z_{\bar{C}C'}^\pi(n)$ almost surely for all $C'\in \textsf{Alt}(C)$. Noting that $\textsf{Alt}(C)=\textsf{Alt}(\bar{C})$, it follows that
\begin{align}
     \liminf_{n\to\infty} M_C^\pi(n) > \limsup_{n\to\infty} M_{\bar{C}}^\pi(n)\quad  \text{almost surely} 
    \label{eq:MC-is-greater-than-MCbar}
\end{align}
for all $\bar{C}\neq C$ such that $a^\star(\bar{C})=a^\star(C)$. Combining \eqref{eq:MC-is-greater-than-MCbar} and \eqref{eq:limsup-M(C')}, we get that under~$\pi=\pi_{\textsf{NS}}^{\star}(L, \eta, R)$, 
\begin{equation}
\bar{C}(n) = C \quad \text{for all }n\text{ large}, \quad \text{almost surely},
\end{equation}
when the underlying assignment of the TPMs is $C$}, from which we may deduce that under $\pi=\pi_{\textsf{NS}}^{\star}(L, \eta, R)$,
\begin{align}
&\lim_{n\to\infty}\ P^{\pi}(A_{n}=a|A_{0:n-1}, \{(\underline{d}(s), \underline{i}(s)):K\leq s\leq n-1\},  \underline{d}(n)=\underline{d}, \underline{i}(n)=\underline{i}
)\nonumber\\
&=\lim_{n\to\infty}\ \lambda_{\eta, R, \bar{C}(n)}(a|\underline{d},\underline{i})\nonumber\\
&=\lambda_{\eta,R, C}(a|\underline{d},\underline{i})
\label{eq:lim-prob-of-selecting-arm-under-policy}
\end{align}
almost surely.
This shows that $\pi_{\textsf{NS}}^{\star}(L, \eta, R)$ {\em eventually} makes the process $\{(\underline{d}(t), \underline{i}(t)):t\geq K\}$ ergodic with $\nu_{\eta,R,C}$ as its ergodic state-action occupancy measure. As a consequence, for all $(\underline{d}, \underline{i})\in \mathbb{S}_R$ and $a\in \mathcal{A}$, it follows that 
\begin{equation}
\lim_{n\to\infty}\ \frac{N(n,\underline{d},\underline{i},a)}{n} = \nu_{\eta, R, C}(\underline{d},\underline{i},a) \ \text{almost surely}.
\label{eq:limit_N(n,d,i,a)/N(n,d,i)}
\end{equation}
Therefore, under $\pi=\pi_{\textsf{NS}}^{\star}(L, \eta, R)$, we have
\begin{align}
\lim_{n\to\infty}\ \frac{Z_{CC'}^{\pi}(n)}{n} 
&= \lim_{n\to\infty}\  \sum_{(\underline{d},\underline{i})\in\mathbb{S}_{R}}\ \sum_{a=1}^{K}\  \sum_{j\in\mathcal{S}}\frac{N(n,\underline{d},\underline{i},a,j)}{n}\,\log\frac{(P_{C}^a)^{d_a}(j|i_a)}{(P_{C'}^{a})^{d_{a}}(j|i_a)}\nonumber\\
&=\sum_{(\underline{d},\underline{i})\in\mathbb{S}_{R}}\ \sum_{a=1}^{K}\  \sum_{j\in\mathcal{S}}\ \lim_{n\to\infty}\ \frac{N(n,\underline{d},\underline{i},a,j)}{n}\,\log\frac{(P_{C}^a)^{d_a}(j|i_a)}{(P_{C'}^{a})^{d_{a}}(j|i_a)}\nonumber\\
&\stackrel{(a)}{=}\sum_{(\underline{d},\underline{i})\in\mathbb{S}_{R}}\ \sum_{a=1}^{K}\  \sum_{j\in\mathcal{S}}\ \nu_{\eta,R,C}(\underline{d}, \underline{i}, a)\cdot (P_{C}^{a})^{d_{a}}(j|i_{a})\cdot \log\frac{(P_{C}^a)^{d_a}(j|i_a)}{(P_{C'}^{a})^{d_{a}}(j|i_a)}\nonumber\\
&=\sum_{(\underline{d},\underline{i})\in\mathbb{S}_{R}}\ \sum_{a=1}^{K}\  \nu_{\eta,R,C}(\underline{d}, \underline{i}, a) \, D((P_{C}^{a})^{d_{a}}(\cdot|i_{a}) \| (P_{C'}^{a})^{d_{a}}(\cdot | i_{a})) \quad \text{almost surely},
\label{eq:proof-of-correct-drift-for-z_{CC'}-1}
\end{align}
where $(a)$ follows from \eqref{eq:limit_for_N(n,d,i,a,j)/N(n,d,i,a)}. Eq.~\eqref{eq:limit_M_h(n)/n} is then immediate from~\eqref{eq:proof-of-correct-drift-for-z_{CC'}-1}.


\section{Proof of Lemma \ref{Lemma:stopping_time_of_policy_goes_to_infinity}}
\label{appndx:proof-of-lemma-stopping-time-goes-to-infty}

Because $\pi=\pi^{\star}(L, \eta, R)$ selects arm $1$ at time $t=0$, arm $2$ at time $t=1$, etc., and arm $K$ at time $t=K-1$, in order to prove the lemma, it suffices to prove that for all $m \geq K$,
\begin{equation}
\lim_{L\to\infty}P_C^{\pi}(\tau(\pi)\leq m)=0.
\end{equation}
Fix an arbitrary $m\geq K$. Then,
\begin{align}
	&\limsup_{L\to\infty}\,P_C^{\pi}(\tau(\pi)\leq m)\nonumber\\
	&=\limsup_{L\to\infty}\,P_C^{\pi}\bigg(\exists~K\leq n\leq m \text{ and }C'\in\mathcal{C}
	\text{ such that }M_{C'}^{\pi}(n)
	\geq \log(L(K-1)(K-1)!)\bigg)\nonumber\\
	&\leq \limsup_{L\to\infty}\sum_{C'\in\mathcal{C}}\ \sum_{n=K}^{m}\ P_{C}^{\pi}(M_{C'}^{\pi}(n)
	\geq \log(L(K-1)(K-1)!))\nonumber\\
	&\leq \limsup_{L\to\infty}\ \frac{1}{\log(L(K-1)(K-1)!)}\ \sum_{C'\in\mathcal{C}}\ \sum_{n=K}^{m}\ \mathbb{E}_{C}^{\pi}[M_{C'}^{\pi}(n)],\label{eq:stop_time_goes_to_infty_1}
\end{align}
where the first and the second inequalities above follow from the union bound and Markov's inequality respectively. We now show that for each $n\in\{K,\ldots,m\}$, the expectation term inside the summation in \eqref{eq:stop_time_goes_to_infty_1} is finite. This will then imply that the limit supremum on the right-hand side of \eqref{eq:stop_time_goes_to_infty_1} is equal to $0$, thus proving the desired result.

Note that
\begin{align}
M_{C'}^{\pi}(n)=\min_{\tilde{C}\in \textsf{Alt}(C')}Z_{C'\tilde{C}}^{\pi}(n)\leq Z_{C'\tilde{C}}^{\pi}(n)\text{ for all }\tilde{C}\in \textsf{Alt}(C').
\label{eq:mod_glr_upper_bounded_by_glr}
\end{align} 
Fix an arbitrary $\tilde{C}\in \textsf{Alt}(C')$ and recall that 
\begin{align}
	Z_{C'\tilde{C}}^{\pi}(n)&=\sum_{a=1}^{K} \log \frac{P_{C'}^{\pi}(X_{a-1}^{a})}{P_{\tilde{C}}^{\pi}(X_{a-1}^{a})} + \sum_{(\underline{d}, \underline{i})\in \mathbb{S}_{R}}\ \sum_{a=1}^{K}\ \sum_{j\in \mathcal{S}}\ N(n, \underline{d}, \underline{i}, a, j) \ \log \frac{(P_{C'}^{a})^{d_{a}}(j|i_{a})}{(P_{\tilde{C}}^{a})^{d_{a}}(j|i_{a})}.
\label{eq:Z_{h_tilde_h'}(n)_upper_bound_1}
\end{align}
Because each row of $(P_{C'}^{a})^{d}$ is mutually absolutely continuous with the corresponding row of $(P_{\tilde{C}}^{a})^{d}$ for all $d\geq 1$,
we may upper bound the second term in \eqref{eq:Z_{h_tilde_h'}(n)_upper_bound_1} as
\begin{align}
\sum_{(\underline{d}, \underline{i})\in \mathbb{S}_{R}}\ \sum_{a=1}^{K}\ \sum_{j\in \mathcal{S}}\ N(n, \underline{d}, \underline{i}, a, j) \ \log \frac{(P_{C'}^{a})^{d_{a}}(j|i_{a})}{(P_{\tilde{C}}^{a})^{d_{a}}(j|i_{a})} &\leq A \cdot \left(\sum_{(\underline{d}, \underline{i})\in \mathbb{S}_{R}}\ \sum_{a=1}^{K}\ \sum_{j\in \mathcal{S}}\ N(n, \underline{d}, \underline{i}, a, j)\right)\nonumber\\
& = A\cdot (n-K+1)\quad \text{almost surely},
\label{eq:Z_{h_tilde_h'}(n)_upper_bound_2}
\end{align}
where
\begin{equation}
A = \max\left\lbrace  \log\frac{P_a^{d}(j|i)}{P_{a'}^{d}(j|i)}:\ P_a^{d}(j|i)\neq 0, \ P_{a'}^{d}(j|i) \neq 0, \ d\in\mathbb{N},\ i,j\in\mathcal{S},\ a,a'\in \mathcal{A}\right\rbrace<\infty.
\label{eq:A}
\end{equation}
Furthermore, suppose that $X_{0}^{a}\sim \phi$ for all $a\in \mathcal{A}$, where $\phi$ is a probability distribution on $\mathcal{S}$ that is independent of $a$ and the underlying assignment of the TPMs $C$. Without loss of generality, let $\phi(i)>0$ for all $i\in \mathcal{S}$. Then, the first term in \eqref{eq:Z_{h_tilde_h'}(n)_upper_bound_1} may be upper bounded as
\begin{align}
\sum_{a=1}^{K} \log \frac{P_{C'}^{\pi}(X_{a-1}^{a})}{P_{\tilde{C}}^{\pi}(X_{a-1}^{a})} &= \log \frac{P_{C'}^{\pi}(X_{0}^{1})}{P_{\tilde{C}}^{\pi}(X_{0}^{1})} + \sum_{a=2}^{K} \log \frac{P_{C'}^{\pi}(X_{a-1}^{a})}{P_{\tilde{C}}^{\pi}(X_{a-1}^{a})}\nonumber\\
&=\sum_{a=2}^{K}\ \sum_{j\in \mathcal{S}}\ \mathbb{I}_{\{X_{a-1}^{a}=j\}}\ \log \frac{P_{C'}^{\pi}(X_{a-1}^{a}=j)}{P_{\tilde{C}}^{\pi}(X_{a-1}^{a}=j)}\nonumber\\
&=\sum_{a=2}^{K} \ \sum_{j\in \mathcal{S}}\ \mathbb{I}_{\{X_{a-1}^{a}=j\}}\ \log \frac{\sum_{i\in \mathcal{S}}\ \phi(i)\cdot (P_{C'}^{a})^{a-1}(j|i)}{\sum_{i\in \mathcal{S}}\ \phi(i)\cdot (P_{\tilde{C}}^{a})^{a-1}(j|i)}\nonumber\\
&\stackrel{(a)}{\leq} \sum_{a=2}^{K}\ \sum_{j\in \mathcal{S}}\ \mathbb{I}_{\{X_{a-1}^{a}=j\}} \ \sum_{i\in \mathcal{S}}\ \frac{\phi(i)\,(P_{C'}^{a})^{a-1}(j|i)}{\sum_{i'\in \mathcal{S}} \phi(i')\,(P_{C'}^{a})^{a-1}(j| i')} \, \log \frac{(P_{C'}^{a})^{a-1}(j|i)}{(P_{\tilde{C}}^{a})^{a-1}(j|i)}\nonumber\\
&\leq A\, (K-1) \quad \text{almost surely},
\label{eq:Z_{h_tilde_h'}(n)_upper_bound_3}
\end{align}
where $(a)$ above follows from the log-sum inequality \cite[Theorem 2.7.1]{thomas2006elements}. Combining \eqref{eq:Z_{h_tilde_h'}(n)_upper_bound_2} and \eqref{eq:Z_{h_tilde_h'}(n)_upper_bound_3}, we get
\begin{equation}
Z_{C'\tilde{C}}^{\pi}(n) \leq A\cdot n \quad \text{almost surely},
\end{equation}
from which it follows that $\mathbb{E}_{C}^{\pi}[Z_{C'\tilde{C}}^{\pi}(n)]\leq A\cdot n$.


\section{Proof of Lemma \ref{Lemma:almost_sure_upper_bound_for_policy_pi_star}}
\label{appndx:proof-of-almost-sure-upper-bound}

By the definition of $\tau(\pi)$, we know that under the assignment of the TPMs $C$, $$ M_{C}^{\pi}(\tau(\pi)-1) < \log (L(K-1)(K-1)!).$$ Therefore, almost surely,
\begin{align}
1 &= \limsup_{L\to\infty}\ \frac{\log (L(K-1)(K-1)!)}{\log L}\nonumber\\
&\geq \limsup_{L\to\infty}\ \frac{M_{C}^{\pi}(\tau(\pi)-1)}{\log L}\nonumber\\
&=\limsup_{L\to\infty}\ \frac{M_{C}^{\pi}(\tau(\pi)-1)}{\tau(\pi)-1} \cdot \frac{\tau(\pi)-1}{\log L}\nonumber\\
&\geq \left(\limsup_{L\to\infty}\ \frac{\tau(\pi)-1}{\log L}\right)\cdot \left(\min_{C'\in \textsf{Alt}(C)}\ \sum_{(\underline{d},\underline{i})\in\mathbb{S}_{R}}\ \sum_{a=1}^{K}\ \nu_{\eta,R,C}(\underline{d},\underline{i},a) \ D((P_{C}^{a})^{d_{a}}(\cdot|i_{a}) \| (P_{C'}^{a})^{d_{a}}(\cdot | i_{a}))\right),
\end{align}
where the last line above is due to \eqref{eq:M_h(N(pi))/N(pi)_has_almost_correct_drift} and the fact that the increment $M_{C}^{\pi}(n)-M_{C}^{\pi}(n-1)$ is bounded for all $n\geq K$. We then note that
\begin{align}
&\min_{C'\in \textsf{Alt}(C)}\ \sum_{(\underline{d},\underline{i})\in\mathbb{S}_{R}}\ \sum_{a=1}^{K}\ \nu_{\eta,R,C}(\underline{d},\underline{i},a) \ D((P_{C}^{a})^{d_{a}}(\cdot|i_{a}) \| (P_{C'}^{a})^{d_{a}}(\cdot | i_{a})) \nonumber\\
&\geq \eta \ \min_{C'\in \textsf{Alt}(C)}\ \left(\sum_{(\underline{d},\underline{i})\in\mathbb{S}_{R}}\ \sum_{a=1}^{K}\ \nu_{C,R}^{\textsf{unif}}(\underline{d},\underline{i},a) \ D((P_{C}^{a})^{d_{a}}(\cdot|i_{a}) \| (P_{C'}^{a})^{d_{a}}(\cdot | i_{a}))\right) \nonumber\\
&\hspace{3cm}+ (1-\eta)\ \min_{C'\in \textsf{Alt}(C)}\ \left(\sum_{(\underline{d},\underline{i})\in\mathbb{S}_{R}}\ \sum_{a=1}^{K}\ \nu_{C,R}^{\star}(\underline{d},\underline{i},a) \ D((P_{C}^{a})^{d_{a}}(\cdot|i_{a}) \| (P_{C'}^{a})^{d_{a}}(\cdot | i_{a}))\right)\nonumber\\
&\geq \eta \ T_{R}^{\textsf{unif}}(C) + (1-\eta)\ T_{R}^{\star}(C),
\end{align}
thus establishing the desired result.


\section{Proof of Proposition \ref{prop:upper_bound}}
\label{appndx:proof_of_upper_bound}

We prove here that the family $\{\tau(\pi)/\log L:L>1\}$, where $\pi=\pi^{\star}(L, \eta, R)$, is uniformly integrable. Towards this, we show that
\begin{equation}
\limsup_{L\to\infty}\ \mathbb{E}_{C}^{\pi}\bigg[\bigg(\frac{\tau(\pi)}{\log L}\bigg)^{2}\bigg]<\infty.
\label{eq:sufficient-condition-for-uniform-integrability}
\end{equation}
Then, \cite[Lemma 3, pp. 227]{shiryaev2016probability} implies the desired uniform integrability result from \eqref{eq:sufficient-condition-for-uniform-integrability} (by using $G(t)=t^2$ in \cite[Lemma 3, pp. 227]{shiryaev2016probability}). Let 
\begin{align}
\psi(L) \coloneqq \frac{\log(L(K-1)(K-1)!)}{\log L},
\end{align}
and let $\pi^\star_C=\pi^\star_C(L,\eta,R)$ denote the version of the policy {\sc $R$-DCR-BAI} that stops only when the event $$M_{C}^{\pi^{\star}_{C}}(n)\geq \log(L(K-1)(K-1)!)$$ occurs. Clearly, $\tau(\pi^\star_C)\geq \tau(\pi)$ almost surely. Then,
\begin{align}
&\limsup_{L\to\infty}\ \mathbb{E}_C^{\pi}\bigg[\bigg(\frac{\tau(\pi)}{\log L}\bigg)^{2}\bigg]\nonumber\\
&=\limsup_{L\to\infty}\int_{0}^{\infty}P_C^{\pi}\bigg(\left(\frac{\tau(\pi)}{\log L}\right)^{2}>x\bigg)\,dx\nonumber\\
&= \limsup_{L\to\infty}\int_{0}^{\infty}P_C^{\pi}\bigg(\tau(\pi) > (\sqrt{x})({\log L})\bigg)\,dx\nonumber\\
&\leq \limsup_{L\to\infty}\int_{0}^{\infty}P_C^{\pi}\bigg({\tau(\pi^\star_C)}> (\sqrt{x})({\log L})\bigg)\,dx\nonumber\\
&\stackrel{(a)}{\leq} \limsup_{L\to\infty}\bigg\lbrace \psi(L)+\int_{\psi (L)}^{\infty}P_C^{\pi}\bigg({\tau(\pi^\star_C)}> (\sqrt{x})({\log L})\bigg)\,dx\bigg\rbrace\nonumber\\
&\leq \limsup_{L\to\infty}\ \psi(L)+\limsup_{L\to\infty}\sum_{n\geq \sqrt{\psi(L)}\, \log L }^{\infty} \frac{2n+1}{(\log L)^{2}}\ P_C^{\pi}(\tau(\pi_{C}^{\star})>n)\nonumber\\
&\leq 1+\limsup_{L\to\infty}\sum_{n\geq \sqrt{\psi(L)}\, \log L }^{\infty} \frac{2n+1}{(\log L)^{2}}\ P_C^{\pi}(M_C^{\pi}(n)<\log(L(K-1)(K-1)!)),\label{eq:uniform_integrability_1}
\end{align}
where $(a)$ above follows by upper bounding the probability term by $1$ for all $x\leq \psi(L)$. Below, we show that $P_C^{\pi}(M_C^{\pi}(n)<\log(L(K-1)(K-1)!))$ is $O(1/n^{3})$, which implies that the infinite sum in \eqref{eq:uniform_integrability_1} is finite. This will then prove that the right-hand side of \eqref{eq:uniform_integrability_1} is finite. Note that
\begin{align}
P_C^{\pi}(M_C^{\pi}(n)<\log(L(K-1)(K-1)!))
&=P_C^{\pi}\left(\min_{C'\in \textsf{Alt}(C)}Z_{CC'}^{\pi}(n)<\log(L(K-1)(K-1)!)\right)\nonumber\\
&\leq \sum_{C'\in \textsf{Alt}(C)}\ P_C^{\pi}\left(Z_{CC'}^{\pi}(n)<\log(L(K-1)(K-1)!)\right),
\label{eq:exp_bound_1}
\end{align}
where the last line above follows from the union bound. 

We now show that each term inside the summation in \eqref{eq:exp_bound_1} is $O(1/n^{3})$. Fix $C'\in \textsf{Alt}(C)$ and observe that
\begin{align}
&P_C^{\pi}\left(Z_{CC'}^{\pi}(n)<\log(L(K-1)(K-1)!)\right)\nonumber\\
&=P_C^{\pi}\left(\frac{Z_{CC'}^{\pi}(n)}{n}<\frac{\log(L(K-1)(K-1)!)}{n}\right)\nonumber\\
&=P_C^{\pi}\bigg(\frac{1}{n}\ \sum_{a=1}^{K} \ \log \frac{P_{C}^{\pi}(X_{a-1}^{a})}{P_{C'}^{\pi}(X_{a-1}^{a})}  \nonumber\\
&\hspace{2cm}+\sum_{(\underline{d}, \underline{i})\in \mathbb{S}_{R}}\ \sum_{a=1}^{K}\ \sum_{j\in \mathcal{S}}\ \frac{N(n, \underline{d}, \underline{i}, a, j)}{n}\ \log \frac{(P_{C}^{a})^{d_{a}}(j|i_{a})}{(P_{C'}^{a})^{d_{a}}(j|i_{a})}<\frac{\log(L(K-1)(K-1)!)}{n}\bigg)\nonumber\\
&\leq P_C^{\pi}\left(\frac{1}{n}\ \sum_{a=1}^{K} \ \log \frac{P_{C}^{\pi}(X_{a-1}^{a})}{P_{C'}^{\pi}(X_{a-1}^{a})}< -\varepsilon\right) \label{exp-upper-bound-prob-term-1}\\
&+ P_C^{\pi}\bigg(\sum_{(\underline{d}, \underline{i})\in \mathbb{S}_{R}}\ \sum_{a=1}^{K}\ \sum_{j\in \mathcal{S}}\ \frac{N(n, \underline{d}, \underline{i}, a, j)}{n}\ \log \frac{(P_{C}^{a})^{d_{a}}(j|i_{a})}{(P_{C'}^{a})^{d_{a}}(j|i_{a})} \nonumber\\
&\hspace{3cm}- \sum_{(\underline{d}, \underline{i})\in \mathbb{S}_{R}}\ \sum_{a=1}^{K}\ \frac{N(n, \underline{d}, \underline{i}, a)}{n}\, D_{\textsf{KL}}((P_{C}^{a})^{d_{a}}(\cdot|i_{a})\| (P_{C'}^{a})^{d_{a}}(\cdot|i_{a})) < -\varepsilon\bigg) \label{exp-upper-bound-prob-term-2}\\
&+P_C^{\pi}\bigg( \sum_{(\underline{d}, \underline{i})\in \mathbb{S}_{R}}\ \sum_{a=1}^{K}\ \frac{N(n, \underline{d}, \underline{i}, a)}{n}\, D_{\textsf{KL}}((P_{C}^{a})^{d_{a}}(\cdot|i_{a})\| (P_{C'}^{a})^{d_{a}}(\cdot|i_{a}))  - 2\varepsilon < \frac{\log(L(K-1)(K-1)!)}{n}\bigg)  \label{exp-upper-bound-prob-term-3}
\end{align}
for all $\varepsilon>0$. Fixing $\varepsilon$, we handle \eqref{exp-upper-bound-prob-term-1}-\eqref{exp-upper-bound-prob-term-3} individually. We also show how to choose $\varepsilon$ (later in \eqref{eq:choice-of-varepsilon}).

\begin{enumerate}
\item Handling \eqref{exp-upper-bound-prob-term-1}: This  term is equal to $0$ for all sufficiently large values of $n$ because the left hand side inside the probability term converges almost surely to $0$, whereas the right hand side is negative.

\item Handling \eqref{exp-upper-bound-prob-term-2}: This term may be expressed as
\begin{align}
P_{C}^{\pi}\bigg(\sum_{(\underline{d}, \underline{i})\in \mathbb{S}_{R}}\ \sum_{a=1}^{K}\ \sum_{j\in \mathcal{S}} \ \bigg[\frac{N(n, \underline{d}, \underline{i}, a, j)}{n}-\frac{N(n, \underline{d}, \underline{i}, a)}{n}\ (P_{C}^{a})^{d_a}(j|i_a)\bigg] \ \log \frac{(P_{C}^{a})^{d_{a}}(j|i_{a})}{(P_{C'}^{a})^{d_{a}}(j|i_{a})} < -\varepsilon\bigg)
\end{align}
We note that $$M_n \coloneqq \left\lbrace\sum_{(\underline{d}, \underline{i})\in \mathbb{S}_{R}}\ \sum_{a=1}^{K}\ \sum_{j\in \mathcal{S}} \ \bigg[N(n, \underline{d}, \underline{i}, a, j)-N(n, \underline{d}, \underline{i}, a)\ (P_{C}^{a})^{d_a}(j|i_a)\bigg] \ \log \frac{(P_{C}^{a})^{d_{a}}(j|i_{a})}{(P_{C'}^{a})^{d_{a}}(j|i_{a})}\right\rbrace_{n\geq K}$$ is a martingale. Indeed, because each row of $(P_{C}^{a})^{d}$ is mutually absolutely continuous with respect to the corresponding row of $(P_{C'}^{a})^{d}$ for all $d\geq 1$, 
\begin{align}
&\mathbb{E}_{C}^\pi\bigg[\sum_{(\underline{d}, \underline{i})\in \mathbb{S}_{R}}\ \sum_{a=1}^{K} \ \sum_{j\in \mathcal{S}}\ \left(N(n, \underline{d}, \underline{i}, h, j)-N(n, \underline{d}, \underline{i}, a)\ (P_{C}^{a})^{d_a}(j|i_a)\right)\ \log \frac{(P_{C}^{a})^{d_{a}}(j|i_{a})}{(P_{C'}^{a})^{d_{a}}(j|i_{a})} \bigg |\ A_{0:n-1}, \bar{X}_{0:n-1}\bigg]\nonumber\\
	&=\sum_{(\underline{d}, \underline{i})\in \mathbb{S}_{R}}\ \sum_{a=1}^{K} \ \sum_{j\in \mathcal{S}}\ \left(N(n-1, \underline{d}, \underline{i}, a, j)-N(n-1, \underline{d}, \underline{i}, a)\ (P_{C}^{a})^{d_a}(j|i_a)\right)\ \log \frac{(P_{C}^{a})^{d_{a}}(j|i_{a})}{(P_{C'}^{a})^{d_{a}}(j|i_{a})}\nonumber\\
	&+\sum_{(\underline{d}, \underline{i})\in \mathbb{S}_{R}}\ \sum_{a=1}^{K}\ \sum_{j\in \mathcal{S}}\ \bigg[ \mathbb{I}_{\{\underline{d}(n)=\underline{d},~\underline{i}(n)=\underline{i}\}}\ \bigg(P_{C}^\pi(A_n=a,\ \bar{X}_n=j|A_{0:n-1}, \bar{X}_{0:n-1})\nonumber\\
	&\hspace{5.5cm}-P_{C}^\pi(A_n=a|A_{0:n-1}, \bar{X}_{0:n-1})\ (P_{C}^{a})^{d_a}(j|i_a)\bigg)\ \log \frac{(P_{C}^{a})^{d_{a}}(j|i_{a})}{(P_{C'}^{a})^{d_{a}}(j|i_{a})}\bigg]\nonumber\\
		&=\sum_{(\underline{d}, \underline{i})\in \mathbb{S}_{R}}\ \sum_{a=1}^{K} \ \sum_{j\in \mathcal{S}}\ \left(N(n-1, \underline{d}, \underline{i}, a, j)-N(n-1, \underline{d}, \underline{i}, a)\ (P_{C}^{a})^{d_a}(j|i_a)\right)\ \log \frac{(P_{C}^{a})^{d_{a}}(j|i_{a})}{(P_{C'}^{a})^{d_{a}}(j|i_{a})},
\end{align}
where the last line follows by noting that when $(\underline{d}(n), \underline{i}(n))=(\underline{d}, \underline{i})$, under the assignment of the TPMs $C$, $$ P_{C}^\pi(\bar{X}_n=j|A_{n}=a,A_{0:n-1}, \bar{X}_{0:n-1})=(P_{C}^{a})^{d_a}(j|i_a).$$ Further, the above martingale is bounded, and its quadratic variation $\langle M_n \rangle$ satisfies
\begin{align}
& \langle M_n\rangle\nonumber\\
& \coloneqq \sum_{t=K}^n \ \mathbb{E}_{C}^\pi\bigg[\bigg(\sum_{(\underline{d}, \underline{i})\in \mathbb{S}_{R}}\ \sum_{a=1}^{K}\ \sum_{j\in \mathcal{S}}\ \mathbb{I}_{\{\underline{d}(t)=\underline{d},\ \underline{i}(t)=\underline{i},\ A_t=a\}}\ \left(\mathbb{I}_{\{\bar{X}_t=j\}}-(P_{C}^{a})^{d_a}(j|i_a)\right)\ \log \frac{(P_{C}^{a})^{d_{a}}(j|i_{a})}{(P_{C'}^{a})^{d_{a}}(j|i_{a})}\bigg)^2 \bigg|\mathcal{F}_{t}\bigg]\nonumber\\
	&\leq \sum_{t=K}^n \ \mathbb{E}_{C}^\pi\bigg[\sum_{(\underline{d}, \underline{i})\in \mathbb{S}_{R}}\ \sum_{a=1}^{K}\ \sum_{j\in \mathcal{S}}\ \mathbb{I}_{\{\underline{d}(t)=\underline{d},\ \underline{i}(t)=\underline{i},\ A_t=a\}}\ \left(\mathbb{I}_{\{\bar{X}_t=j\}}-(P_{C}^{a})^{d_a}(j|i_a)\right)^{2}\ \bigg(\log \frac{(P_{C}^{a})^{d_{a}}(j|i_{a})}{(P_{C'}^{a})^{d_{a}}(j|i_{a})}\bigg)^2 \bigg|\mathcal{F}_{t}\bigg]\nonumber\\
	&\stackrel{(a)}{\leq} 4 A^{2}\ \sum_{t=K}^n \ \mathbb{E}_{C}^\pi\bigg[\sum_{(\underline{d}, \underline{i})\in \mathbb{S}_{R}}\ \sum_{a=1}^{K}\ \sum_{j\in \mathcal{S}}\ \mathbb{I}_{\{\underline{d}(t)=\underline{d},~\underline{i}(t)=\underline{i},~A_t=a\}}\bigg| \mathcal{F}_{t}\bigg]\nonumber\\
	&\leq n\ (4A^{2} |\mathcal{S}|)
	\label{eq:exp_bound_4}
\end{align}
almost surely, where $A$ above is as defined in \eqref{eq:A}. We then have
\begin{align}
	& P_{C}^\pi\left(\sum_{(\underline{d}, \underline{i})\in \mathbb{S}_{R}}\ \sum_{a=1}^{K}\ \sum_{j\in \mathcal{S}}\ \left(N(n, \underline{d}, \underline{i}, a, j)-N(n, \underline{d}, \underline{i}, a)\ (P_{C}^{a})^{d_a}(j|i_a)\right)\ \log \frac{(P_{C}^{a})^{d_{a}}(j|i_{a})}{(P_{C'}^{a})^{d_{a}}(j|i_{a})} < -n\varepsilon \right)\nonumber\\
	&\leq P_{C}^\pi\left(\left\lvert \sum_{(\underline{d}, \underline{i})\in \mathbb{S}_{R}}\ \sum_{a=1}^{K}\ \sum_{j\in \mathcal{S}}\ \left(N(n, \underline{d}, \underline{i}, a, j)-N(n, \underline{d}, \underline{i}, a)\ (P_{C}^{a})^{d_a}(j|i_a)\right)\ \log \frac{(P_{C}^{a})^{d_{a}}(j|i_{a})}{(P_{C'}^{a})^{d_{a}}(j|i_{a})} \right\rvert > n\varepsilon \right)\nonumber\\
	&\leq P^\pi\left(\sup_{K\leq t\leq n}~\left\lvert\sum_{(\underline{d}, \underline{i})\in \mathbb{S}_{R}}\ \sum_{a=1}^{K}\ \sum_{j\in \mathcal{S}}\ \left(N(n, \underline{d}, \underline{i}, a, j)-N(n, \underline{d}, \underline{i}, a)\ (P_{C}^{a})^{d_a}(j|i_a)\right)\ \log \frac{(P_{C}^{a})^{d_{a}}(j|i_{a})}{(P_{C'}^{a})^{d_{a}}(j|i_{a})}\right\rvert > n\varepsilon \right)\nonumber\\
	&\stackrel{(a)}{\leq}  \frac{\mathbb{E}_{C}^\pi\left[\left(\sup\limits_{K\leq t\leq n}\ \left\lvert\sum\limits_{(\underline{d}, \underline{i})\in \mathbb{S}_{R}}\ \sum\limits_{a=1}^{K}\ \sum\limits_{j\in \mathcal{S}}\ \left(N(n, \underline{d}, \underline{i}, a, j)-N(n, \underline{d}, \underline{i}, a)\ (P_{C}^{a})^{d_a}(j|i_a)\right)\ \log \frac{(P_{C}^{a})^{d_{a}}(j|i_{a})}{(P_{C'}^{a})^{d_{a}}(j|i_{a})}\right\rvert\right)^6\right]}{n^6\ \varepsilon^6}\nonumber\\
	&\stackrel{(b)}{\leq} \frac{B}{n^6\ \varepsilon^6}\ \mathbb{E}_{C}^\pi[|\langle M_n \rangle|^{3}]\nonumber\\
	&\stackrel{(c)}{\leq }\frac{B}{n^6\ \varepsilon^6}\cdot n^{3}\cdot \left(4A^{2}|\mathcal{S}|\right)^{3}\nonumber\\
	&= \frac{A'}{n^{3}},
	\label{eq:exp_bound_5}
\end{align}
where $(a)$ above is due to Markov's inequality, $(b)$ is due to Burkholder's inequality \cite[pp. 414]{chow2012probability}, and $(c)$ follows from \eqref{eq:exp_bound_4}. We have thus shown that \eqref{exp-upper-bound-prob-term-2} is $O(1/n^3)$.

\item Handling \eqref{exp-upper-bound-prob-term-3}: 
Observe that \eqref{exp-upper-bound-prob-term-3} may be upper bounded by
\begin{equation}
P_C^{\pi}\bigg( \frac{N(n, \underline{d}, \underline{i}, a)}{n}\, D_{\textsf{KL}}((P_{C}^{a})^{d_{a}}(\cdot|i_{a})\| (P_{C'}^{a})^{d_{a}}(\cdot|i_{a}))  - 2\varepsilon < \frac{\log(L(K-1)(K-1)!)}{n}\bigg)
\label{eq:O(1/n3)_1}
\end{equation}
for all $(\underline{d}, \underline{i})\in \mathbb{S}_{R}$ and $a\in \mathcal{A}$. Fix $\underline{d}=\underline{d}^{\star}=(K, K-1, \ldots, 1)$, $\underline{i}=\underline{i}^{\star}\in \mathcal{S}^{K}$, and $a\in \mathcal{A}$. Using the convergence in \eqref{eq:limit_N(n,d,i,a)/N(n,d,i)}, we get that for every $\epsilon'>0$, under the policy $\pi_{\textsf{NS}}^{\star}(L, \eta, R)$,
\begin{equation}
\frac{N(n, \underline{d}^{\star}, \underline{i}^{\star}, a)}{n} \geq \nu_{\eta,R,C}(\underline{d}^{\star}, \underline{i}^{\star},a)(1-\epsilon')
\end{equation}
for all $n$ large, almost surely. Leveraging this, we define 
\begin{equation}
E_{n} \coloneqq \left\lbrace \omega\in \Omega: \frac{N(n, \underline{d}^{\star}, \underline{i}^{\star}, a, \omega)}{n} \geq   \nu_{\eta,R,C}(\underline{d}^{\star}, \underline{i}^{\star},a)(1-\epsilon') \right\rbrace,
\label{eq:event-En}
\end{equation}
and write \eqref{eq:O(1/n3)_1} as the sum of two terms, one of which is 
\begin{align}
& P_C^{\pi}\bigg( E_{n} \bigcap \left\lbrace\frac{N(n, \underline{d}^{\star}, \underline{i}^{\star}, a)}{n}\, D_{\textsf{KL}}((P_{C}^{a})^{d^{\star}_{a}}(\cdot|i^{\star}_{a})\| (P_{C'}^{a})^{d^{\star}_{a}}(\cdot|i^{\star}_{a}))  - 2\varepsilon < \frac{\log(L(K-1)(K-1)!)}{n}\right\rbrace\bigg)\nonumber\\
& \leq P_{C}^{\pi} \left( (1-\epsilon')\ \nu_{\eta,R,C}(\underline{d}^{\star}, \underline{i}^{\star},a)\ D_{\textsf{KL}}((P_{C}^{a})^{d^{\star}_{a}}(\cdot|i^{\star}_{a})\| (P_{C'}^{a})^{d^{\star}_{a}}(\cdot|i^{\star}_{a})) - 2\varepsilon <  \frac{\log(L(K-1)(K-1)!)}{n}\right),
\label{eq:term-1}
\end{align}
and the other is
\begin{align}
& P_C^{\pi}\bigg( E_{n}^{c} \bigcap \left\lbrace\frac{N(n, \underline{d}^{\star}, \underline{i}^{\star}, a)}{n}\, D_{\textsf{KL}}((P_{C}^{a})^{d^{\star}_{a}}(\cdot|i^{\star}_{a})\| (P_{C'}^{a})^{d^{\star}_{a}}(\cdot|i^{\star}_{a}))  - 2\varepsilon < \frac{\log(L(K-1)(K-1)!)}{n}\right\rbrace\bigg)\nonumber\\
& \leq P_{C}^{\pi}\left(\frac{N(n, \underline{d}^{\star}, \underline{i}^{\star}, a)}{n} < \nu_{\eta,R,C}(\underline{d}^{\star}, \underline{i}^{\star},a)(1-\epsilon')\right).
\label{eq:term-2}
\end{align}
We shall see how to choose $\epsilon'$ (later in \eqref{eq:choice-of-epsilon-prime}). Choosing $\varepsilon$ such that
\begin{equation}
    (1-\epsilon')\ \nu_{\eta,R,C}(\underline{d}^{\star}, \underline{i}^{\star},a)\ D_{\textsf{KL}}((P_{C}^{a})^{d^{\star}_{a}}(\cdot|i^{\star}_{a})\| (P_{C'}^{a})^{d^{\star}_{a}}(\cdot|i^{\star}_{a})) - 2\varepsilon > 0, 
    \label{eq:choice-of-varepsilon}
\end{equation}
we see that the left hand side of the probability term in \eqref{eq:term-1} is strictly positive, whereas the right hand side goes to $0$ as $n\to\infty$. Thus, for all sufficiently large values of $n$, \eqref{eq:term-1} equals $0$.

It now remains to show that \eqref{eq:term-2} is $O(1/n^{3})$. 
\vspace{0.5cm}

\underline{\textbf{Showing that \eqref{eq:term-2} is $O(1/n^{3})$:}}\\
Let $M$ be a large, positive integer such that \eqref{eq:P_a^M_strictly_pos_entries} holds for all $m\geq M$. 
Along the lines of the proof of irreducibility presented in Appendix \ref{appndx:proof-of-lemma-ergodicity-under-unif-SRS-policy}, it can be shown that for all $(\underline{d}, \underline{i})\in \mathbb{S}_{R}$ and $T_{0}$,
 \begin{align}
P_{C}^{\pi}(\underline{d}(T_{0}+N)=\underline{d}^{\star}, \underline{i}(T_{0}+N)=\underline{i}^{\star}|\underline{d}(T_{0})=\underline{d}, \underline{i}(T_{0})=\underline{i}) \geq \left(\frac{\eta}{K}\ \mu_{R}^{\textsf{min}}\right)^{M+K}\ \bar{\varepsilon}^{K}
	\label{eq:verification_of_assumption_a5_1}
\end{align}
for $N=M+K$, where $\mu_{R}^{\textsf{min}}$ is as defined in \eqref{eq:mu-min}, and $\bar{\varepsilon}$ is as defined in \eqref{eq:bar-varepsilon}. Eq. \eqref{eq:verification_of_assumption_a5_1} states that under the policy {\sc $R$-DCR-BAI}, the probability of starting from any $(\underline{d}, \underline{i})\in \mathbb{S}_{R}$ and reaching the state $(\underline{d}^{\star}, \underline{i}^{\star})$ after $N=M+K$ time instants may be lower bounded uniformly over all starting states. In the literature on controlled Markov processes, such a phenomenon is referred to as Doeblin's minorisation condition \cite[Eq.~(5)]{kontoyiannis2005relative}. Writing $\rho$ to denote the constant on the right hand side of \eqref{eq:verification_of_assumption_a5_1}, we get that for all $n\geq M+K$, 
\begin{align}
&P_{C}^{\pi}(\underline{d}(n)=\underline{d}^{\star}, \underline{i}(n)=\underline{i}^{\star}) \nonumber\\
&=\sum_{(\underline{d}, \underline{i})\in \mathbb{S}_{R}}\ \Big[P_{C}^{\pi}(\underline{d}(n)=\underline{d}^{\star}, \underline{i}(n)=\underline{i}^{\star}|\underline{d}(n-M-K)=\underline{d}, \underline{i}(n-M-K)=\underline{i})\nonumber\\
&\hspace{5cm} \cdot P_{C}^{\pi}(\underline{d}(n-M-K)=\underline{d}, \underline{i}(n-M-K)=\underline{i})\Big]\nonumber\\
&\geq \rho\cdot \sum_{(\underline{d}, \underline{i})\in \mathbb{S}_{R}}\  P_{C}^{\pi}(\underline{d}(n-M-K)=\underline{d}, \underline{i}(n-M-K)=\underline{i})\nonumber\\
& = \rho.
\label{eq:doeblin-condition-consequence}
\end{align}
We now note that for all $n\geq M+K$,
\begin{align}
N(n, \underline{d}^{\star}, \underline{i}^{\star}, a) &=\sum_{t=K}^{n}\ \mathbb{I}_{\{\underline{d}(t)=\underline{d}^{\star}, \ \underline{i}(t)=\underline{i}^{\star}, \ A_{t}=a\}}\nonumber\\
&\geq  \sum_{t=M+K}^{n}\ \mathbb{I}_{\{\underline{d}(t)=\underline{d}^{\star}, \ \underline{i}(t)=\underline{i}^{\star}, \ A_{t}=a\}}\quad  \text{almost surely}.
\label{eq:exp-upper-bound-2}
\end{align}
Denoting the right hand side of \eqref{eq:exp-upper-bound-2} by $N'(n,\underline{d}^{\star}, \underline{i}^{\star}, a)$, we note that
\begin{align}
\mathbb{E}_{C}^{\pi}[N'(n,\underline{d}^{\star}, \underline{i}^{\star}, a)] &= \sum_{t=M+K}^{n}\ P_{C}^{\pi}(\underline{d}(t)=\underline{d}^{\star}, \ \underline{i}(t)=\underline{i}^{\star}, \ A_{t}=a)\nonumber\\
&\geq \sum_{t=M+K}^{n}\ \rho\cdot P_{C}^{\pi}(A_{t}=a|\underline{d}(t)=\underline{d}^{\star}, \ \underline{i}(t)=\underline{i}^{\star})\nonumber\\
&=\sum_{t=M+K}^{n}\ \rho\cdot \lambda_{\eta, R, \bar{C}(t)}(a|\underline{d}^{\star}, \underline{i}^{\star})\nonumber\\
&\geq (n-M-K+1)\cdot \rho \cdot \left(\frac{\eta}{K}\ \mu_{R}^{\textsf{min}}\right),
\label{eq:expected-N'(n,d,i,a)}
\end{align}
where $\mu_{R}^{\textsf{min}}$ is as defined in \eqref{eq:mu-min}.

For all $n\geq M+K$, we then have
\begin{align}
	&P_{C}^\pi\bigg(N(n, \underline{d}^{\star}, \underline{i}^{\star}, a) < n\,\nu_{\eta,R,C}(\underline{d}^{\star}, \underline{i}^{\star}, a)(1-\epsilon')\bigg)\nonumber\\
	&\leq P_{C}^\pi\bigg(N'(n, \underline{d}^{\star}, \underline{i}^{\star}, a) < n\,\nu_{\eta,R,C}(\underline{d}^{\star}, \underline{i}^{\star}, a)(1-\epsilon')\bigg)\nonumber\\
	&=P_{C}^\pi\bigg(N'(n, \underline{d}^{\star}, \underline{i}^{\star}, a) -\mathbb{E}_{C}^\pi[N'(n, \underline{d}^{\star}, \underline{i}^{\star}, a)] < n\,\nu_{\eta,R,C}(\underline{d}^{\star}, \underline{i}^{\star}, a)(1-\epsilon') - \mathbb{E}_{C}^\pi[N'(n, \underline{d}^{\star}, \underline{i}^{\star},a)]\bigg)\nonumber\\
	&\leq P_{C}^\pi\bigg(N'(n, \underline{d}^{\star}, \underline{i}^{\star}, a) -\mathbb{E}_{C}^\pi[N'(n, \underline{d}^{\star}, \underline{i}^{\star}, a)] < n\,\nu_{\eta,R,C}(\underline{d}^{\star}, \underline{i}^{\star}, a)(1-\epsilon') \nonumber\\
	&\hspace{8.5cm} - (n-M-K+1)\cdot \rho\cdot \frac{\eta}{K}\cdot \mu_{R}^{\textsf{min}}\bigg)\nonumber\\
	&\leq P_{C}^\pi\bigg(N'(n, \underline{d}^{\star}, \underline{i}^{\star}, a) -\mathbb{E}_{C}^\pi[N'(n, \underline{d}^{\star}, \underline{i}^{\star}, a)] < n\bigg\lbrace\nu_{\eta,R,C}(\underline{d}^{\star}, \underline{i}^{\star}, a)(1-\epsilon') \nonumber\\
	&\hspace{8cm}- \left(\frac{n-M-K+1}{n}\right)\cdot \frac{\rho\,\eta\, \mu_{R}^{\textsf{min}}}{K} \bigg\rbrace\bigg).
	\label{eq:exp_bound_13}
\end{align}
Noting that $$ \frac{n-M-K+1}{n} \geq  \frac{1}{2} $$ for all $n$ sufficiently large, we choose $\epsilon'$ such that $$ \nu_{\eta,R,C}(\underline{d}^{\star}, \underline{i}^{\star}, a)(1-\epsilon') - \frac{\rho\,\eta\, \mu_{R}^{\textsf{min}}}{2K} < 0. $$ For instance, it suffices to set
\begin{equation}
    \epsilon'= 1 - \frac{1}{2}\ \frac{\rho\,\eta\, \mu_{R}^{\textsf{min}}}{2K\ \nu_{\eta,R,C}(\underline{d}^{\star}, \underline{i}^{\star}, a)}. 
    \label{eq:choice-of-epsilon-prime}
\end{equation}
For this choice of $\epsilon'$, it follows that the probability term in \eqref{eq:exp_bound_13} may be bounded above exponentially by using concentration inequalities for sub-gaussian random variables \cite[p. 25]{boucheron2013concentration}; here, $N'(n, \underline{d}^{\star}, \underline{i}^{\star}, a)$ is a sum of (not necessarily independent) indicator random variables, each of which is sub-gaussian with variance factor $1/4$. This implies that $N'(n, \underline{d}^{\star}, \underline{i}^{\star}, a)$ is also sub-gaussian \cite[Lemma 17]{karthik2021learning}. Therefore, for all sufficiently large $n$, the probability term in \eqref{eq:exp_bound_13} is $O(1/n^{3})$.
\end{enumerate}

\section{Proof of Lemma \ref{lemma:lim-T-R-star-matches-T-star-when-TPMs-have-identical-rows}}
\label{appndx:proof-of-convergence-of-T-R-star-to-T-star}

Suppose that $P_k(\cdot|i)=\mu_k(\cdot)$ for all $k=1,\ldots, K$ and $i\in \mathcal{S}$. Fixing $C\in \mathcal{C}$, it follows that for all $d\in \mathbb{N}$, $i\in \mathcal{S}$, and $C'\in \textsf{Alt}(C)$,
\begin{align}
    D_{\textsf{KL}}((P_{C}^{a})^{d}(\cdot|i) \| (P_{C'}^{a})^{d}(\cdot|i)) = D_{\textsf{KL}}(\mu_C^a \| \mu_{C'}^a),
    \label{eq:KL-divergence-does-not-depend-on-d-i}
\end{align}
where $\mu_C^a$ denotes the stationary distribution associated with the TPM $P_C^a$. As a consequence of \eqref{eq:KL-divergence-does-not-depend-on-d-i}, we have
\begin{align}
    T^\star(C) 
    &= \sup_{\nu}\ \min_{C'\in \textsf{Alt}(C)} \ \sum_{(\underline{d},\underline{i})\in\mathbb{S}}\ \sum_{a=1}^{K}\  \nu(\underline{d},\underline{i},a) \ D_{\textsf{KL}}((P_C^a)^{d_a}(\cdot|i_a)\|(P_{C'}^a)^{d_a}(\cdot|i_a))\nonumber\\
    &=\sup_{\nu}\ \min_{C'\in \textsf{Alt}(C)} \ \sum_{(\underline{d},\underline{i})\in\mathbb{S}}\ \sum_{a=1}^{K}\  \nu(\underline{d},\underline{i},a) \ D_{\textsf{KL}}(\mu_C^a \| \mu_{C'}^a)\nonumber\\
    &=\sup_{\kappa}\ \min_{C'\in \textsf{Alt}(C)}\ \sum_{a=1}^{K}\ \kappa(a) \  D_{\textsf{KL}}(\mu_C^a \| \mu_{C'}^a),
    \label{eq:modified-T-star-C}
\end{align}
where $\kappa(a)\coloneqq \sum_{(\underline{d}, \underline{i})\in \mathbb{S}} \nu(\underline{d}, \underline{i}, a)$ for all $a\in \mathcal{A}$, and the supremum in \eqref{eq:modified-T-star-C} is over all $\kappa$ which are probability distributions on the set of arms $\mathcal{A}$. 

For a fixed $R\in \mathbb{N}\cap (K, \infty)$, suppose that $$ \mathbb{S}_1 = \bigcup_{a=1}^{K}\ \mathbb{S}_{R,a}, \quad \mathbb{S}_2=\mathbb{S}_R \setminus \mathbb{S}_1. $$ Then,
\begin{align}
    & T_R^\star(C) \nonumber\\
    &=  \sup_{\nu}\ \min_{C'\in \textsf{Alt}(C)} \ \sum_{(\underline{d},\underline{i})\in\mathbb{S}_R}\ \sum_{a=1}^{K}\  \nu(\underline{d},\underline{i},a) \ D_{\textsf{KL}}((P_C^a)^{d_a}(\cdot|i_a)\|(P_{C'}^a)^{d_a}(\cdot|i_a))\nonumber\\
    &=\sup_{\nu}\ \min_{C'\in \textsf{Alt}(C)} \ \sum_{(\underline{d},\underline{i})\in\mathbb{S}_R}\ \sum_{a=1}^{K}\  \nu(\underline{d},\underline{i},a) \ D_{\textsf{KL}}(\mu_C^a \| \mu_{C'}^a)\nonumber\\
    &=\sup_{\nu}\ \min_{C'\in \textsf{Alt}(C)} \bigg\lbrace  \sum_{(\underline{d},\underline{i})\in\mathbb{S}_1}\ \sum_{a=1}^{K}\  \nu(\underline{d},\underline{i},a) \ D_{\textsf{KL}}(\mu_C^a \| \mu_{C'}^a) + \sum_{(\underline{d},\underline{i})\in\mathbb{S}_2}\ \sum_{a=1}^{K}\  \nu(\underline{d},\underline{i},a) \ D_{\textsf{KL}}(\mu_C^a \| \mu_{C'}^a)\bigg\rbrace\nonumber\\
    &=\sup_{\nu}\ \min_{C'\in \textsf{Alt}(C)} \bigg\lbrace  \sum_{a=1}^{K}\ \sum_{(\underline{d},\underline{i})\in\mathbb{S}_{R,a}}\ \nu(\underline{d},\underline{i},a) \ D_{\textsf{KL}}(\mu_C^{a} \| \mu_{C'}^{a})  + \sum_{(\underline{d},\underline{i})\in\mathbb{S}_2}\ \sum_{a=1}^{K}\  \nu(\underline{d},\underline{i},a) \ D_{\textsf{KL}}(\mu_C^a \| \mu_{C'}^a)\bigg\rbrace\nonumber\\
    &=\sup_{\nu}\ \min_{C'\in \textsf{Alt}(C)} \bigg\lbrace  \sum_{a=1}^{K}\ D_{\textsf{KL}}(\mu_C^a \| \mu_{C'}^a)\ \bigg(\sum_{(\underline{d},\underline{i})\in\mathbb{S}_{R,a}} \nu(\underline{d},\underline{i},a) + \sum_{(\underline{d},\underline{i})\in\mathbb{S}_2}\   \nu(\underline{d},\underline{i},a) \bigg) \bigg\rbrace\nonumber\\
    &\stackrel{(a)}{=}\sup_{\nu}\ \min_{C'\in \textsf{Alt}(C)} \bigg\lbrace  \sum_{a=1}^{K}\ D_{\textsf{KL}}(\mu_C^a \| \mu_{C'}^a)\ \bigg(\sum_{(\underline{d},\underline{i})\in\mathbb{S}_{R,a}} \nu(\underline{d},\underline{i},a) + \sum_{a'\neq a}\ \sum_{(\underline{d},\underline{i})\in\mathbb{S}_{R,a'}} \nu(\underline{d},\underline{i},a) +  \sum_{(\underline{d},\underline{i})\in\mathbb{S}_2}\   \nu(\underline{d},\underline{i},a) \bigg) \bigg\rbrace\nonumber\\
    &=\sup_{\nu}\ \min_{C'\in \textsf{Alt}(C)} \bigg\lbrace  \sum_{a=1}^{K}\ D_{\textsf{KL}}(\mu_C^a \| \mu_{C'}^a)\ \bigg(\sum_{(\underline{d},\underline{i})\in\mathbb{S}_R} \nu(\underline{d},\underline{i},a)\bigg) \bigg\rbrace\nonumber\\
    &=\sup_{\kappa}\ \min_{C'\in \textsf{Alt}(C)}\ \sum_{a=1}^{K}\ \kappa(a) \  D_{\textsf{KL}}(\mu_C^a \| \mu_{C'}^a),
    \label{eq:modified-T-R-star-C}
\end{align}
where $(a)$ above follows from the observation that any $\nu$ participating in the supremum in $(a)$ meets the $R$-max-delay constraint in \eqref{eq:additional-condition-with-delay-restriction}, and therefore satisfies $\nu(\underline{d}, \underline{i}, a)=0$ for all $(\underline{d}, \underline{i})\in \mathbb{S}_{R,a'}$, $a'\neq a$. From \eqref{eq:modified-T-star-C} and \eqref{eq:modified-T-R-star-C}, we see that $T_R^\star(C)=T^\star(C)$ for all $R$, thus proving that $\lim_{R\to\infty} T_R^\star(C)=T^\star(C)$ in the special case when each of the arm TPMs have identical rows.

\IEEEpeerreviewmaketitle

\bibliographystyle{IEEEtran}
\bibliography{myref2-with-delay-constraints}

\begin{thebibliography}{10}
\providecommand{\url}[1]{#1}
\csname url@samestyle\endcsname
\providecommand{\newblock}{\relax}
\providecommand{\bibinfo}[2]{#2}
\providecommand{\BIBentrySTDinterwordspacing}{\spaceskip=0pt\relax}
\providecommand{\BIBentryALTinterwordstretchfactor}{4}
\providecommand{\BIBentryALTinterwordspacing}{\spaceskip=\fontdimen2\font plus
\BIBentryALTinterwordstretchfactor\fontdimen3\font minus
  \fontdimen4\font\relax}
\providecommand{\BIBforeignlanguage}[2]{{%
\expandafter\ifx\csname l@#1\endcsname\relax
\typeout{** WARNING: IEEEtran.bst: No hyphenation pattern has been}%
\typeout{** loaded for the language `#1'. Using the pattern for}%
\typeout{** the default language instead.}%
\else
\language=\csname l@#1\endcsname
\fi
#2}}
\providecommand{\BIBdecl}{\relax}
\BIBdecl

\bibitem{Chernoff1959}
H.~Chernoff, ``Sequential design of experiments,'' \emph{The Annals of
  Mathematical Statistics}, vol.~30, no.~3, pp. 755--770, 1959.

\bibitem{albert1961sequential}
A.~E. Albert, ``The sequential design of experiments for infinitely many states
  of nature,'' \emph{The Annals of Mathematical Statistics}, pp. 774--799,
  1961.

\bibitem{garivier2016optimal}
A.~Garivier and E.~Kaufmann, ``Optimal best arm identification with fixed
  confidence,'' in \emph{Conference on Learning Theory,}.\hskip 1em plus 0.5em
  minus 0.4em\relax PMLR, 2016, pp. 998--1027.

\bibitem{moulos2019optimal}
V.~Moulos, ``Optimal best {M}arkovian arm identification with fixed
  confidence,'' \emph{Advances in Neural Information Processing Systems},
  vol.~32, 2019.

\bibitem{Kaufmann2016}
E.~Kaufmann, O.~Capp{\'e}, and A.~Garivier, ``On the complexity of best-arm
  identification in multi-armed bandit models,'' \emph{Journal of Machine
  Learning Research}, vol.~17, no.~1, pp. 1--42, 2016.

\bibitem{karthik2021detecting}
P.~N. Karthik and R.~Sundaresan, ``Detecting an odd restless {M}arkov arm with
  a trembling hand,'' \emph{IEEE Transactions on Information Theory}, vol.~67,
  no.~8, pp. 5230--5258, 2021.

\bibitem{karthik2021learning}
------, ``Learning to detect an odd restless {M}arkov arm with a trembling
  hand,'' \emph{arXiv preprint arXiv:2105.03603}, 2021.

\bibitem{lai1985asymptotically}
T.~L. Lai and H.~Robbins, ``Asymptotically efficient adaptive allocation
  rules,'' \emph{Advances in Applied Mathematics}, vol.~6, no.~1, pp. 4--22,
  1985.

\bibitem{anantharam1987asymptotically}
V.~Anantharam, P.~Varaiya, and J.~Walrand, ``Asymptotically efficient
  allocation rules for the multiarmed bandit problem with multiple plays-part
  ii: Markovian rewards,'' \emph{IEEE Transactions on Automatic Control},
  vol.~32, no.~11, pp. 977--982, 1987.

\bibitem{bubeck2012regret}
S.~Bubeck and N.~Cesa-Bianchi, ``Regret analysis of stochastic and
  nonstochastic multi-armed bandit problems,'' \emph{Machine Learning}, vol.~5,
  no.~1, pp. 1--122, 2012.

\bibitem{mannor2004sample}
S.~Mannor and J.~N. Tsitsiklis, ``The sample complexity of exploration in the
  multi-armed bandit problem,'' \emph{Journal of Machine Learning Research},
  vol.~5, no.~6, pp. 623--648, 2004.

\bibitem{even2002pac}
E.~Even-Dar, S.~Mannor, and Y.~Mansour, ``Pac bounds for multi-armed bandit and
  {M}arkov decision processes,'' in \emph{International Conference on
  Computational Learning Theory}.\hskip 1em plus 0.5em minus 0.4em\relax
  Springer, 2002, pp. 255--270.

\bibitem{karnin2013almost}
Z.~Karnin, T.~Koren, and O.~Somekh, ``Almost optimal exploration in multi-armed
  bandits,'' in \emph{International Conference on Machine Learning}.\hskip 1em
  plus 0.5em minus 0.4em\relax PMLR, 2013, pp. 1238--1246.

\bibitem{jamieson2014lil}
K.~Jamieson, M.~Malloy, R.~Nowak, and S.~Bubeck, ``lil'ucb: An optimal
  exploration algorithm for multi-armed bandits,'' in \emph{Conference on
  Learning Theory}.\hskip 1em plus 0.5em minus 0.4em\relax PMLR, 2014, pp.
  423--439.

\bibitem{audibert2010best}
J.-Y. Audibert, S.~Bubeck, and R.~Munos, ``Best arm identification in
  multi-armed bandits.'' in \emph{Conference on Learning Theory}.\hskip 1em
  plus 0.5em minus 0.4em\relax JMLR, 2010, pp. 41--53.

\bibitem{kalyanakrishnan2012pac}
S.~Kalyanakrishnan, A.~Tewari, P.~Auer, and P.~Stone, ``Pac subset selection in
  stochastic multi-armed bandits.'' in \emph{International Conference on
  Machine Learning}, vol.~12.\hskip 1em plus 0.5em minus 0.4em\relax PMLR,
  2012, pp. 655--662.

\bibitem{kaufmann2013information}
E.~Kaufmann and S.~Kalyanakrishnan, ``Information complexity in bandit subset
  selection,'' in \emph{Conference on Learning Theory}.\hskip 1em plus 0.5em
  minus 0.4em\relax PMLR, 2013, pp. 228--251.

\bibitem{gupta2021best}
S.~Gupta, G.~Joshi, and O.~Ya{\u{g}}an, ``Best arm identification in correlated
  multi-armed bandits,'' \emph{IEEE Journal on Selected Areas in Information
  Theory}, vol.~2, no.~2, pp. 549--563, 2021.

\bibitem{zhong2020best}
Z.~Zhong, W.~C. Cheung, and V.~Y.~F. Tan, ``Best arm identification for
  cascading bandits in the fixed confidence setting,'' in \emph{International
  Conference on Machine Learning}.\hskip 1em plus 0.5em minus 0.4em\relax PMLR,
  2020, pp. 11\,481--11\,491.

\bibitem{dekel2014bandits}
O.~Dekel, J.~Ding, T.~Koren, and Y.~Peres, ``Bandits with switching costs:
  ${T}^{2/3}$ regret,'' in \emph{Proceedings of the ACM Symposium on Theory of
  Computing}, 2014, pp. 459--467.

\bibitem{zhong2021prob}
Z.~Zhong, W.~C. Cheung, and V.~Y.~F. Tan, ``Probabilistic sequential shrinking:
  A best arm identification algorithm for stochastic bandits with
  corruptions,'' in \emph{International Conference on Machine Learning}.\hskip
  1em plus 0.5em minus 0.4em\relax PMLR, 2021, pp. 12\,772--12\,781.

\bibitem{karthik2020learning}
P.~N. Karthik and R.~Sundaresan, ``Learning to detect an odd {M}arkov arm,''
  \emph{IEEE Transactions on Information Theory}, vol.~66, no.~7, pp.
  4324--4348, 2020.

\bibitem{deshmukh2018controlled}
A.~Deshmukh, S.~Bhashyam, and V.~V. Veeravalli, ``Controlled sensing for
  composite multihypothesis testing with application to anomaly detection,'' in
  \emph{Asilomar Conference on Signals, Systems, and Computers}.\hskip 1em plus
  0.5em minus 0.4em\relax IEEE, 2018, pp. 2109--2113.

\bibitem{deshmukh2021sequential}
A.~Deshmukh, V.~V. Veeravalli, and S.~Bhashyam, ``Sequential controlled sensing
  for composite multihypothesis testing,'' \emph{Sequential Analysis}, vol.~40,
  no.~2, pp. 259--289, 2021.

\bibitem{prabhu2020sequential}
G.~R. Prabhu, S.~Bhashyam, A.~Gopalan, and R.~Sundaresan, ``Sequential
  multi-hypothesis testing in multi-armed bandit problems: An approach for
  asymptotic optimality,'' \emph{IEEE Transactions on Information Theory},
  2022.

\bibitem{borkar1988control}
V.~S. Borkar, ``Control of {M}arkov chains with long-run average cost
  criterion,'' in \emph{Stochastic Differential Systems, Stochastic Control
  Theory and Applications}.\hskip 1em plus 0.5em minus 0.4em\relax Springer,
  1988, pp. 57--77.

\bibitem{levin2017markov}
D.~A. Levin and Y.~Peres, \emph{Markov Chains and Mixing Times}.\hskip 1em plus
  0.5em minus 0.4em\relax American Mathematical Society, 2017, vol. 107.

\bibitem{victor1999general}
H.~Victor, ``A general class of exponential inequalities for martingales and
  ratios,'' \emph{The Annals of Probability}, vol.~27, no.~1, pp. 537--564,
  1999.

\bibitem{thomas2006elements}
M.~Thomas and J.~A. Thomas, \emph{Elements of Information Theory}.\hskip 1em
  plus 0.5em minus 0.4em\relax Wiley-Interscience, 2006.

\bibitem{shiryaev2016probability}
A.~N. Shiryaev, \emph{Probability-1}.\hskip 1em plus 0.5em minus 0.4em\relax
  Springer, 2016, vol.~95.

\bibitem{chow2012probability}
Y.~S. Chow and H.~Teicher, \emph{Probability Theory: Independence,
  Interchangeability, Martingales}.\hskip 1em plus 0.5em minus 0.4em\relax
  Springer Science \& Business Media, 2012.

\bibitem{kontoyiannis2005relative}
I.~Kontoyiannis, L.~A. Lastras-Monta{\~n}o, and S.~P. Meyn, ``Relative entropy
  and exponential deviation bounds for general {M}arkov chains,'' in
  \emph{Proceedings of International Symposium on Information Theory,}.\hskip
  1em plus 0.5em minus 0.4em\relax IEEE, 2005, pp. 1563--1567.

\bibitem{boucheron2013concentration}
S.~Boucheron, G.~Lugosi, and P.~Massart, \emph{Concentration {I}nequalities:
  {A} {N}onasymptotic {T}heory of {I}ndependence}.\hskip 1em plus 0.5em minus
  0.4em\relax Oxford {U}niversity {P}ress, 2013.

\end{thebibliography}
\end{document}